%% file: main.tex
\documentclass[runningheads]{llncs}
\usepackage{graphicx}

\usepackage{tikz}
\usepackage{comment}
\usepackage{amsmath,amssymb} %
\usepackage{color}
\usepackage{booktabs}
\usepackage{multirow}

\usepackage{caption}
\usepackage{subcaption}

\usepackage{pgfplots} 
\usetikzlibrary{patterns}

\newcommand{\mathbbm}[1]{\text{\usefont{U}{bbm}{m}{n}#1}} 
\newcommand{\norm}[1]{\left\lVert#1\right\rVert}

\newcommand{\PLH}{{\mkern-2mu\times\mkern-2mu}}

\usepackage[accsupp]{axessibility}  %

\DeclareMathOperator*{\argmax}{arg\,max}

\usepackage{xspace}
\newcommand*{\ie}{i.e.\@\xspace}
\newcommand*{\eg}{e.g.\@\xspace}

\usepackage{longtable}
\begin{document}
\pagestyle{headings}
\mainmatter
\def\ECCVSubNumber{499}  %

\title{Demystifying Unsupervised Semantic Correspondence Estimation} %

\titlerunning{Demystifying Unsupervised Semantic Correspondence Estimation}
\author{Mehmet Ayg\"un \quad Oisin Mac Aodha\index{Mac Aodha, Oisin}}
\authorrunning{Ayg\"un and Mac Aodha}

\institute{University of Edinburgh\\
\url{https://mehmetaygun.github.io/demistfy}
}
\maketitle

\begin{abstract}

We explore semantic correspondence estimation through the lens of unsupervised learning. 
We thoroughly evaluate several recently proposed unsupervised methods across multiple challenging datasets using a standardized evaluation protocol where we vary factors such as the backbone architecture, the pre-training strategy, and the pre-training and finetuning datasets. 
To better understand the failure modes of these methods, and in order to provide a clearer path for improvement, we provide a new diagnostic framework along with a new performance metric that is better suited to the semantic matching task. 
Finally, we introduce a new unsupervised correspondence approach which utilizes the strength of pre-trained features while encouraging better matches during training. 
This results in significantly better matching performance compared to current state-of-the-art methods. 

\end{abstract}

\section{Introduction}

In metaphysics, the correspondence theory of truth posits that without the notion of correspondence, there cannot be truth~\cite{david2016correspondence}. 
Analogously, correspondence estimation also holds a very important place as one of the core problems in computer vision. %
The ability to reliably obtain accurate pixel-level correspondence underpins a diverse range of tasks from stereo estimation, optical flow, structure-from-motion, through to visual tracking. 
Distinct from these lower-level objectives, semantic correspondence estimation, the task of matching different regions, parts, and landmarks across distinct object instances, is crucial to developing systems that can perform higher-level visual reasoning in diverse environments with objects that can vary significantly in both appearance and the configuration of their constituent parts.  

Manually obtaining semantic correspondence supervision, for example in the form of annotated object landmarks, is an arduous and time consuming task. 
As a result, several works have instead attempted to understand to what extent semantic regions and parts emerge from conventionally trained supervised image classification networks~\cite{long2014convnets,zeiler2014visualizing,zhou2016cvpr,gonzalez2018semantic}. 
These works have shown such semantic information is indeed present in the representations encoded by these networks, at least to some degree. 
Recently, a body of work has emerged that aims to learn semantic correspondence through self-supervision alone, \ie without the need for ground truth supervision at training time~\cite{thewlis2017unsupervised,thewlis2019unsupervised,cheng2021equivariant,karmali2022lead}.  

While we have observed progress on unsupervised semantic correspondence estimation, a number of questions are still underexplored and unanswered. 
For instance, it is not clear how well current approaches generalize beyond more simplified object categories such as human faces to more complex non-rigidly deforming categories that vary in terms of both pose and appearance. 
Recent works have also been able to avail of advances in self-supervised learning of general visual representations \cite{cheng2021equivariant,karmali2022lead}, thus making it difficult to properly understand how they compare to older methods that do not utilize such self-supervised pre-training.
In this work, we attempt to shine light on the above questions in addition to exploring the role other factors such as the impact of pre-training and finetuning data, backbone models, and the underlying evaluation criteria used to asses performance. 
Inspired by detailed benchmarking investigation in human pose estimation~\cite{ruggero2017benchmarking}, we provide a thorough evaluation of the success and failure modes of current methods to provide guidance for future progress. 

We make the following three contributions: 
(i) A standardized evaluation of multiple existing approaches for unsupervised semantic correspondence estimation across five challenging datasets. 
(ii) A new, conceptually simple, unsupervised training objective that results in superior semantic matching performance. %
(iii) A detailed breakdown of the current failure cases for current best performing approaches and our proposed new unsupervised method.

\section{Related Work} 

\noindent\textbf{Supervised Semantic Correspondence.} Pre-deep learning work tackled semantic correspondence estimation as a local region matching problem using hand crafted features~\cite{liu2010sift,kim2013deformable,bristow2015dense}, or as offset matching using object proposals~\cite{ham2016proposal}. 
In the deep learning era, several works investigated if objects parts and regions emerge from image classification models \cite{zeiler2014visualizing,zhou2016cvpr,gonzalez2018semantic}, \ie models trained only with image-level class supervision. 
\cite{long2014convnets} showed that deep CNN features could actually be used for semantic matching. 
Subsequent work built on this by proposing new architectures specifically designed for semantic matching \cite{choy2016universal,han2017scnet,kim2017fcss,rocco2017convolutional,huang2019dynamic,lee2019sfnet,kim2018recurrent}. 
Some of these approaches focused on combining multilevel features (\ie hypercolumn features) from deep networks~\cite{ufer2017deep,min2019hyperpixel,min2020learning,zhao2021multi}, aggregating information from features using 4D convolutions~\cite{rocco2018neighbourhood,rocco2020efficient,li2020correspondence,lee2021patchmatch}, leveraging geometric relations via Hough transforms~\cite{min2021convolutional}, or using optimal transport~\cite{sarlin2020superglue,liu2020semantic}. 
Some matching methods formulate the problem as one of flow estimation between images~\cite{liu2010sift,min2019hyperpixel}. 
However, unlike optical flow, semantic correspondence methods need to be able to handle intra and inter class variations when matching points.
Recently, the use of transformer-based models has also been explored~\cite{cho2021cats,jiang2021cotr}. 
In contrast to most of the above works, we focus on the unsupervised setting, whereby no supervised keypoint annotations are used to train our models. 

\noindent\textbf{Unsupervised Semantic Correspondence.} 
Recent progress in self-supervised learning has resulted in a suite of methods that are capable of extracting discriminative whole image representations without requiring explicit supervision \cite{van2018representation,wu2018unsupervised,chen2020simple,grill2020bootstrap,he2020momentum}. 
While the majority of these methods optimize objectives to discriminate global image representations by using augmented image pairs, \cite{cheng2021equivariant,karmali2022lead} showed that these approaches can also be utilized in correspondence estimation. Recently, several approaches proposed optimizing alternative objectives on a denser level \cite{roh2021spatially,wei2021aligning,araslanov2021dense,o2020unsupervised,wang2021dense,wang2021exploring,zhong2021pixel}. 
However, these methods have been applied to tasks such as object detection and segmentation, but not directly for semantic correspondence. 
Another line of work proposed methods to discover semantic keypoint locations in an unsupervised way \cite{jakab2018unsupervised,zhang2018unsupervised,kulkarni2019unsupervised,jakab2020self,ryou2021weakly}.

For the problem of correspondence estimation, images augmented with artificial spatial deformations were used by~\cite{kanazawa2016warpnet,rocco2017convolutional} to learn transformations between image pairs without any external supervision. 
Instead of learning a function to match image pairs, \cite{thewlis2017unsupervised,thewlis2017unsupervisedneurips} framed the problem as one of learning a function that can extract local features which can be used for semantic matching across all instances of a category of interest. To introduce greater invariance for intra-category differences, DVE~\cite{thewlis2019unsupervised} extended EQ~\cite{thewlis2017unsupervisedneurips} with the use of additional non-augmented auxiliary images during training.

More recent work has been able to make use of advances in self-supervised learning in order to learn more effective representations. 
CL~\cite{cheng2021equivariant} proposed a two-stage approach, combining image-level instance-based discrimination~\cite{he2020momentum} together with dense equivariant learning. 
They trained a linear projection head on top of frozen learned features computed via an image-level self-supervised pre-training task, where the goal of the projection step was to enforce the dense features to be spatially distinct within an image. 
LEAD~\cite{karmali2022lead} also followed a similar two-stage approach, starting with instance-level discrimination using~\cite{grill2020bootstrap}. 
In the second stage, instead of encouraging the features to be spatially distinct, their projection operation minimized the dissimilarity between feature correlation maps from the instance-level features and correlation maps from the projected features. 
This can be viewed as a form of dimensionality reduction as the projected features are smaller in size compared to the original features. 

The above methods, while effective on some datasets, have limitations. 
EQ~\cite{thewlis2017unsupervisedneurips} is only able to learn invariances that can be expressed via image augmentations. 
DVE~\cite{thewlis2019unsupervised} assumes that the images have the same visible keypoints, and can thus be negatively impacted by incorrect matches on background pixels. 
The projection step used by CL~\cite{cheng2021equivariant} runs the risk of discarding  invariances learned during the pre-training stage.
While LEAD~\cite{karmali2022lead} maintains learned invariances from the first stage, if the pre-trained features generate incorrect matches, their loss can end up optimizing possibly incorrect feature correlations. 
In this work, we thoroughly benchmark the performance of these approaches by evaluating them on several challenging datasets. 
We also propose a new semantic correspondence loss, which learns more effective dense features by both preserving the learned invariances while also making the features more distinct. 
   
\noindent\textbf{Performance Evaluation and Error Diagnosis.} Benchmarking model performance with a single summary metric is one of the best tools that we have for objectively measuring progress on a given task. 
However, accurately understanding the limitations and improvements provided by new methods is even more crucial for future progress.
Several works have introduced different diagnostic tools and frameworks to analyze methods across a variety of problems \cite{hoiem2012diagnosing,russakovsky2013detecting,everingham2015pascal,zhang2016far,sigurdsson2017actions,alwassel2018diagnosing}.
For the semantic correspondence problem, the vast majority of existing works only report performance via single summary metrics, \eg the Percentage of Correct Keypoints (PCK) with a fixed distance threshold. 
This allows us to get an overall sense of performance but does not reveal \emph{why} a given method performs better than others. 
Recent works~\cite{musgrave2020metric,choe2020evaluating} have emphasized the importance of detailed evaluation in order to better understand what components specific performance improvements can be attributed to. 
In this work, in the spirit of~\cite{ruggero2017benchmarking}, we introduce a more thorough evaluation for analyzing semantic correspondence methods. 
We also propose a new version of PCK which better captures correspondence errors and present standardized baseline results across multiple datasets to fairly compare semantic correspondence performance.

\section{Semantic Correspondence Estimation}
\label{sec:semantic_correspondence}

\subsection{Problem Setup}
Given a source-target image pair, $\mathbf{x}_{s}$ and $\mathbf{x}_{t}$, the goal of correspondence estimation is to find the locations of a set of points of interest from the source image in the target image.
Unlike in optical flow or stereo estimation, where the task is to compute correspondence across time or viewpoint, in the case of semantic correspondence, the goal is to find matching locations across different depictions of the same object category. 
This is a challenging setting as the objects of interest can vary in terms of appearance, pose, and shape, in addition to difficulty arising from other nuisance factors such as the background, occlusion, and lighting.

\newcommand\myeq{\mkern1.5mu{=}\mkern1.5mu}
We pose the correspondence problem as a nearest-neighbor matching task in a learned local feature embedding space. 
Formally, for a pixel location, $u \in \Omega \myeq \{1, ..., H\} \times \{1, ..., W\}$, in a source image of size $H\times W$, we find the corresponding point $\hat{u}$ in the target image $\mathbf{x}_t$ as, $\hat{u} \myeq \argmax_{k \in \Omega} f(\Phi_{u}(\mathbf{x}_{s}), \Phi_{k}(\mathbf{x}_{t})),$ where $\Phi_{u}(\mathbf{x}_{s})$ represents an embedding vector of the point $u$ from image $\mathbf{x}_{s}$, and $f$ is a similarity function. 
We use a deep neural network as our embedding function $\Phi$, and the similarity is computed via the dot product of the $\ell_{2}$ normalized embedding vectors. 
In practice, we decompose the embedding function into a feature encoder, followed by a projection step, \ie $\Phi(\mathbf{x}) \myeq \rho(\Psi(\mathbf{x}))$, where the encoder is deep network. 
The purpose of the projection is to reduce the dimensionality of the feature, and could be a linear operation~\cite{cheng2021equivariant} or a network~\cite{karmali2022lead}.

\begin{figure}[t]
    \centering
    \includegraphics[width=\textwidth]{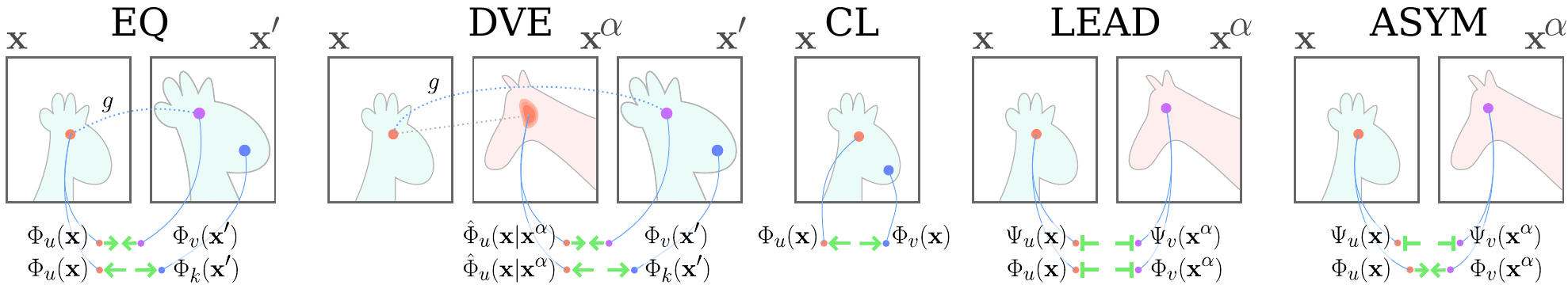}
    \caption{Unsupervised approaches for semantic correspondence estimation. 
    $\mathbf{x}'$ is a synthetically augmented version of image $\mathbf{x}$, and $\mathbf{x}^\alpha$ is a different image of the same semantic category. 
    EQ~\cite{thewlis2017unsupervisedneurips} minimizes the distance between embeddings of point pairs with known geometric transformations $g$. 
    DVE~\cite{thewlis2019unsupervised} builds on EQ by using an additional auxiliary image. 
    CL~\cite{cheng2021equivariant} maximizes distance between embeddings of points within an image. 
    LEAD~\cite{karmali2022lead} enforces the same distance between pre-trained and projected embeddings. 
    Our ASYM method extends LEAD by enforcing projected embeddings to be closer in the feature space.}
    \label{fig:overview}
\end{figure}

In the next section, we review several existing unsupervised methods designed for learning dense representations with an emphasis on matching (see Fig.~\ref{fig:overview} for an overview). 
While more sophisticated methods have been proposed for estimating semantic correspondence, \eg using optimal transport~\cite{sarlin2020superglue,liu2020semantic}, distance re-weighting with spatial regularizers~\cite{min2019hyperpixel}, or restricting the search area with class activation maps~\cite{zhou2016cvpr} as in~\cite{liu2020semantic}, we focus on learning embedding functions as recent work has shown that combining self-supervised representation learning with correspondence specific finetuning produces state of the art results~\cite{cheng2021equivariant,karmali2022lead}.

\subsection{Unsupervised Semantic Correspondence Learning}
\label{ssec:un_losses}
EQ~\cite{thewlis2017unsupervisedneurips} proposed an unsupervised method that utilizes the equivariance principle to learn dense matchable features. 
During training, their model takes an image $\mathbf{x}$ along with an augmented version of it $\mathbf{x}'$ and tries to minimize feature similarity of known corresponding pixel locations $u$ and $v$. 
Here, $\mathbf{x}'$ is derived from $\mathbf{x}$ using artificial spatial and appearance-based augmentations and the pixel coordinates $u$ and $v$ are locations from the two images which are related by a known transformation $g$, such that $v=gu$. 
They minimize the following loss,
\begin{equation}
    \label{eq:equivariance2}
        \mathcal{L}_{\mathit{eq}} = \frac{1}{|\Omega|^2} \sum_{u \in \Omega}\sum_{v \in \Omega}\|gu-v\| ~p(v | u; \Phi, \mathbf{x}, \mathbf{x}', \tau),
\end{equation}

\begin{equation}
\label{eq:softdiv}
    p(v |u; \Phi, \mathbf{x}, \mathbf{x'},\tau) = \frac{\text{exp}(\langle \Phi_u(\mathbf{x}), \Phi_v(\mathbf{x}') \rangle/\tau)}{\sum_{k \in \Omega} \text{exp}(\langle \Phi_u(\mathbf{x}), \Phi_k(\mathbf{x}') \rangle/\tau)},
\end{equation}
\noindent where $\tau$ is temperature parameter for the softmax function and $\Omega$ is the set of possible pixel locations on the image grid. 
In essence, the model aims to embed corresponding points nearby in the learned embedding space, while also pushing other points further away.

EQ uses artificially augmented image pairs and can thus only learn invariances up to those expressible by these augmentations. 
Subsequently, DVE~\cite{thewlis2019unsupervised} extended EQ using an auxiliary image, $\mathbf{x}^\alpha$, to calculate correspondence from $\mathbf{x} \rightarrow \mathbf{x}^\alpha$ and then $ \mathbf{x}^\alpha \rightarrow \mathbf{x}'$. 
This is achieved by replacing the $\Phi_u(\mathbf{x})$ term in Eqn.~\ref{eq:softdiv} with $\hat{\Phi}_u(\mathbf{x}|\mathbf{x}^\alpha) = \sum_w \Phi_u(\mathbf{x}^\alpha)p(w |u; \Phi, \mathbf{x}, \mathbf{x}^\alpha, \tau)$. 
Importantly, the ground truth correspondence to the auxiliary image does not need to be known as the mapping from $\mathbf{x} \rightarrow \mathbf{x}'$ is available. 

Recently, two stage methods for learning dense embeddings have been proposed~\cite{cheng2021equivariant,karmali2022lead}.  
In these approaches, the first stage makes use of an image-level self-supervised training objective (\eg \cite{he2020momentum,grill2020bootstrap}) in order to train the feature encoder. 
Then the projection head is tuned to refine the representation so that it is better for matching. 
Like EQ, CL~\cite{cheng2021equivariant} also aims to make features distinct within the image. 
However, in contrast to EQ, the dense $D$ dimensional feature vectors from $\Psi(\mathbf{x})$ are linearly projected to a lower dimension $D'$ using a linear projection with weights $\mathbf{w} \in \mathbb{R}^{D\times D'}$. 
They use the same loss as Eqn.~\ref{eq:equivariance2}, but simply use $\mathbf{x}$ instead of $\mathbf{x}'$, \ie they do \emph{not} use a pair of augmented images. 

LEAD~\cite{karmali2022lead} also employs a two stage approach, but aims to maximize the similarity between feature correlation maps calculated using the original self-supervised features $\Psi(\mathbf{x})$ and the projected features $\Phi(\mathbf{x})$. 
The first term in their loss represents the probability that point $u$ from image $\mathbf{x}$ is matched with point $v$ in image $\mathbf{x}^\alpha$ using embeddings from the feature encoder $\Psi$. 
In the second term, embeddings are projected to a lower dimensional space using the combined encoder and projection head,
\begin{equation}
    \label{eq:lead}
        \mathcal{L}_{\mathit{lead}} = \frac{1}{|\Omega|^2} \sum_{u \in \Omega}\sum_{v \in \Omega}~-p(v | u; \Psi, \mathbf{x}, \mathbf{x}^\alpha, \tau) \log{p(v | u; \Phi, \mathbf{x}, \mathbf{x}^\alpha, \tau)}.
\end{equation}
\noindent LEAD uses `real' image pairs, as opposed to augmented images, \ie $\mathbf{x}^\alpha$ is not a synthetically augmented version of $\mathbf{x}$, but instead it is an auxiliary real image depicting the same object class. 
This is possible as their formulation does not require any ground truth correspondence during training. 
In essence, LEAD implements a form of learned dimensionality reduction, which can be effective if the pre-trained features already contain useful information for matching. 

EQ and DVE were originally designed such that their embedding network $\Phi$ was trained in an end-to-end manner, while CL and LEAD separately trained the encoder network $\Psi$, followed by the learned projection function $\rho$. 
Existing methods often use different network architectures for the encoder and decoder which makes it challenging to compare the objective functions directly. 
To fairly evaluate these approaches, in our experiments we use frozen pre-trained networks as the encoder $\Psi$, and train a separate linear projection head $\rho$, \ie $\Phi(\mathbf{x}) = \rho(\Psi(\mathbf{x}))$, for each of the losses.

\subsection{Unsupervised Asymmetric Correspondence Loss}

The LEAD objective aims to preserve distances between features before and after they have been projected into a lower dimensional feature space. 
Given two points, $u$ and $v$, from different images, the loss term effectively tries to enforce $f(\Psi_{u}(\mathbf{x}), \Psi_{v}(\mathbf{x}^\alpha))$ and $f(\Phi_{u}(\mathbf{x}), \Phi_{v}(\mathbf{x}^\alpha))$ to be as close as possible. 
The projection tries to maintain both what is similar and not similar between point pairs by preserving their distance. 
However, the structure of the embedding space does \emph{not} change after this projection step which means that performance is bounded by the quality of the features in the original feature space. 

We make a conceptually simple change to the LEAD objective in order to provide the flexibility to allow the model to change distances in the projected feature space. 
Unlike LEAD, instead of using the same temperature value in the softmax function for both feature spaces, we utilize a different temperature  when we calculate the similarity between point embeddings. 
Specially, we use a smaller temperature for the original feature space and a larger one for the projected feature space, \ie $\tau_{1} <\tau_{2}$, resulting in the following loss,
\begin{equation}
    \label{eq:asym}
        \mathcal{L}_{\mathit{asym}} = \frac{1}{|\Omega|^2} \sum_{u \in \Omega}\sum_{v \in \Omega}||~p(v | u; \Psi, \mathbf{x}, \mathbf{x}^\alpha, \tau_{1}) - p(v | u; \Phi, \mathbf{x}, \mathbf{x}^\alpha, \tau_{2})||.
\end{equation}

\noindent A smaller temperature makes the distance between closer points smaller and far away points larger. 
To match these same distance scores, the projection needs to make embeddings of closer points closer and vice versa. 
Moreover, the objective also preserves the order of distances of point pairs, \ie close points remain closer compared to further away ones.  
As a result, the projection needs to capture what is common between already matching point pairs in order to optimize the loss which leads to better embeddings for matching. 
While this is a relatively small change in the loss formulation, it results in a significant improvement in the performance. As we use different temperature parameters, we refer to our asymmetric projection loss as ASYM.
The other difference between ASYM and LEAD is that we make use of Euclidean distance instead of cross entropy as we found this to be more effective. We compare the impact of these design choices via detailed ablation experiments in our supplementary material.

\section{Evaluation Protocol}

\subsection{Evaluation Metrics}
There are two dominant approaches for benchmarking the performance of unsupervised correspondence estimation methods: (i) landmark regression and (ii) feature matching. 
For landmark regression, an additional supervised regression head is trained for each of the landmarks of interest (\eg the keypoints of a human face) on top of the representation learned by the correspondence network. %
For matching, one simply computes the distance in feature space to all the points in the second image for a given point of interest in a source image and then selects the closest match as the corresponding point.

We argue that matching is a better task for evaluating the power of learned feature embeddings as regression requires ground truth supervision to train the additional parameters. 
As matching uses raw feature embeddings it cannot incorporate biases from datasets, \eg exploiting the average locations of keypoints. %
While current literature tends to focus on regression evaluation, there are some exceptions to this. 
However, by and large, matching results are only presented for comparably easier datasets. 
For example, \cite{thewlis2019unsupervised,cheng2021equivariant,karmali2022lead} only present matching results on the MAFL dataset~\cite{MAFL}. 
MAFL contains cropped and aligned images of human faces, and current methods perform very well on it, with matching errors close to two pixels on average.
\subsubsection{Percentage of Correct Keypoints (PCK).}
Traditionally, matching performance is measured using the PCK metric. 
Given a set of ground truth keypoints $\mathcal{P}=\{\mathbf{p}_{m}\}_{m=1}^{M}$ and predictions $\mathcal{\hat{P}}=\{\hat{\mathbf{p}}_{m}\}_{m=1}^{M}$, $\mathrm{PCK}$ is calculated as $
    PCK(\mathcal{P}, \mathcal{\hat{P}}) = \frac{1}{M}\sum_{m=1}^{M} \mathbbm{1} [\norm{\hat{\mathbf{p}}_{m} - \mathbf{p}_{m}} \leq d ].$
Here, $d=\alpha \max{(W^b, H^b)}$  is a distance threshold, chosen as a proportion (\eg $\alpha = 0.1$ of the maximum side length) of the object bounding box (with width $W^b$ and height $H^b$) size. 
A prediction is counted as correct if it is inside of the target keypoint area. 
\subsubsection{Detailed Error Evaluation.}
\input{figs/errors}
Inspired by~\cite{ruggero2017benchmarking}, we define additional error metrics to analyze performance of different methods in more detail. 
A visual overview is illustrated in Fig.~\ref{fig:error_types}. 
If a point is matched with a point that not is close to any of the keypoints in the target image, we denote this error as a `miss'.
This error generally occurs when a point is matched with the image background: $E_{miss} = \mathbbm{1}[d < \min \{\norm{\hat{\mathbf{p}}_{m} - \mathbf{p} }  |  ~\mathbf{p} \in \mathcal{P} \}]$. If a prediction is in the correct vicinity, but outside of the defined distance threshold, we denote this a `jitter', $E_{jitter} = \mathbbm{1}[d < \norm{\hat{\mathbf{p}}_{m} - \mathbf{p}_{m}} < 2d]$. The last error type is a `swap' which occurs when a point matches in an area that is closer to a different keypoint, $E_{swap} = \mathbbm{1}[\delta \neq \norm{\hat{\mathbf{p}}_{m} - \mathbf{p}_{m}}  \wedge~ d > \delta]$, where $\delta = \min \{\norm{\hat{\mathbf{p}}_{m} - \mathbf{p}} | ~\mathbf{p} \in P \}$.

The miss and jitter errors are also counted as incorrect by the PCK metric, but swaps may still be counted as correct. 
For instance, a prediction which is in the middle of a pair of eyes might still be counted as correct according to PCK even if it closer to the wrong eye since it could be still within the distance threshold. 
As our goal is to estimate semantic correspondence, we should aim to match with the \emph{correct} semantic part. 
As a result, we propose a new version of PCK which penalizes these swaps. 
Under this metric, to make a correct prediction, a point needs to both match close to the corresponding keypoint \emph{and} the closest keypoint should be the same semantic keypoint, 
\begin{align}
    PCK^{\dag}(\mathcal{P}, \mathcal{\hat{P}}) = \frac{1}{M}\sum_{m=1}^{M} \mathbbm{1} [\norm{\hat{\mathbf{p}}_{m} - \mathbf{p}_{m}} \leq d ~\wedge~ \delta = \norm{\hat{\mathbf{p}}_{m} - \mathbf{p}_{m}}].
\end{align}
\subsection{Evaluation Datasets}
In order to evaluate semantic correspondence performance we perform experiments on five different datasets: AFLW~\cite{AFLW}, Spair-71k~\cite{min2019spair}, CUB-200-2011 (CUB)~\cite{CUB}, Stanford Dogs Extra (SDog)~\cite{khosla2011novel,biggs2020left}, and Awa-Pose~\cite{xian2018zero,awapose}. 
These datasets were chosen as they span a range of object category types (\eg from man made to  natural world classes) and exhibit different levels of difficultly (\eg from topologically simply human faces to deformable animals). 
AFLW~\cite{AFLW} contains images of human faces with various backgrounds from different view points. 
However, due to structured nature of faces, the visual difference between images are limited and thus the task is relatively easy compared to the other datasets. 
SDog~\cite{khosla2011novel,biggs2020left} and CUB~\cite{CUB} contain images of fine-grained visual categories (dogs and birds respectively) and include highly varying appearance, diverse backgrounds, and non-rigid poses which results in a challenging matching task. 
Awa-Pose~\cite{xian2018zero,awapose} contains images from 35 different animal species and allows us to asses inter-class correspondence as the keypoints are shared across the species. 
SPair-71k~\cite{min2019spair} contains scenes with multiple man made objects present with complex backgrounds, but the pairs come from same class and the size of the datasets is relatively small. 
An overview can be found in Table~\ref{table:datasets}. 

Only the annotations in SPair-71K were explicitly collected with a focus on semantic correspondence evaluation. %
For the other datasets there are no pre-defined image pairs or standardized correspondence evaluation splits. 
In the existing literature random image pairs are selected that makes direct comparison between alternative methods challenging~\cite{zhao2021multi,li2020correspondence,choy2016universal}. 
As the keypoint annotations are semantically consistent across instances in these datasets, we create splits for each dataset, where random image pairs are selected from test splits of the datasets. 
We will publish these splits in order to aid future evaluation. 
\input{tables/datasets}
\subsection{Implementation Details} 
We perform experiments with two different types of backbones models for our feature encoder $\Psi$. 
For the CNN, unless otherwise specified, we extract features from images resized to $384\PLH384$, and use the $1024$ dimensional features from the conv3 layer of a ResNet-50~\cite{he2016deep}. 
For the Transformer, $8\PLH8$ patches from $224\PLH224$ images with stride 8 are used as input (similar to \cite{amir2021deep}) and we extract $736$ dimensional features from 9th layer. 
We also investigate supervised and self-supervised trained backbones. 
The supervised and self-supervised CNNs are from \cite{he2016deep} and \cite{chen2021mocov3} and the Transformer models are from \cite{dosovitskiy2020image} and \cite{caron2021emerging} respectively. 
Unless stated otherwise, we report results using the standard PCK metric with $\alpha = 0.1$ for direct comparison to other methods. We set the temperature $\tau_1$ to 0.2 and $\tau_2$ to 0.4 for ASYM. We provide an evaluation of different temperature values and additional implementation details in the supplementary material.

\section{Experiments}
In our experiments, we attempt to answer the following questions: i) how well do current unsupervised correspondence methods perform on challenging datasets, ii) how does the choice of backbone architecture and pre-training objective impact performance, iii) how does the pre-training data source impact performance, iv) how does the data source used for finetuning the correspondence model impact performance, and finally, v) what are the current source of errors, and thus what needs to be done to close the gap between current state-of-the art supervised and unsupervised methods.

\subsection{Impact of Unsupervised Correspondence Objective}
\label{ssec:exp_projs}

\input{tables/projection}

To evaluate the unsupervised correspondence methods outlined in  Sec.~\ref{sec:semantic_correspondence}, in Table~\ref{table:projection} we train a linear projection head $\rho$ on top of the embeddings from a frozen pre-trained backbone $\Psi$. 
Additional baselines are also presented, including: pre-trained features directly from the backbone models with no projection (None), Non-Negative Matrix Factorization (NMF), Principal Component Analysis (PCA), projection using a Random weight matrix, and Supervised projection where we optimize the objective in Eqn.~\ref{eq:equivariance2} using ground truth keypoint pairs. 
We explore CNNs and Transformers as backbones that are pre-trained either in a supervised or self-supervised fashion. %

Overall, our proposed ASYM approach obtains better scores than other unsupervised methods on all datasets, independent of the choice of backbone or pre-training method, with the exception of the AFLW face dataset. 
Compared to LEAD, our proposed adaptation improves performance on datasets where the visual diversity is high (\ie non-face datasets).
EQ and DVE perform poorly on the datasets where the visual appearance is high across instances, but it is worth noting that these methods were originally designed for the end-to-end trained setting.  
CL obtains good performance in some cases and is the best on AFLW. 
However, our ASYM method is still consistently strong. 
Perhaps somewhat surprisingly, PCA based projection performs better than most of the baselines, while NMF did not perform well. 
PCA's performance can be partially explained by the strength of the original features (\ie None). 
Although the performance of unsupervised methods differs across different backbones, the relative ordering stays the same -- Sup, ASYM, CL, PCA, NONE, LEAD, NMF, EQ, and DVE.

\subsection{Impact of Backbone Model and Pre-training Objective}
\label{ssec:exp_models}

While \cite{cho2021cats} claims that the choice of CNNs or Transformers as the backbone model does not affect the performance, recently \cite{amir2021deep} presented impressive correspondence results using a Transformer-based model. 
In order to explore further, we compared features from models pre-trained on Imagenet with either supervised (Sup.) or unsupervised (Unsup.) objectives. 

When a projection layer is trained with keypoint supervision, the performance difference between architectures diminishes, as can be observed by comparing the supervised baseline to original embeddings (None) in Table~\ref{table:projection}. 
However, when the projection layer is trained using no supervision, the best results are obtained in the cases where the initial embeddings were the best on a given dataset. 
For instance, the unsupervised pre-trained Transformer obtains the best results with no projection on the SDog and Awa datasets compared to other backbone models. Training the unsupervised methods from these embeddings also results in the best performance compared to other pre-trained backbones. 
In summary, if keypoint supervision is available, the choice of backbone does not significantly impact the end result. However, in the unsupervised case, starting with good performing embeddings is important. Furthermore, the pre-training strategy does not affect the performance of CNNs, while unsupervised Transformers generally performs better than supervised one (see Table~\ref{table:projection}).

\subsection{Impact of Pre-training Dataset}
\label{ssec:exp_pre_data}

\begin{figure}[t]
\centering 
\includegraphics[trim={5 20 10 15},clip,width=0.90\textwidth]{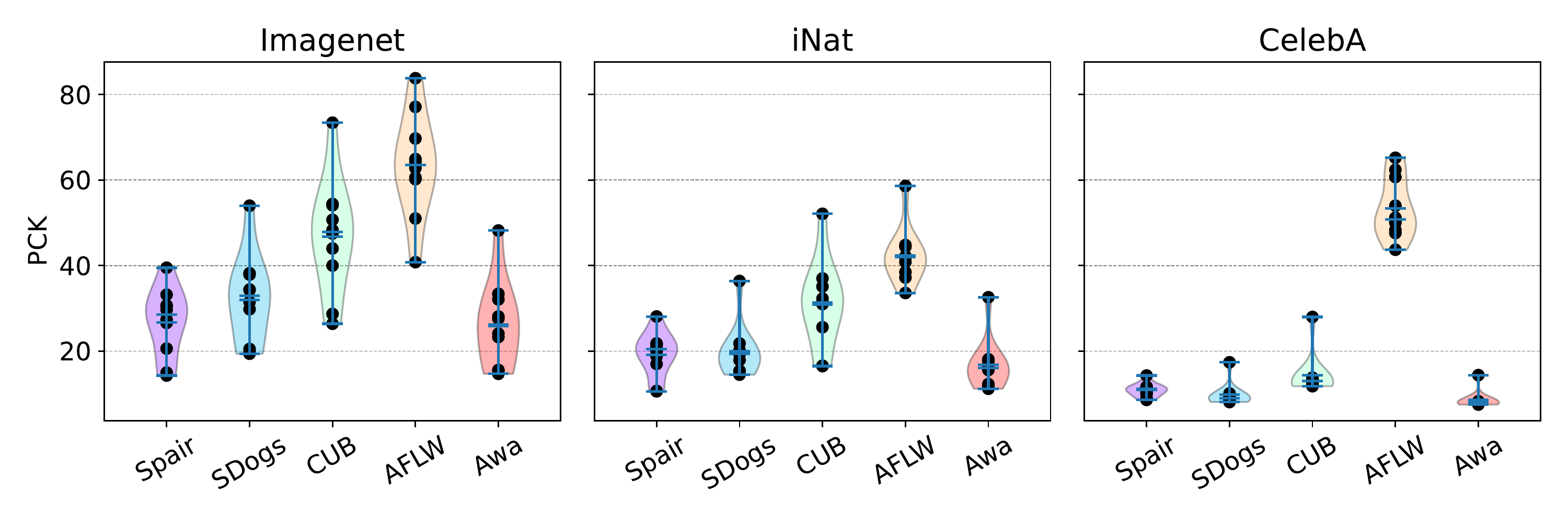}
\caption{Impact of different pre-training datasets used to train a CNN feature encoder using self-supervised training. 
For each of the three datasets we report the performance of different methods shown as individual dots. %
} 
\label{fig:diff_data_source}
\end{figure}

Here we explore the impact of the pre-training data source used to train the feature encoder. 
We train correspondence losses using embeddings from a CNN trained via contrastive self-supervision on either  Imagenet~\cite{russakovsky2015imagenet} (various categories), iNat2021~\cite{van2021benchmarking} (natural world categories), or Celeb-A~\cite{liu2015deep} (human faces). 
Specifically, we use MoCov3 from \cite{chen2021mocov3} for Imagenet, MoCov2~\cite{mocov2} for iNat from \cite{van2021benchmarking},  and MoCov2 from \cite{cheng2021equivariant} for CelebA. 
These results are presented in Fig.~\ref{fig:diff_data_source}. 

It is clear that the choice of pre-training data has an impact on all unsupervised methods, with Imagenet outperforming other sources. 
The CelebA model performs poorly on all tasks with the exception of AFLW, as the features likely only contain information about faces. 
iNat2021 does not contain any man-made objects or dog categories, and as a result, models trained on it perform worse on SDog and Spair. 
While iNat2021 contains many bird images, it contains an order of magnitude less mammals making it less effective on Awa-Pose. 

\subsection{Impact of Finetuning Correspondence Dataset}
\label{ssec:exp_proj_data}
\begin{figure}[t]
        \centering
        \includegraphics[trim={10 18 15 18},clip,width=0.92\textwidth]{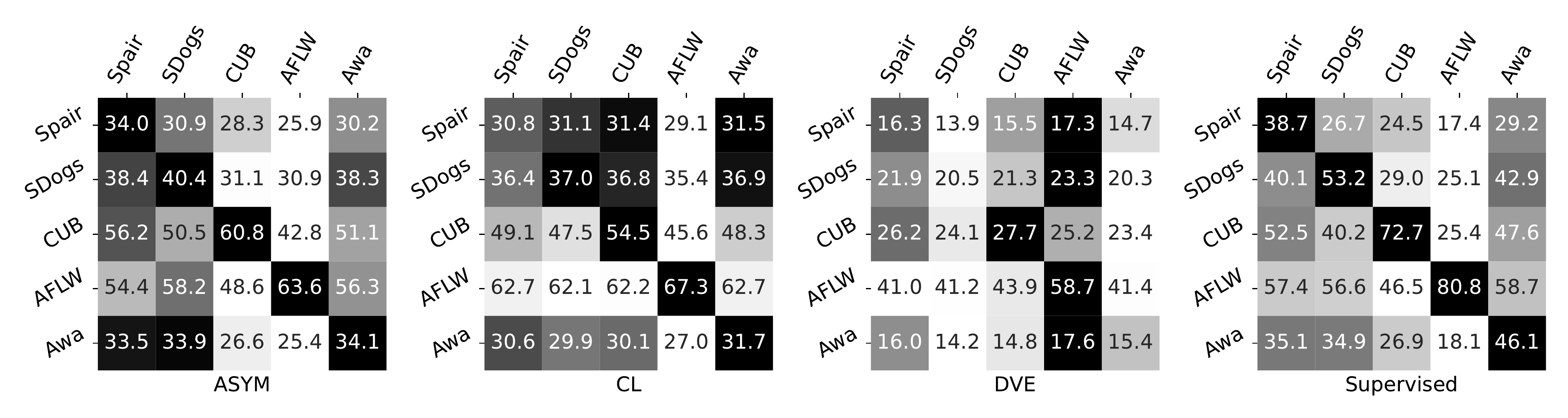}
         \caption{Cross dataset evaluation results. Each row represents the test source data, and each column is the dataset that a given correspondence loss  is trained on.
         Note that the colormaps are row normalized. 
         These results use the same initial encoder as the `Sup. pre-trained - CNN' results in Table~\ref{table:projection}.
         }
         \label{fig:cross_evalution}
\end{figure}

Next we explore how transferable are the embeddings trained on one dataset and evaluated on another. 
For instance, what happens if the linear projection is trained on dog images and then tested on birds, or in an extreme case, trained on human faces and tested on animal categories.
The correspondence losses are trained on top of the sup. CNN from Table~\ref{table:projection}. 
The results are outlined in Fig.~\ref{fig:cross_evalution}.

The generalization performance across other datasets is poor for supervised losses compared to the unsupervised ones. 
The performance drop is largest for models trained on faces, but when training on other data and tested on faces, the performance does not drop significantly. 
Models trained on Spair-71k generally perform reasonably well on other datasets.
\subsection{Detailed Error Analysis}
\label{ssec:exp_error_analysis}
Here we break down the different error types in order to better understand where the different methods fail and thus require improvement. 
We compare unsupervised correspondence losses and supervised projection to the current best performing methods CATs~\cite{cho2021cats}, CHM~\cite{min2021convolutional}, and MMNet~\cite{zhao2021multi} on Spair-71K. 
Results are presented in Table~\ref{table:error_analysis}. 

For the supervised methods, MMNet has significantly lower miss errors compared to all other methods, although it results in a lot of swaps. As this method combines correlation maps from different layers, it is able to capture more global context, which helps reduce misses. 
However, while CATs and CHM produce more misses compared to MMNet, swaps are reduced, as they use more sophisticated aggregation methods (6D convolution and attention) to resolve ambiguities during matching. 
Moreover, as these two lines of work complement each other in the error types, they could potentially be combined to obtain better results. 

For the unsupervised methods, we see that the most common error type is miss across all methods. 
While ASYM reduces misses compared to other unsupervised methods, it is not as good as the supervised approaches. As swaps are instances where a match has occurred, but to the wrong keypoint, methods with high number of misses will not have many swaps by definition. ASYM results in fewer misses, which is desirable, but this increases the chance that swaps can occur.
The `Supervised' baseline reduces misses, but compared to the more sophisticated supervised approaches it generates more swaps. 
We argue that while more supervision might help to reduce misses, in order to reduce swaps, better matching mechanisms are needed, as in~\cite{cho2021cats,min2021convolutional}.

Finally, we can see that our $\text{PCK}^{\dag}$ metric is reduced by $\sim20\%$ compared to the original PCK metric in all cases. 
This indicates that in one in five cases, the source point matches an area closer to another keypoint instead of the correct corresponding point. 
For some applications these errors might not affect the end performance drastically, while for others,  this disparity could be significant. 
We provide additional analysis and $\text{PCK}^{\dag}$ scores for other datasets in the supplementary material.

\input{tables/error_analysis}

\subsection{Discussion and Limitations}
Our exhaustive experiments show that evaluating with varied challenging datasets is crucial in order to see the benefits of current methods as human face data results (\eg AFLW) alone can be misleading (Table~\ref{table:projection}). 
While unsupervised performance may not yet be at the level of fully supervised baselines, they are not far off but have the benefit of generalizing better across datasets  (Fig.~\ref{fig:cross_evalution}). 
Current performance metrics (\ie PCK) do not penalize all error types and thus result in overly optimistic performance (Table~\ref{table:error_analysis}). 
The choice of pre-training can have a big impact, but in most instances Imagenet pre-training is superior (Fig.~\ref{fig:diff_data_source}). 

It is not feasible to control all hyper-parameter values as the space too large. 
As a result, to ensure fair and controlled comparisons, we adopted a two stage pipeline, with frozen backbone models, as advocated in recent start-of-the-art work~\cite{cheng2021equivariant}. 
We justified the important design choices and provide additional experiments in the supplementary material. 
Finally, the keypoints used for evaluating correspondence are derived from object landmarks which are detectable and salient by design. 
In future work, it would be interesting to use additional annotations from other object parts which are not necessarily easily annotated but still have semantically meaningful correspondences across instances. 

\section{Conclusion}
We presented a thorough evaluation of existing unsupervised methods for semantic correspondence estimation and presented a new approach that consistently outperforms existing methods. 
We showed that while matching performance on human face data is strong, there is still a way to go on more challenging datasets. 
Our analysis sheds light on some of the reasons for failure as well as providing some further insight into the role of data, models, and losses which we hope will enable others to make further progress on this important task. 
\small{
\par \noindent \textbf{Acknowledgements:} Thanks to Hakan Bilen and Omiros Pantazis for their valuable feedback.
This work was in part supported by the Turing 2.0 `Enabling Advanced Autonomy' project funded by the EPSRC and the Alan Turing Institute. 
}

\bibliographystyle{splncs04}
\bibliography{main}

\clearpage
\appendix
\setcounter{table}{0}
\renewcommand{\thetable}{A\arabic{table}}
\setcounter{figure}{0}
\renewcommand{\thefigure}{A\arabic{figure}}

\input{supp_content}

\end{document}

%% file: figs/errors.tex
\begin{figure}[t]
     \centering
     \begin{subfigure}[b]{0.19\textwidth}
        \centering
         \includegraphics[height=1.8cm, trim=0 20 0 35, clip]{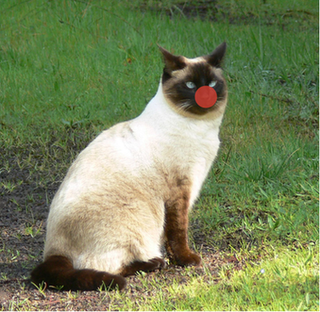}
         \caption{Source}
         \label{fig:source_img}
     \end{subfigure}
     \hfill
     \begin{subfigure}[b]{0.19\textwidth}
     \centering
         \includegraphics[height=1.8cm]{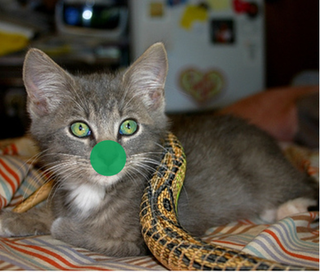}
         \caption{Match}
         \label{fig:pck_match}
     \end{subfigure}
     \hfill
    \begin{subfigure}[b]{0.19\textwidth}
         \centering
         \includegraphics[height=1.8cm]{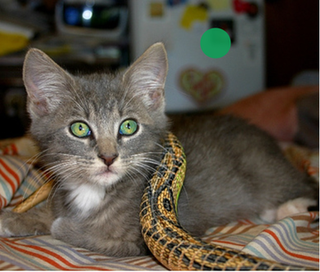}
         \caption{Miss}
         \label{fig:err_miss}
     \end{subfigure}
     \hfill
    \begin{subfigure}[b]{0.19\textwidth}
         \centering
         \includegraphics[height=1.8cm]{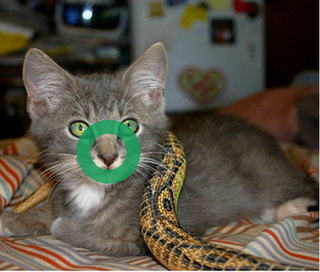}
         \caption{Jitter}
         \label{fig:err_jitter}
     \end{subfigure}
     \hfill
    \begin{subfigure}[b]{0.19\textwidth}
         \centering
         \includegraphics[height=1.8cm]{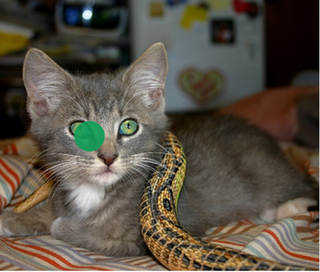}
         \caption{Swap}
         \label{fig:err_swap}
     \end{subfigure}
        \caption{For the keypoint denoted in red in the source image (a), we see the correct match in (b). If the point matches with the background it is a miss (c), if it is close to the correct location it is a jitter (d). If the match is in the correct vicinity but closer to another semantic part, it is a swap error (e).}
        \label{fig:error_types}
\end{figure}    

%% file: tables/datasets.tex
\begin{table}[t]
\footnotesize
\centering
\caption{Summary of the different datasets that we use for evaluating semantic correspondence performance. 
We also report the metadata that is provided with each dataset: KP (keypoints/landmarks) and Bbox (bounding boxes).
With the exception of Spair-71k, there is no pre-defined evaluation pairs for the datasets.}
\resizebox{1.0\textwidth}{!}{
\begin{tabular}{lcccrr}
\toprule
Dataset Name & ~~\# Images~~ & ~~\# Pairs~~ & ~~\# Classes~~ & ~~Annotations & ~~ Matching Diversity\\ \midrule
SPair-71k~\cite{min2019spair} & 2k & 70k & 18 & KP (3-30), Bbox & Med \\
Stanford Dogs (SDog)~\cite{biggs2020left} & 10k & 10k & 120 & KP (24), Bbox & Med\\
CUB-200-2011 (CUB)~\cite{CUB} & 11k & 10k &  200 & KP(15), Bbox & Med\\
AFLW~\cite{AFLW} & 13k & 10k & - & KP(5) & Low\\
Awa-Pose~\cite{awapose} & 10k & 10k & 36 & KP (30-40), Bbox & High \\ 
\bottomrule
\end{tabular}
}
\label{table:datasets}
\end{table}

%% file: tables/projection.tex
\begin{table}[t]
\caption{Comparison of different unsupervised semantic correspondence  methods. 
Here we vary the backbone models and pre-training strategies. 
The unsupervised correspondence methods are trained on the respective evaluation datasets.} 
\subfloat[$\Psi$ = Sup. pre-trained - CNN]{
\resizebox{0.48\linewidth}{!}{
\begin{tabular}{lccccc}
\toprule
Projection($\rho$)~ &  Spair-71K & SDogs & CUB & AFLW & Awa \\ \midrule
None & 31.8 & 34.9 & 51.3 & 57.4 & 28.8  \\
NMF & 27.4 & 33.9 & 49.6 & 53.6 & 28.0 \\
PCA & 32.2 & 35.5 & 53.1 & 57.8 & 29.7 \\
Random & 26.9 & 30.5 & 43.1 & 54.9 & 23.4 \\ \midrule
Supervised & 38.7 & 53.2 & 72.7 & 80.8 & 46.1 \\ \midrule
EQ\cite{thewlis2017unsupervisedneurips} & 16.4 & 21.2 & 28.1 & 48.5 & 15.6  \\
DVE\cite{thewlis2019unsupervised} & 16.3 & 20.5 & 27.7 & 58.7 & 15.4  \\
CL\cite{cheng2021equivariant} & 30.8 & 37.0 & 54.5 & \textbf{67.3} & 31.7  \\
LEAD\cite{karmali2022lead} & 31.7 & 35.1 & 51.5 & 58.0 & 29.1  \\ \midrule
ASYM (Ours) & \textbf{34.0} & \textbf{40.4} & \textbf{60.8} & 63.6 & \textbf{34.1}  \\
\bottomrule
\end{tabular}
}
}
\subfloat[$\Psi$ = Unsup. pre-trained - CNN]{
\resizebox{0.48\linewidth}{!}{
\begin{tabular}{lccccc}
\toprule
Projection($\rho$)~  &  Spair-71K & SDogs & CUB & AFLW & Awa \\ \midrule
None & 30.7 & 34.3 & 47.5 & 64.3 & 27.6  \\
NMF  & 20.6 & 19.9 & 44.0 & 40.8 & 15.6  \\
PCA  & 27.4 & 29.8 & 50.7 & 51.0 & 24.1  \\ 
Random  & 26.6 & 31.5 & 40.0 & 60.2 & 23.3  \\\midrule
Supervised  & 39.5 & 54.0 & 73.4 & 83.8 & 48.2  \\ \midrule
EQ\cite{thewlis2017unsupervisedneurips} & 14.3 & 20.5 & 26.4 & 62.8 & 15.5  \\
DVE\cite{thewlis2019unsupervised} &  15.0 & 19.4 & 28.7 & 60.6 & 14.7  \\
CL\cite{cheng2021equivariant} & 29.7 & 37.9 & 54.1 & \textbf{77.1} & \textbf{33.4} \\
LEAD\cite{karmali2022lead}  & 30.5 & 34.4 & 48.3 & 64.9 & 28.1  \\\midrule
ASYM (Ours) & \textbf{33.2} &\textbf{38.2} & \textbf{54.4} & 69.7 & 32.1  \\
\bottomrule
\end{tabular}
}
}

\subfloat[$\Psi$ = Sup. pre-trained - Transformer]{
\resizebox{0.48\linewidth}{!}{
\begin{tabular}{lccccc}
\toprule
Projection($\rho$)~  &  Spair-71K & SDogs & CUB & AFLW & Awa \\ \midrule
None & \textbf{33.5} & 38.0 & 66.3 & 54.1 & 34.1  \\  
NMF  & 23.3 & 29.2 & 55.5 & 51.5 & 24.7  \\
PCA  & 33.0 & 38.1 & 66.4 & 53.9 & 34.1  \\
Random  & 31.9 & 36.9 & 63.3 & 52.9 & 31.8  \\  \midrule
Supervised  & 38.5 & 48.2 & 78.2 & 70.5 & 47.9  \\ \midrule
EQ\cite{thewlis2017unsupervisedneurips} & 15.5 & 15.9 & 24.0 & 60.2 & 11.7  \\
DVE\cite{thewlis2019unsupervised} & 15.4 & 17.5 & 23.8 & 55.6 & 11.8  \\
CL\cite{cheng2021equivariant} & 30.5 & 35.8 & 67.1 & \textbf{68.4} & 31.0  \\
LEAD\cite{karmali2022lead}  & 32.7 & 37.6 & 65.8 & 53.8 & 33.9  \\ \midrule
ASYM (Ours) & 33.2 & \textbf{41.7} & \textbf{72.2} & 54.2 & \textbf{38.5}  \\
\bottomrule
\end{tabular}
}
}
\subfloat[$\Psi$ = Unsup. pre-trained - Transformer]{
\resizebox{0.48\linewidth}{!}{
\begin{tabular}{lccccc}
\toprule
Projection($\rho$)~  &  Spair-71K & SDogs & CUB & AFLW & Awa \\ \midrule
None & \textbf{34.1} & 42.7 & 61.0 & 64.2 & 36.1  \\  
NMF  & 26.3 & 39.0 & 51.9 & 61.0 & 32.9  \\
PCA  & 34.0 & 42.7 & 61.0 & 64.2 & 36.1  \\ 
Random  & 32.3 & 42.1 & 59.6 & 61.9 & 34.6  \\ \midrule
Supervised  & 38.1 & 52.7 & 72.9 & 92.0 & 47.4  \\ \midrule
EQ\cite{thewlis2017unsupervisedneurips} & 9.0 & 12.5 & 15.0 & 62.5 & 8.8  \\
DVE\cite{thewlis2019unsupervised} & 8.5 & 13.1 & 14.1 & 60.6 & 9.0  \\
CL\cite{cheng2021equivariant} & 25.8 & 32.3 & 54.1 & \textbf{81.8} & 25.0  \\
LEAD\cite{karmali2022lead}  & 33.6 & 42.5 & 60.8 & 64.2 & 35.8  \\ \midrule
ASYM (Ours) & 32.9 & \textbf{45.2} & \textbf{65.2} & 65.9 & \textbf{39.9} \\
\bottomrule
\end{tabular}
}
}
\label{table:projection}
\end{table}

%% file: tables/error_analysis.tex
\begin{table}[t]
\caption{Detailed error types for both unsupervised and supervised correspondence losses on Spair using two different distance thresholds. FT indicates if the backbone was finetuned with keypoint supervision. Our baselines use the `Sup. pre-trained - CNN' encoder from Table~\ref{table:projection}, in other cases we use the public models by the authors. 
All models use a ResNet backbone, except MMNet-FCN\cite{zhao2021multi}. 
}
\centering
\subfloat[$\alpha = 0.1$]{
\resizebox{0.48\linewidth}{!}{
\begin{tabular}{lllccccc}
\toprule
 ~ & ~FT~ & Method & ~Miss$\downarrow$~ & ~Jitter$\downarrow$~ & ~Swap$\downarrow$~ & ~PCK$\uparrow$~ & ~$\text{PCK}^{\dag}$$\uparrow$\\ \midrule
\parbox[t]{2mm}{\multirow{5}{*}{\rotatebox[origin=c]{90}{Unsup.}}} && CL & 51.5 & 13.7 & 24.3 & 30.8 & 24.2 \\
& & EQ & 68.3 & 15.0 & 18.9 & 16.4 & 12.8 \\
& & DVE & 67.9 & 14.9 & 19.7 & 16.3 & 12.4 \\
& & LEAD & 47.1 & 13.6 & 27.4 & 31.7 & 25.4 \\
& & ASYM & 44.1 & 13.2 & 28.6 & 34.0 & 27.2 \\
\midrule
\parbox[t]{2mm}{\multirow{5}{*}{\rotatebox[origin=c]{90}{Sup.}}} & & Supervised &  40.2 & 14.9 & 29.4 & 38.7 & 30.4  \\ 
& & CATs~\cite{cho2021cats} & 46.3 & 21.0 & 21.9 & 42.4 &  31.7 \\ 
 & ~$\checkmark$& CATs~\cite{cho2021cats} & 40.1 & 19.1 & 20.3 & 49.9 &  39.6 \\
 & ~$\checkmark$ & CHM \cite{min2021convolutional} & 40.3 & 18.2 & 23.8 & 44.2 & 35.8  \\
 & ~$\checkmark$ & MMNet-FCN\cite{zhao2021multi} & 28.5 & 14.7 & 28.8 & 52.2 & 42.6 \\ 
\bottomrule
\end{tabular}
}
}
\subfloat[$\alpha = 0.05$ ]{
\resizebox{0.48\linewidth}{!}{
\begin{tabular}{lllccccc}
\toprule
 ~ & ~FT~ & Method & ~Miss$\downarrow$~ & ~Jitter$\downarrow$~ & ~Swap$\downarrow$~ & ~PCK$\uparrow$~ & ~$\text{PCK}^{\dag}$$\uparrow$\\ \midrule
\parbox[t]{2mm}{\multirow{5}{*}{\rotatebox[origin=c]{90}{Unsup.}}} & & CL & 71.5 & 13.2 & 12.9 & 17.7 & 15.6 \\
& & EQ & 85.1 & 8.8 & 8.0 & 7.6 & 6.9 \\
& & DVE & 85.3 & 9.0 & 8.3 & 7.3 & 6.5 \\
& & LEAD & 66.9 & 12.4 & 15.9 & 19.3 & 17.3 \\
&&ASYM & 63.3 & 12.6 & 17.5 & 21.5 & 19.2 \\
\midrule
\parbox[t]{2mm}{\multirow{5}{*}{\rotatebox[origin=c]{90}{Sup.}}}& & Supervised &  61.1 & 14.6 & 17.6 & 24.1 & 21.3  \\ 
&&CATs~\cite{cho2021cats} &  71.0 & 20.7 & 10.8 & 21.6 & 18.1 \\ 
&~$\checkmark$& CATs~\cite{cho2021cats}  & 64.8 & 22.2 & 10.7 & 27.7 & 24.4  \\
&~$\checkmark$ & CHM~\cite{min2021convolutional} & 64.5 & 18.7 & 12.4 & 25.6 & 23.1  \\
&~$\checkmark$& MMNet-FCN\cite{zhao2021multi} & 51.7 & 19.0 & 18.1 & 33.3 & 30.2 \\ 
\bottomrule
\end{tabular}
}
}
\label{table:error_analysis}
\end{table}

%% file: supp_content.tex
\section{Additional Experiments and Implementation Details}
Here we evaluate some of the implementation choices made in the main paper and provide additional implementation details.

\subsection{Implementation Details} 
We perform experiments with two different types of backbones models for our feature encoder $\Psi$. 
For the CNN, unless otherwise specified, we extract features from images resized to $384\PLH384$, and use the $1024$ dimensional features from the conv3 layer of a ResNet-50~\cite{he2016deep}. 
We use a ResNet-50 trained on Imagenet~\cite{russakovsky2015imagenet} as our supervised baseline, and MoCov3~\cite{chen2021mocov3} as our unsupervised CNN. 
For the Transformer, $8\PLH8$ patches from $224\PLH224$ images with stride 8 are used as input (similar to \cite{amir2021deep}) and we extract $736$ dimensional features from 9th layer. 
We also investigate supervised and self-supervised trained backbones. 
The supervised and self-supervised CNNs are from \cite{he2016deep} and \cite{chen2021mocov3} and the Transformer models are from \cite{dosovitskiy2020image} and \cite{caron2021emerging} respectively. 
During training, we upsample feature maps to $64\PLH64$ via bilinear interpolation. 
For our projection head $\rho$, a single $1\PLH1$ 2D convolution is trained and the dimension of the features is reduced to $256$. 
During training, as in \cite{cheng2021equivariant}, we freeze the feature encoder $\Psi$. 
The projection head is trained for 50 epochs using Adam~\cite{kingma2014adam} optimizer with learning rate of $0.001$. 
Unless stated otherwise, we report results using the standard PCK metric with $\alpha = 0.1$ for direct comparison to other methods.
For EQ, DVE and LEAD we set the temperature $\tau$ to 0.05 and 0.14 for CL as in described in their papers, and set $\tau_1$ to 0.2 and $\tau_2$ to 0.4 for ASYM. 
We provide an evaluation of different temperature values in the supplementary material. 

\subsection{Impact of the Temperature Value}
In Table~\ref{table:temp_exp}, we explore the impact of the temperature for the different unsupervised losses. 
While the performance of LEAD, ASYM, and DVE do not change significantly with different temperature choices, the performance of CL is impacted drastically, \ie when using the recommended value of 0.14 from their paper, we obtain a PCK of 30.8 for Spair-71K in Table~2 in the main paper. 
As noted in the main paper, for EQ, DVE, and LEAD we
set the temperate $\tau$ to 0.05 and use 0.14 for CL based on the recommendations in the original papers. We use the same temperature values for all datasets.
\input{tables/temp_exp}

\subsection{Impact of Design Choices for ASYM}
As our new proposed ASYM loss is an adaptation of LEAD, here we present experiments ablating our design choices.  %
ASYM differs from LEAD in two respects: (i) ASYM uses different temperature values for the correlation maps for the original features and the projected features, and  (ii) ASYM uses a mean square error (MSE), as opposed to cross entropy (CE) which is used in LEAD. 
As can be seen in Table~\ref{table:mse_ces}, the MSE loss performs worse for LEAD while it improves performance of ASYM. 
However, the main difference in overall performance is not a result of the choice of penalty function (\ie MSE versus CE), but the usage of different temperature parameters. 
In Table~\ref{table:mse_ces} we can see that changing the temperature for LEAD has no significant impact on the final performance.  
\input{tables/mse_ce}

Due to changes in the formulation, the objectives that ASYM and LEAD optimize also differ. 
For a given pair of points and their similarity score, LEAD  reduces the dimensionality of the embeddings for these points while maintaining the same similarity scores as the input feature space. 
This is achieved by capturing both what is common and not common between the pair of points. Using higher or lower temperature values does not change the feature distances in the LEAD. 
However, in our ASYM objective, for a point pair which has a high similarity score, the projection needs to make these points even closer in order to match with the same similarity score from the input features as the projected embeddings use a higher temperature value. 
A visualization of the result of this can be observed in  Fig.~\ref{fig:lead_vs_asym}. 
As expected, for a given keypoint and a target image LEAD produces a very similar similarity map compared to the one  calculated with the original features. 
In contrast, ASYM produces a more `peaked' similarity map, since matching points from original features become closer in the new embedding space. 

We also compare how the similarity scores change after unsupervised projection. 
For a source keypoint, we calculate the cosine similarity scores for all pixel embeddings in the target image. If a point is within the threshold area of a target keypoint we refer to these points as `correct' matches, otherwise they are classed as `wrong' matches. 
We visualize the histogram of these scores for all datasets in Fig.~\ref{fig:hist_lead_vs_asym}. 
As can be seen from the distributions, LEAD results in histograms that are very similar to original input features (\ie None). 
However, ASYM reduces the overlap between the correct and wrong distributions. 
As expected, if the similarity scores for correct matches are not larger than wrong matches, ASYM cannot improve the embeddings significantly, as seen in the Awa dataset.  

\input{figs/lead_vs_asym/lead_vs_asym}
\input{figs/lead_vs_asym/hist_lead_asym}

\subsection{Impact of Encoder Feature Layer}
In Table~\ref{table:layers} we experiment with using features from different feature layers from a CNN (Resnet50~\cite{he2016deep}) trained using supervision on Imagenet. 
The third convolution layer performs best on all datasets, and so we use features from it in all of our experiments for CNNs. 
For Transformer backbones~\cite{dosovitskiy2020image,caron2021emerging}, we used the 9th layer as the initial features, as they were shown to perform best in~\cite{amir2021deep}.
\input{tables/feature_layer}

\subsection{Impact of Input Image Resolution}
\begin{figure}
    \centering
    \begin{subfigure}[b]{0.40\textwidth}
        \includegraphics[trim={5 10 5 10},clip,width=\textwidth]{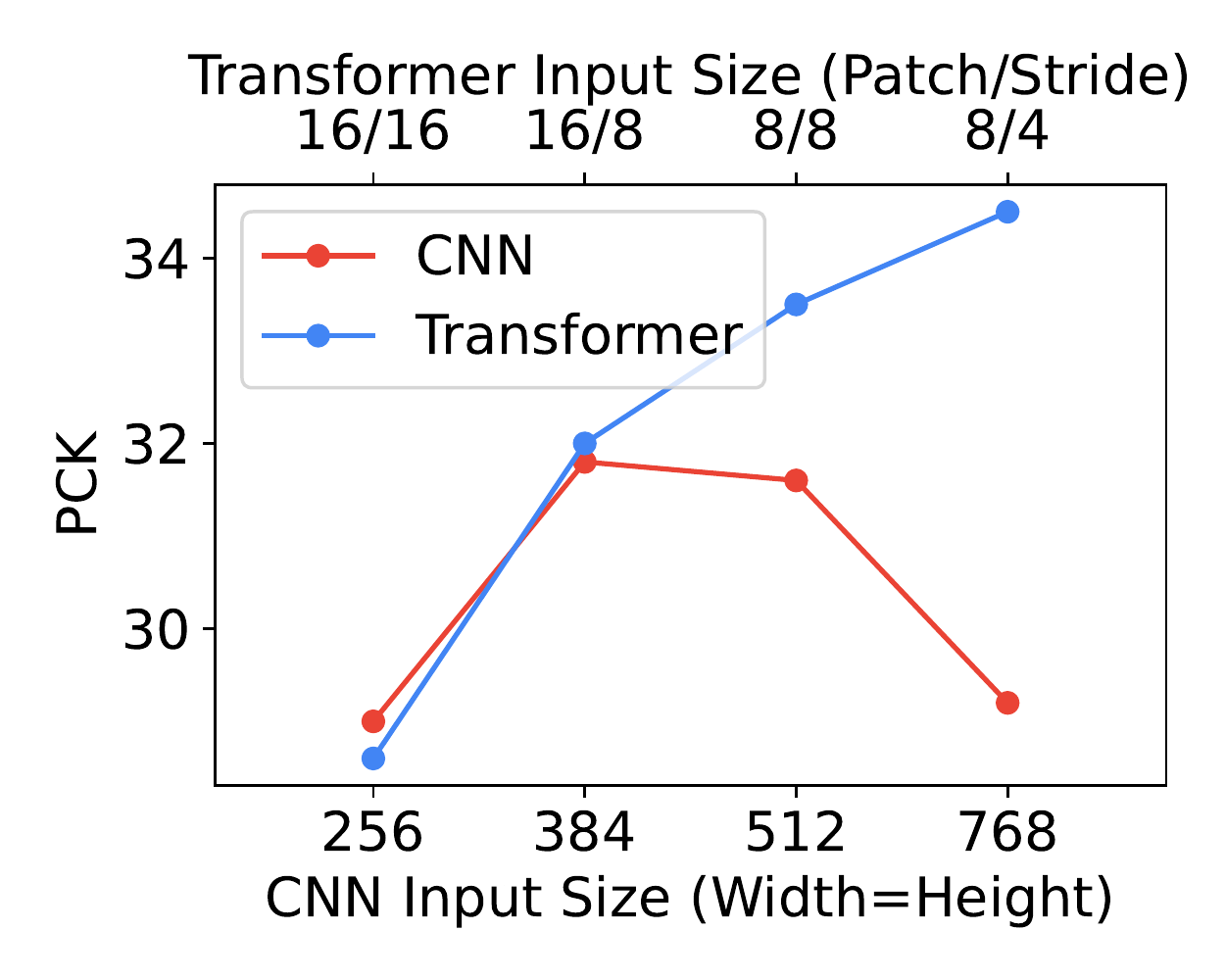} 
         \caption{Supervised Pre-training}
         \label{fig:sup_cnn_transformer}
     \end{subfigure}
    \begin{subfigure}[b]{0.40\textwidth}
        \includegraphics[trim={5 10 5 10},clip,width=\textwidth]{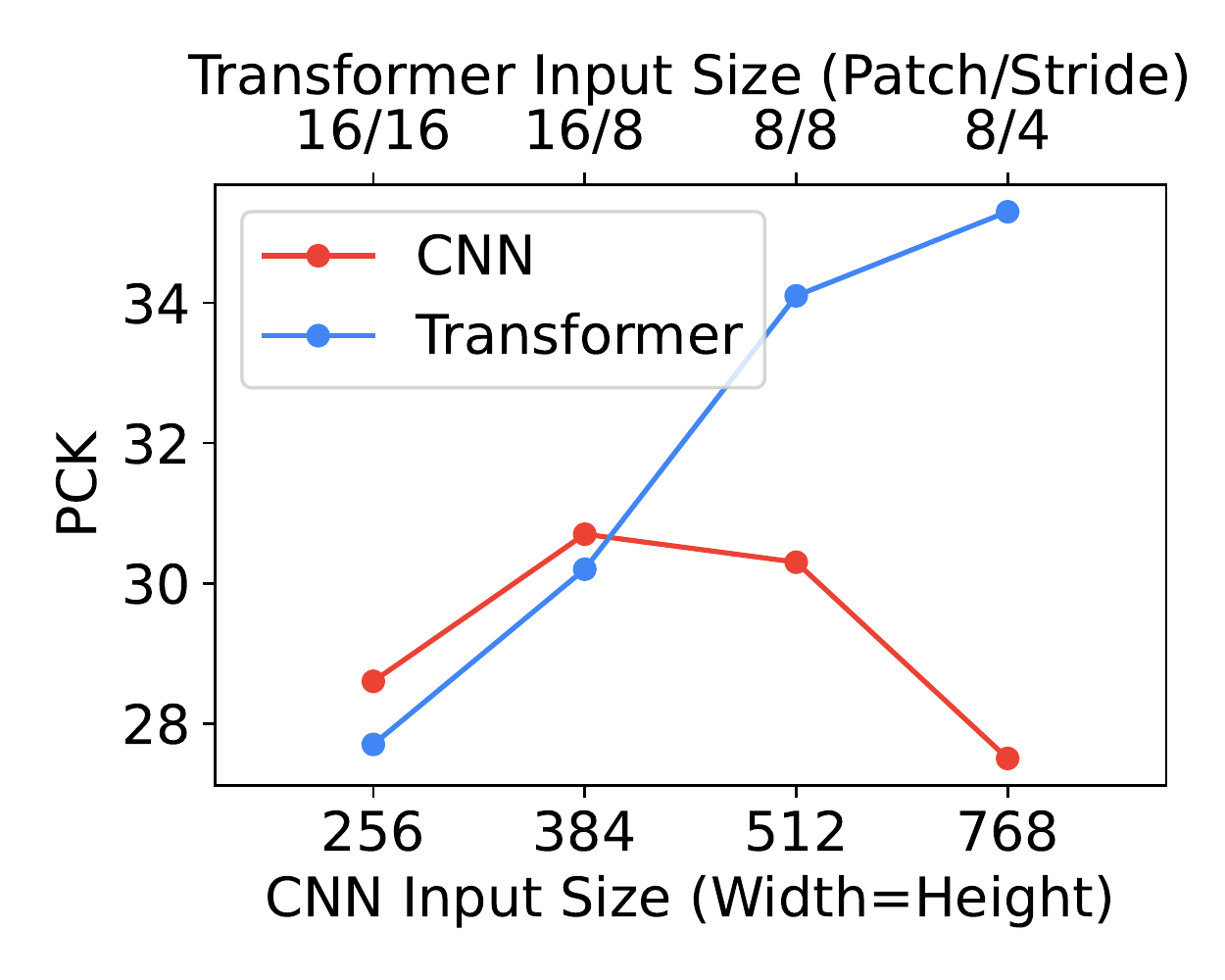} 
         \caption{Unsupervised Pre-training}
         \label{fig:unsup_cnn_transformer}
     \end{subfigure}
    \caption{Semantic correspondence performance of CNNs and Transformers with different input sizes on Spair-71K with no projection. Pre-trained features from models trained on Imagenet with (a) supervised  or (b) unsupervised  losses are used. Image resolution is fixed to 224x224 for the Transformers. Note that the effective resolution of feature maps from CNNs and Transformers are not comparable for each vertical position in the plots.}
\label{fig:cnn_vs_transformer}
\end{figure}

In Fig.~\ref{fig:cnn_vs_transformer}, we explore the impact of different input image resolutions, using pre-trained embeddings without any projection (\ie None), for CNN and Transformer backbones. 
We used CNNs are from \cite{he2016deep} and \cite{chen2021mocov3} as the supervised and unsupervised CNN, \cite{dosovitskiy2020image} and \cite{caron2021emerging} as the supervised and unsupervised Transformer. 
Transformers scale well as the number of tokens increases, while the performance of the CNNs saturates as the image resolution is increased. 
We argue that this is due to not-adaptive nature of the receptive field sizes of CNNs which may overfit to the trained image resolution. 
As CNNs best performed using an input resolution of 384x384, we use that resolution for in our experiments.
While 8x8 patches with stride 4 is the best performing version for transformers, due to computational constraints, we used 8x8 patches with stride 8 as the transformer input in our experiments. 

\section{Additional Results and Analysis}

Here we present additional results and more detailed analysis for each of the  datasets of interest.

\subsection{Detailed Error Analysis for Additional Datasets}
We present the detailed error analysis and report scores using our $\text{PCK}^{\dag}$ metric in Table~\ref{table:pckdag_alldatasets} for each dataset not shown in the main paper.  
Similar to the Spair-71k results from the main paper, the most common error type is `miss' among all datasets.
Our ASYM approach generally reduces misses compared to other unsupervised losses. 
With the exception of the AFLW dataset, there is a noticeable difference between $\text{PCK}^{\dag}$ and PCK scores. 
For AFLW, the keypoints that correspond to each other are well defined and far apart from each other as the faces are large. 
As a result, there are far fewer swaps, and so $\text{PCK}^{\dag}$ scores are close to their PCK counterparts. 
In contrast, for CUB, most of the points are distributed close to the head region of the birds which leads to a lot of swaps and a drop in scores for our new proposed metric. 
This highlights the importance of using a proper metric for evaluating the semantic correspondence task.
Matching a keypoint from the beak of a bird to the eye of another bird is not a correct semantic match, but with the current PCK metric it would be labeled as correct if it was within the distance threshold. 
\input{tables/pckdag_otherdatasets}

\subsection{Example Images and Qualitative Results}
Random instance pairs from each dataset are depicted in Fig.~\ref{fig:examples}. 
Spair-71K contains examples of different classes, spanning man-made objects to animal classes. StanfordDogs (SDogs) contains different breeds of dogs in challenging poses with varying appearance. 
CUB contains bird species. %
AFLW contains human faces which occupy most of the frame. 
Unlike CUB and SDogs which only contains images from one species, Awa includes different vertebrate animal categories which enables us to assess inter-category correspondence performance.
\input{figs/examples_from_data}

We also present some qualitative results for the different unsupervised losses, for all datasets, in Fig~\ref{fig:qual} and Fig~\ref{fig:qual2}. 
While ASYM generally improves the predictions compared to other unsupervised losses, it still lags behind supervised projection which makes use of ground truth matches for training. 
AFLW generally contains easy examples with a small percentage of background pixels and only minor changes in pose which makes the task easier. 
While the PCK scores for AFLW and CUB are close to each other,  as can be seen from qualitative results, this can be explained by how PCK evaluates matches which does not necessarily reflect the difficult of the dataset in some cases. 

\input{figs/qualitative}
\input{figs/qualitative2}
\input{figs/qualitative3}

\subsection{Visualizing Learned Feature Embeddings}

We present 2d t-SNE~\cite{van2008visualizing} visualizations of the keypoint embeddings for the AFLW, CUB, and SDogs datasets in Fig~\ref{fig:tsne}. Since Spair contains different classes wherein the keypoints are not semantically consistent across classes, we did not present t-SNE visualization of Spair. 
Moreover, the Awa dataset contains more than 30 keypoints which makes visualizing them difficult, thus we exclude that as well. 
To create these plots, we first extracted embeddings from only the keypoint locations. 
These are 1024 dimensional for the None projection and 256 for other unsupervised methods. 
We then project these embeddings to 2D using t-SNE, and finally plot them. 
Each color represents a different keypoint type, which is different depending on the dataset.  

LEAD and ASYM look similar to original feature space. One interesting thing is that, CL manages to separate overlapping embeddings when compared to the `no projection' baseline on the AFLW dataset.  
This is reflected by their superior PCK scores for this dataset.  
However, for CUB there are cases where it splits clusters of a keypoints which were a single prominent cluster in the original embeddings space. 
This perhaps indicates that applying CL can sometimes destroy invariances that were captured in the pre-trained features, thus leading to undesirable changes in the embedding space. 
\input{figs/tsne/tsne}

\subsection{Keypoint Regression Evaluation}
As noted in the main paper, the two common types of evaluation paradigms for semantic correspondence estimation are:  (i) landmark/keypoint regression and (ii) feature matching. 
We chose to use feature matching for our results as is does not require additional supervision. 
However, for completeness here we evaluate embeddings from different unsupervised methods using the regression protocol on two face datasets; MAFL~\cite{MAFL} and AFLW$_M$\cite{AFLW}. 
AFLW$_M$ contains crops from the MTFL\cite{MTFL} dataset, which contains 2,995 examples for testing and 10,122 for training. 
This is the same dataset that we consider in our main paper as AFLW, as it was referred as AFLW$_M$ in some papers \cite{thewlis2019unsupervised,cheng2021equivariant,karmali2022lead} we present here as AFLW$_M$ as well. We report percentage of inter-ocular distance similar to previous work. Please note that lower is better in this metric.

We follow same approach as in \cite{thewlis2019unsupervised,cheng2021equivariant,karmali2022lead}, \ie we freeze the embedder $\Phi$ and train an additional regression head on top of these features. 
We use the unsupervised CNN trained on Imagenet for the feature encoder $\Psi$, and the unsupervised losses are finetuned on the AFLW dataset for both datasets to obtain embeddings which are input to the regression head. 
The results can be seen in Table~\ref{table:regression}. 
\input{tables/regression_exp}

We also compared the results taken directly from the original papers. 
While our re-implementation obtains reasonable scores, they are slightly worse than the original reported numbers. 
This can be explained by the fact that we use a basic encoder which produce dense feature maps in a lower spatial dimension. 
Compared to CL~\cite{cheng2021equivariant}, we use single layer features before projection, as opposed to higher dimensional hypercolumn features. 
Unlike the original LEAD~\cite{karmali2022lead} implementation, our projection operation is a single layer 1x1 2D convolution compared to a fully convolutional decoder which produces higher resolution features used in their paper. 
Unlike DVE~\cite{thewlis2019unsupervised}, we do not preform end-to-end finetuning. 
Also, for consistency with our other results the unsupervised losses in our implementations are finetuned on the AFLW dataset instead of CelebA~\cite{liu2015deep}, which is a larger dataset. 
While one may expect a large drop in performance due to these differences, there is in fact only a one pixel drop. 
This level of error is likely to be on the order, if not smaller, than human annotation inconsistency. 
This perhaps highlights the inadequacy of the regression evaluation as the supervision used during training makes the evaluation unfair. 
Furthermore, it again emphasizes that these types of face datasets are perhaps reaching saturation.

\subsection{Pre-training Source and Cross Dataset Evaluation}
Here we present the raw numbers for the pre-training data source and cross dataset evaluation experiments from main paper. 
The results can be found in Table~\ref{table:different_datasource} and Table~\ref{table:cross_projection_sup}, and correspond to the results in Fig.~3 and Fig.~4 in the main paper. 
\input{tables/pretraining_dataset}

\input{tables/cross_proj_sup}

%% file: tables/temp_exp.tex
\begin{table}[t]
\caption{Temperature ablation experiment for unsupervised losses on Spair-71K. Here we use the `Sup. pre-trained - CNN' encoder from main paper. 
With the exception of ASYM, all methods use $\tau_{1}$ as their $\tau$ and do not use $\tau_{2}$ at all.}
\centering
\begin{tabular}{llccccc}
\toprule
Metric~ & $\tau_{1}$~~ & ~~$\tau_{2}$~~ & ~DVE~ & ~CL~ & ~LEAD~ & ~ASYM~  \\ \midrule
\multirow{5}{*}{PCK} & 0.02 & 0.04 & 16.5 & 9.2 & 31.9 & 31.7 \\ %
& 0.05 & 0.1 & 16.3 & 8.2 & 31.7 & 32.1 \\
& 0.1 & 0.2 & 16.0 & 17.2 & 31.9 & 33.0 \\
& 0.2 & 0.4 & 15.7 & 26.6 & 31.4 & 34.0 \\ 
& 0.4 & 0.8 & 9.2 & 15.8 & 30.1 & 29.5 \\ %

\midrule
\multirow{5}{*}{$\text{PCK}^{\dag}$}& 0.02 & 0.04 & 12.9 & 7.5 & 25.5 & 25.4 \\ %
& 0.05 & 0.1 & 12.4 & 6.6 & 25.4 & 25.8 \\
& 0.1 & 0.2 & 12.4 & 13.8 & 25.4 & 26.6 \\
& 0.2 & 0.4 & 12.1 & 20.0 & 25.1 & 27.2 \\
& 0.4 & 0.8 & 6.9 & 11.2 & 23.8 & 23.1 \\ %
\bottomrule
\end{tabular}
\label{table:temp_exp}
\end{table}

%% file: tables/mse_ce.tex
\begin{table}[h]
\caption{Loss and temperature ablation for ASYM and LEAD on Spair-71K. 
For both methods, Mean Square Error (MSE) and Cross-Entropy (CE) losses are used. 
ASYM using CE with the same temperature value for both $\tau_1$  and $\tau_2$ is equivalent to LEAD. 
}
\centering
\begin{tabular}{llccc}
\toprule
Method~ & $\tau_{1}$~~ & ~~$\tau_{2}$~~ & ~MSE~ & ~CE~  \\ \midrule
\multirow{3}{*}{LEAD} & 0.05 & - & 31.5 & 31.7 \\
& 0.1 & - & 30.6 & 31.9 \\
& 0.2 & - & 29.9 & 31.4 \\ 
& 0.4 & - & 27.4 & 30.3 \\ \midrule
\multirow{3}{*}{ASYM} & 0.05 & 0.1 & 32.1 & 32.0 \\
& 0.1 & 0.2 & 33.0 & 32.8 \\
& 0.2 & 0.4 & 34.0 & 32.0 \\
\bottomrule
\end{tabular}
\label{table:mse_ces}
\end{table}

%% file: figs/lead_vs_asym/lead_vs_asym.tex
\begin{figure}[t]
     \centering
     \begin{subfigure}[b]{0.16\textwidth}
        \centering
         \includegraphics[height=2.1cm, trim=0 120 0 0, clip]{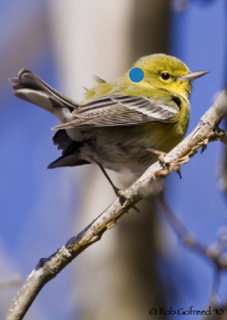}
         \caption{Source}
     \end{subfigure}
    \hfill \hfill
        \begin{subfigure}[b]{0.16\textwidth}
        \centering
         \includegraphics[height=2.1cm, trim=50 0 50 0, clip]{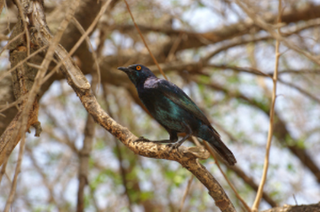}
         \caption{Target}
     \end{subfigure}
     \hfill \hfill
    \begin{subfigure}[b]{0.16\textwidth}
        \centering
         \includegraphics[height=2.1cm, trim=0 0 0 0, clip]{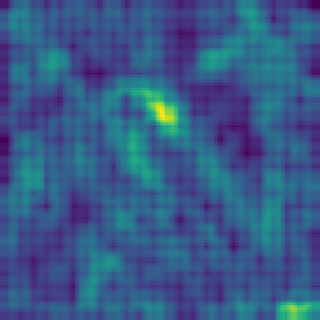}
         \caption{None}
     \end{subfigure}
    \hfill
    \begin{subfigure}[b]{0.16\textwidth}
        \centering
         \includegraphics[height=2.1cm, trim=0 0 0 0, clip]{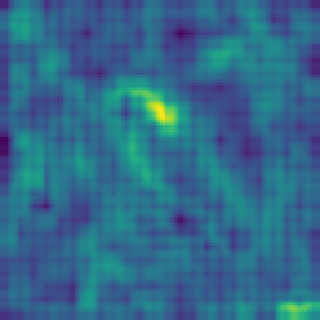}
         \caption{LEAD}
     \end{subfigure}
    \hfill
         \begin{subfigure}[b]{0.16\textwidth}
        \centering
         \includegraphics[height=2.1cm, trim=0 0 0 0, clip]{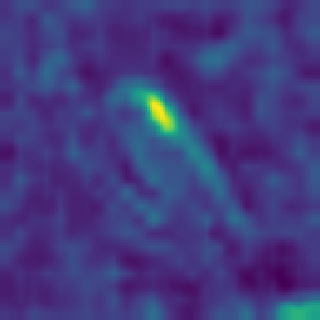}
         \caption{ASYM}
     \end{subfigure}
    \caption{Feature matching scores for different methods for the keypoint on on the birds head (indicated in blue) from the source images in (a) to the target in (b). 
    By design, LEAD matches the distribution from the original feature space shown in (c).
    We can see that our ASYM method results in a much more sharper distribution around the correct location compared to LEAD. }
    \label{fig:lead_vs_asym}
\end{figure}

%% file: figs/lead_vs_asym/hist_lead_asym.tex
\begin{figure}
    \vspace{-10pt}\parbox{\textwidth}{\center \textbf{Spair-71K}}
     \centering
     \begin{subfigure}[b]{0.30\textwidth}
        \centering
         \includegraphics[width=\textwidth, trim=0 0 0 0, clip]{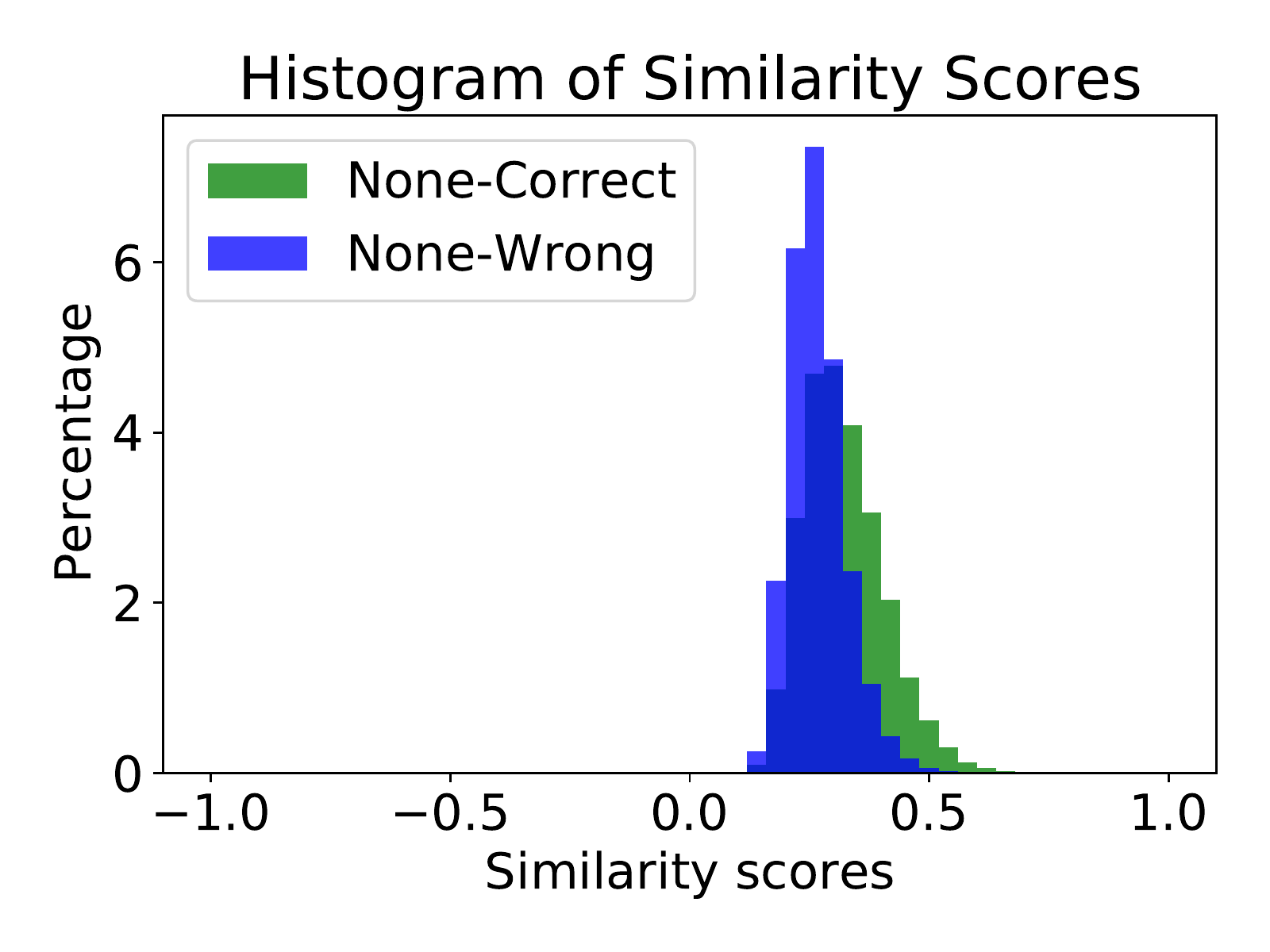}
     \end{subfigure}
    \hfill
        \begin{subfigure}[b]{0.30\textwidth}
        \centering
         \includegraphics[width=\textwidth, trim=0 0 0 0, clip]{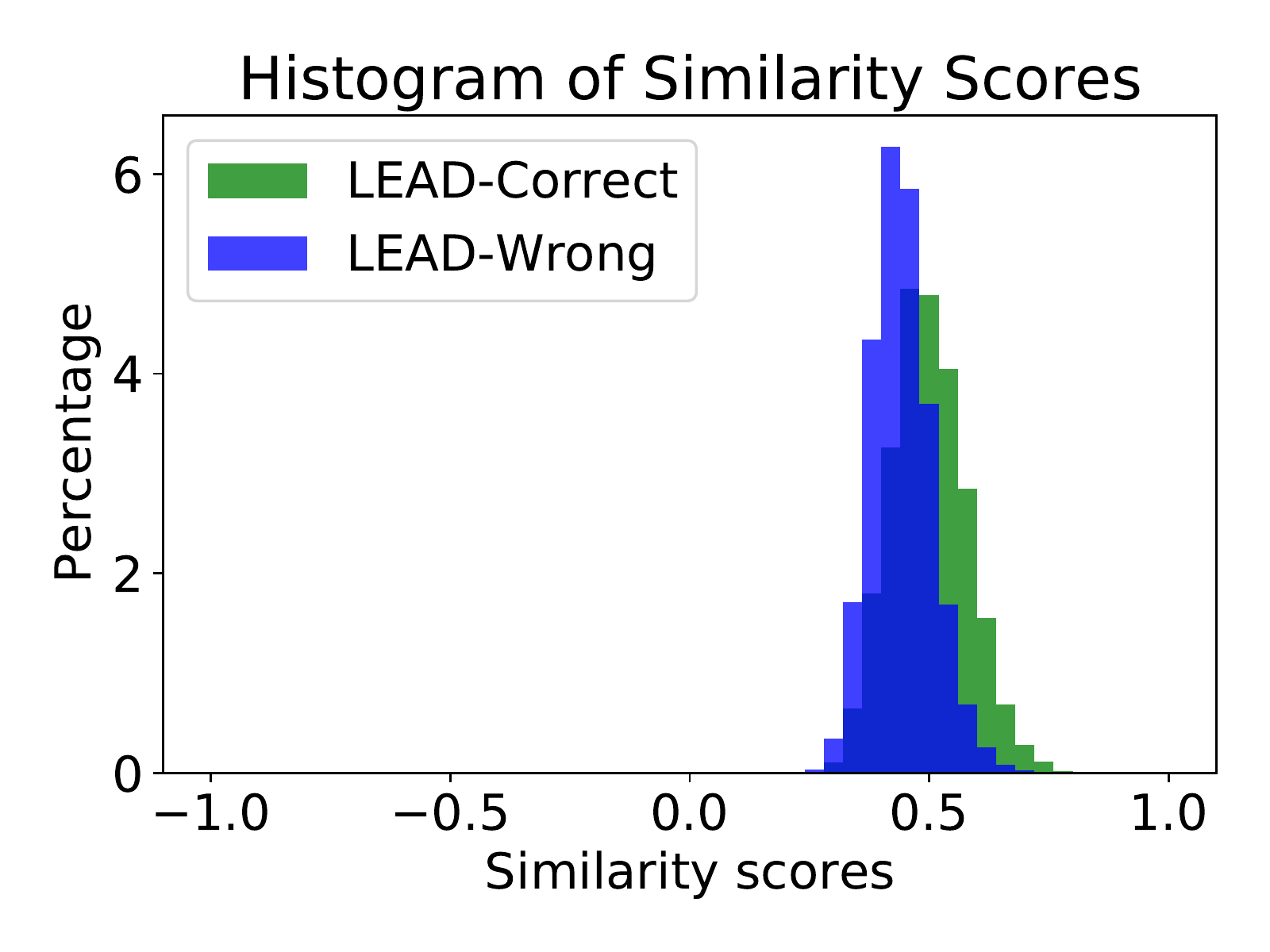}
     \end{subfigure}
     \hfill
    \begin{subfigure}[b]{0.30\textwidth}
        \centering
         \includegraphics[width=\textwidth, trim=0 0 0 0, clip]{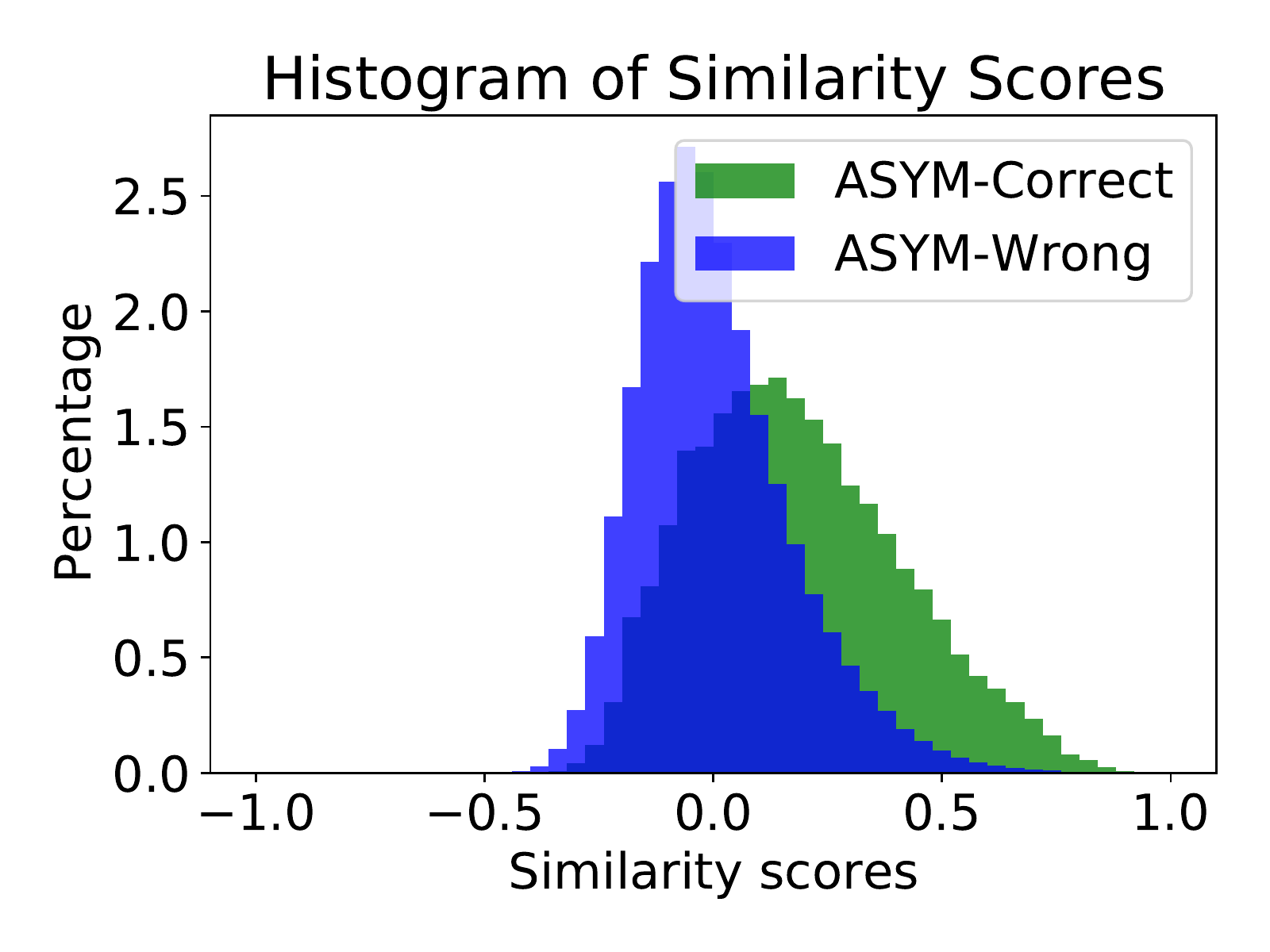}
     \end{subfigure}
      \parbox{\textwidth}{\center \textbf{Sdogs}}
     \centering
     \begin{subfigure}[b]{0.30\textwidth}
        \centering
         \includegraphics[width=\textwidth, trim=0 0 0 0, clip]{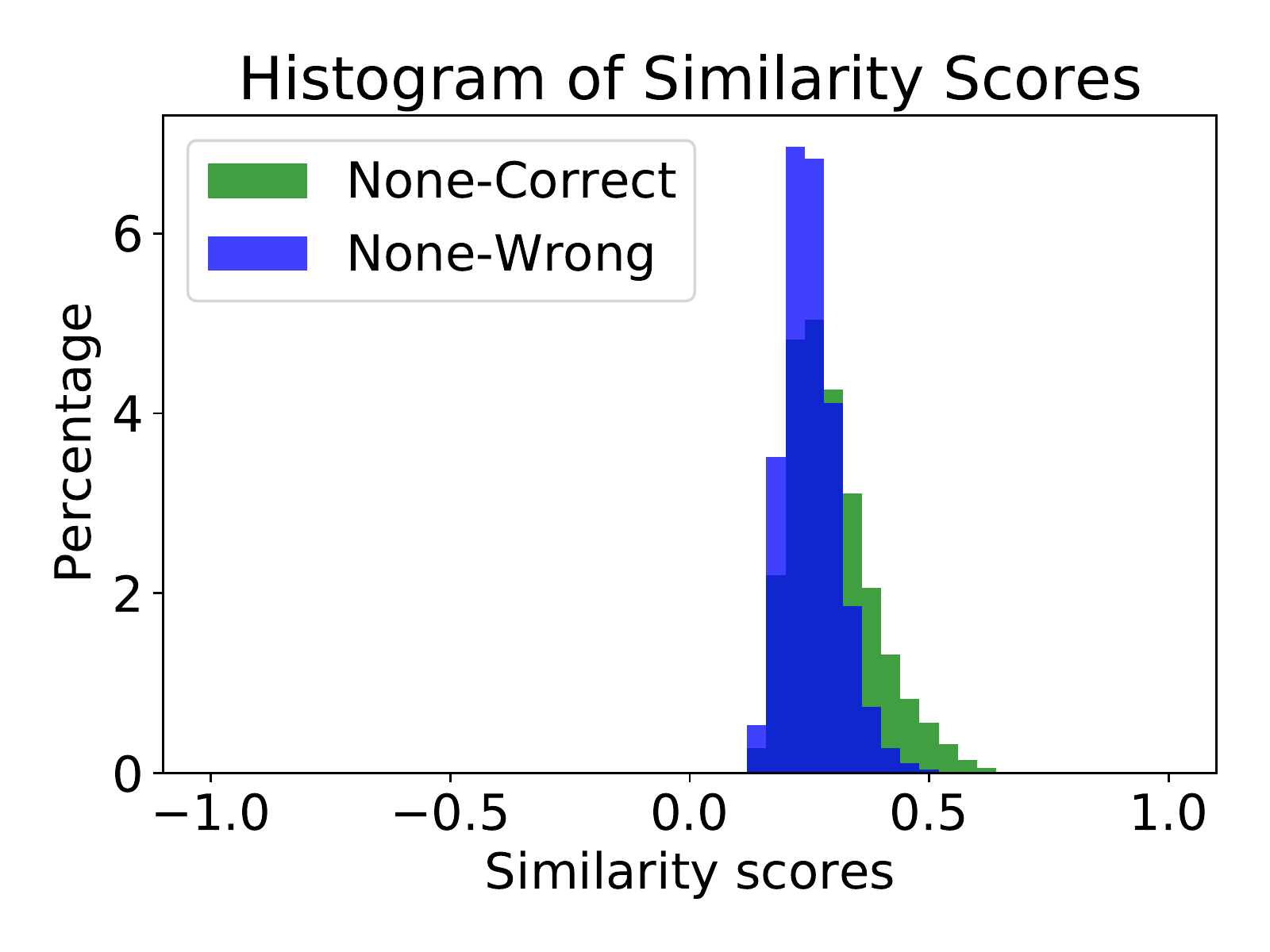}
     \end{subfigure}
    \hfill
        \begin{subfigure}[b]{0.30\textwidth}
        \centering
         \includegraphics[width=\textwidth, trim=0 0 0 0, clip]{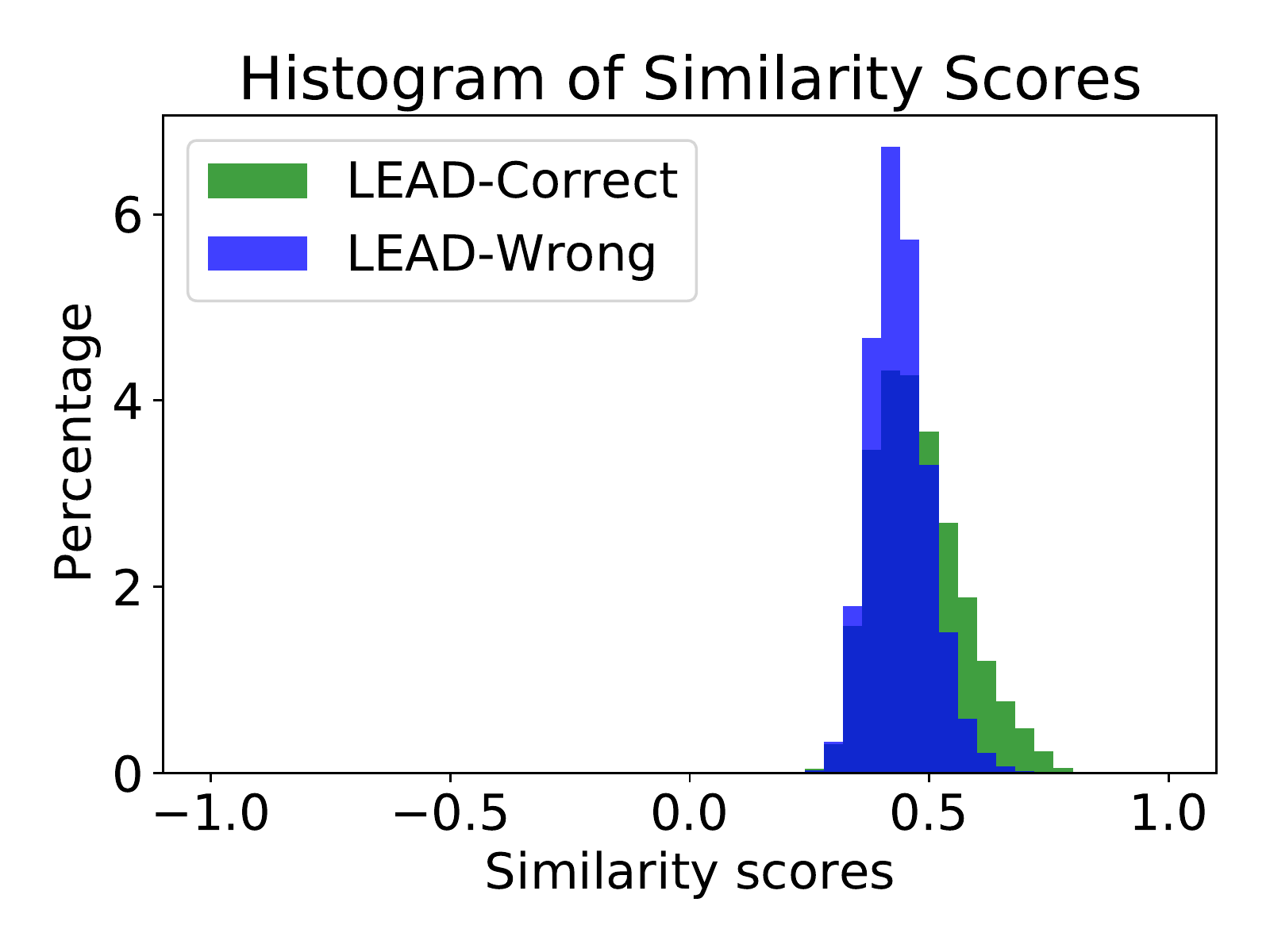}
     \end{subfigure}
     \hfill
    \begin{subfigure}[b]{0.30\textwidth}
        \centering
         \includegraphics[width=\textwidth, trim=0 0 0 0, clip]{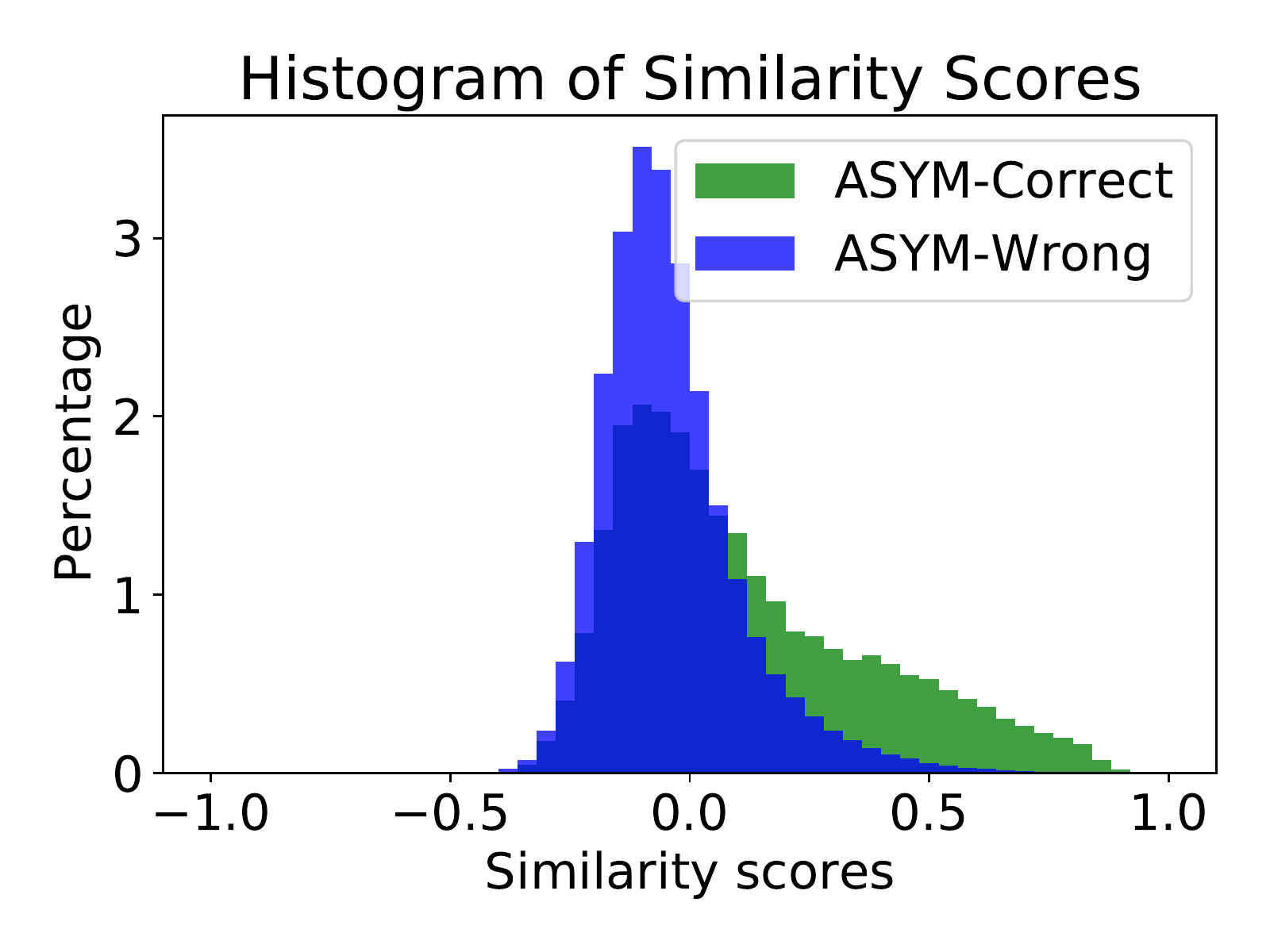}
     \end{subfigure}
      \parbox{\textwidth}{\center \textbf{CUB}}
     \centering
     \begin{subfigure}[b]{0.32\textwidth}
        \centering
         \includegraphics[width=\textwidth, trim=0 0 0 0, clip]{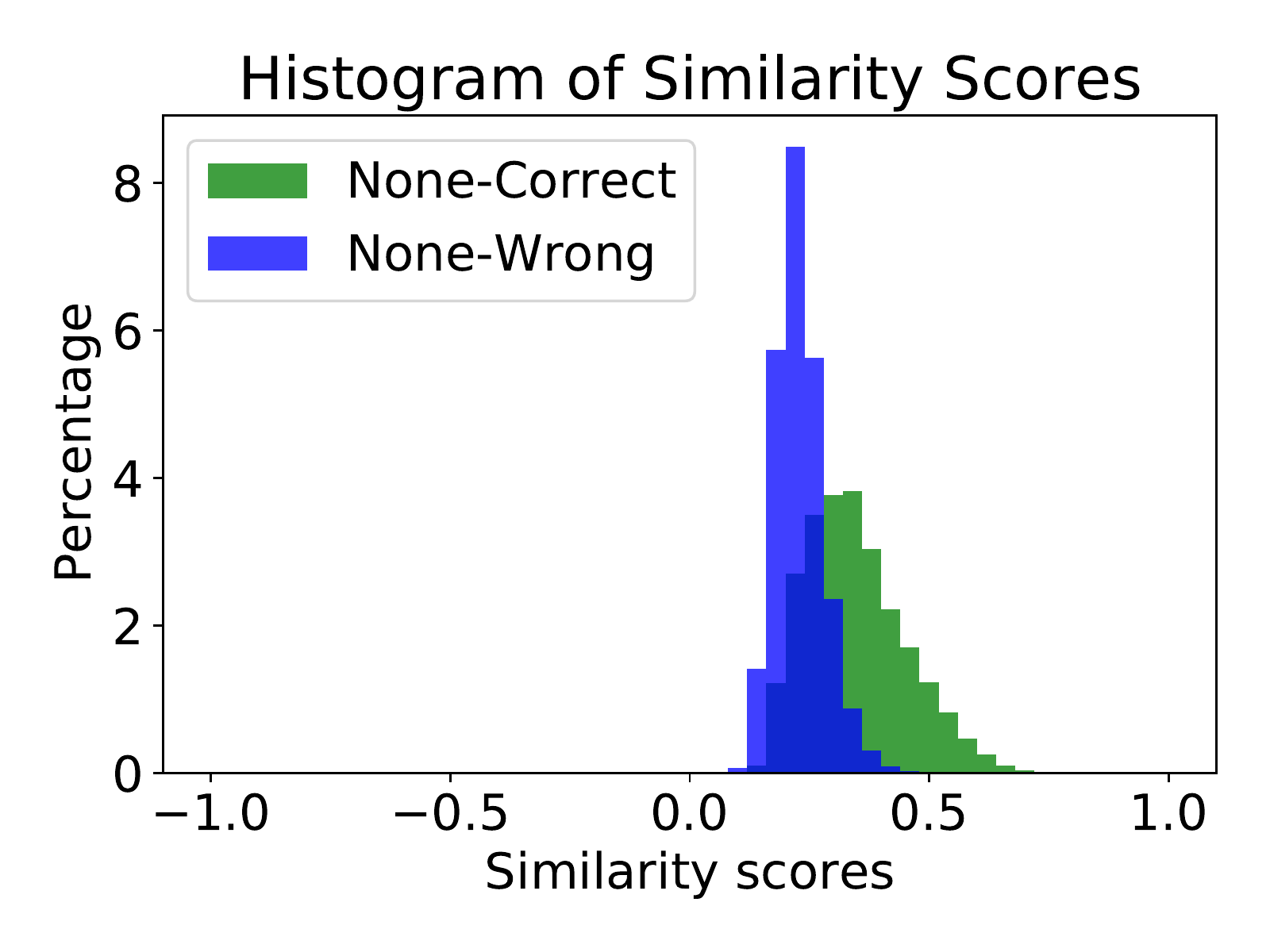}
     \end{subfigure}
    \hfill
        \begin{subfigure}[b]{0.32\textwidth}
        \centering
         \includegraphics[width=\textwidth, trim=0 0 0 0, clip]{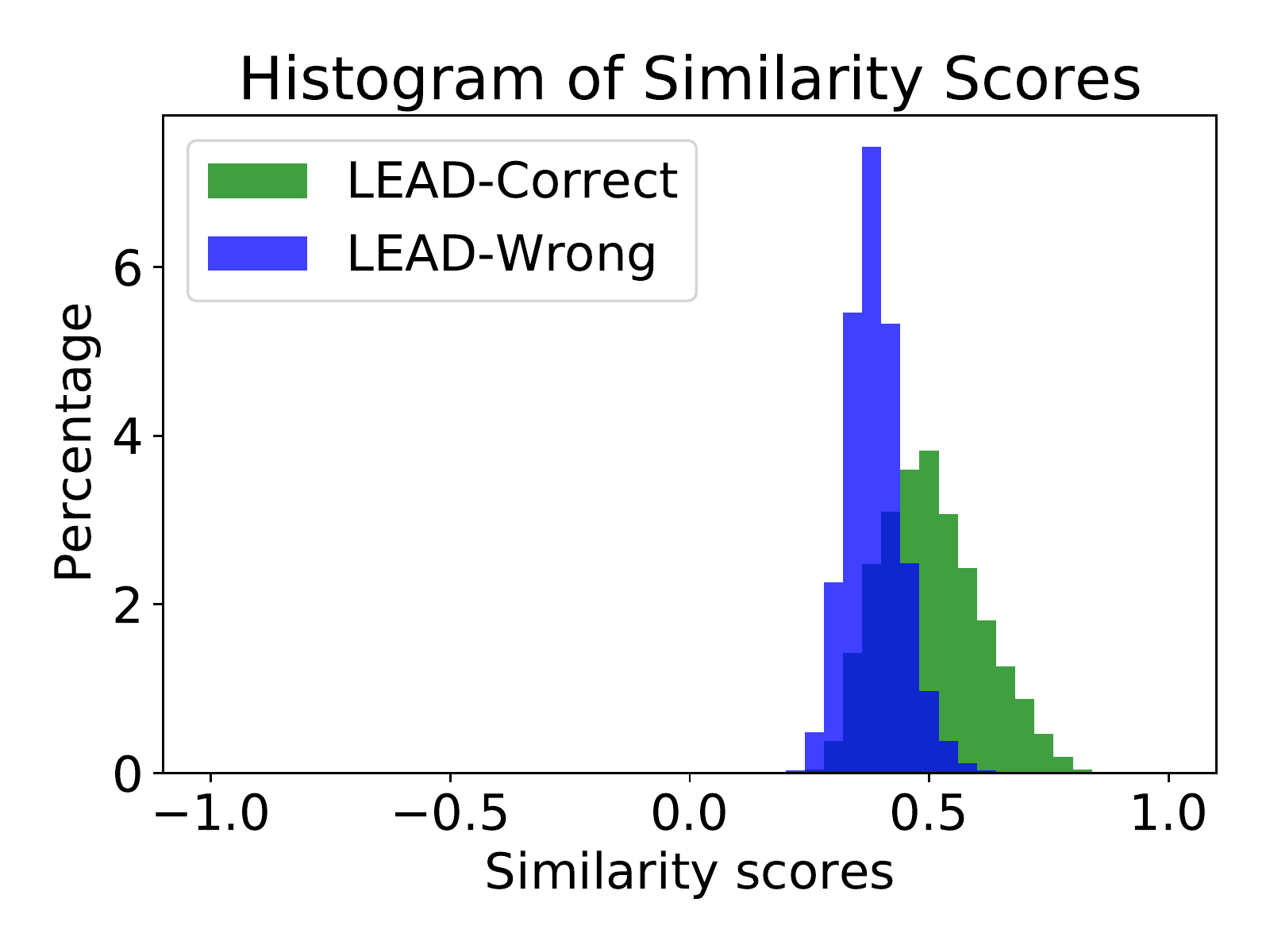}
     \end{subfigure}
     \hfill
    \begin{subfigure}[b]{0.32\textwidth}
        \centering
         \includegraphics[width=\textwidth, trim=0 0 0 0, clip]{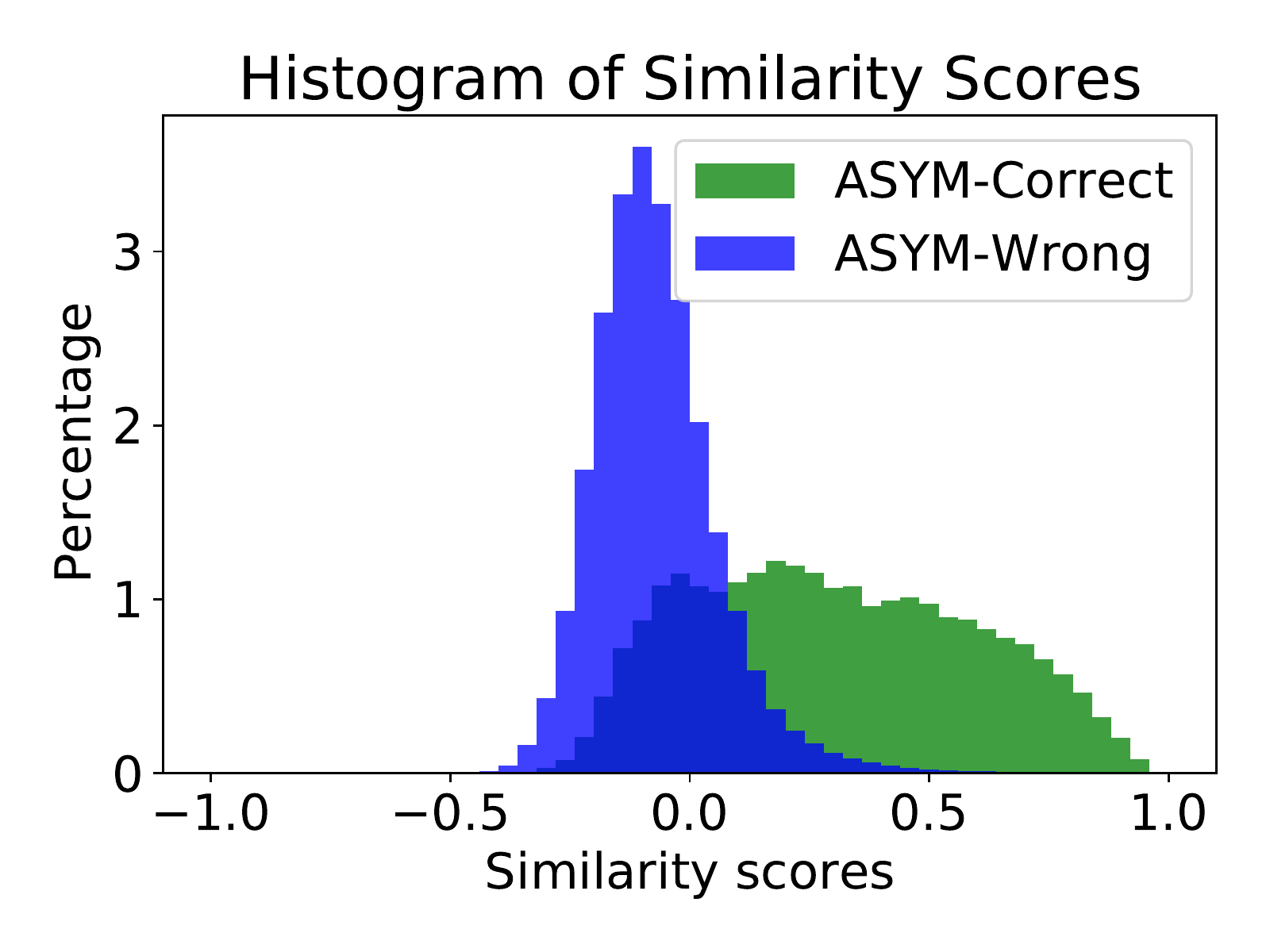}
         
     \end{subfigure}
      \parbox{\textwidth}{\center \textbf{AFLW}}
    \centering
     \begin{subfigure}[b]{0.32\textwidth}
        \centering
         \includegraphics[width=\textwidth, trim=0 0 0 0, clip]{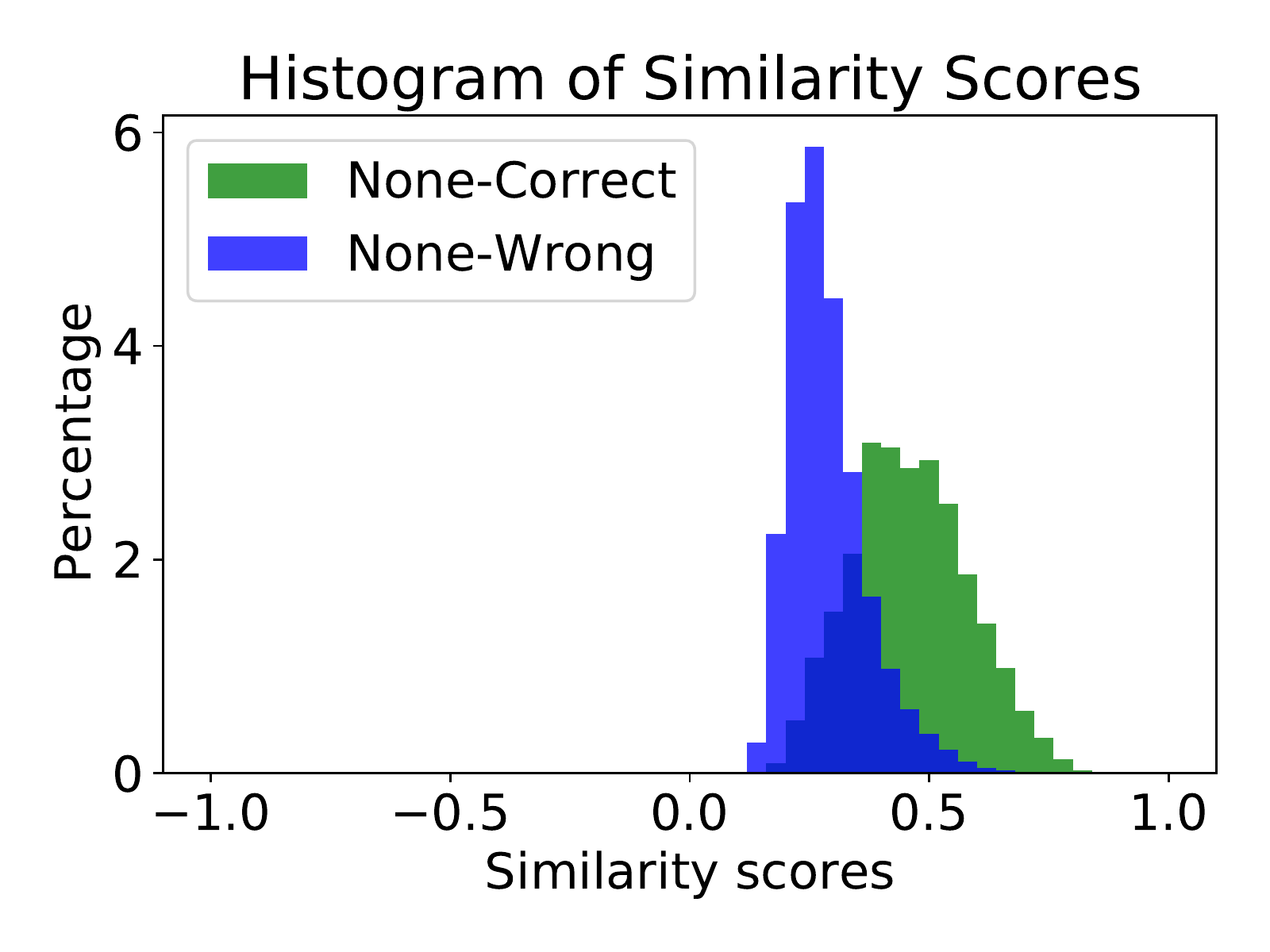}
     \end{subfigure}
    \hfill
        \begin{subfigure}[b]{0.32\textwidth}
        \centering
         \includegraphics[width=\textwidth, trim=0 0 0 0, clip]{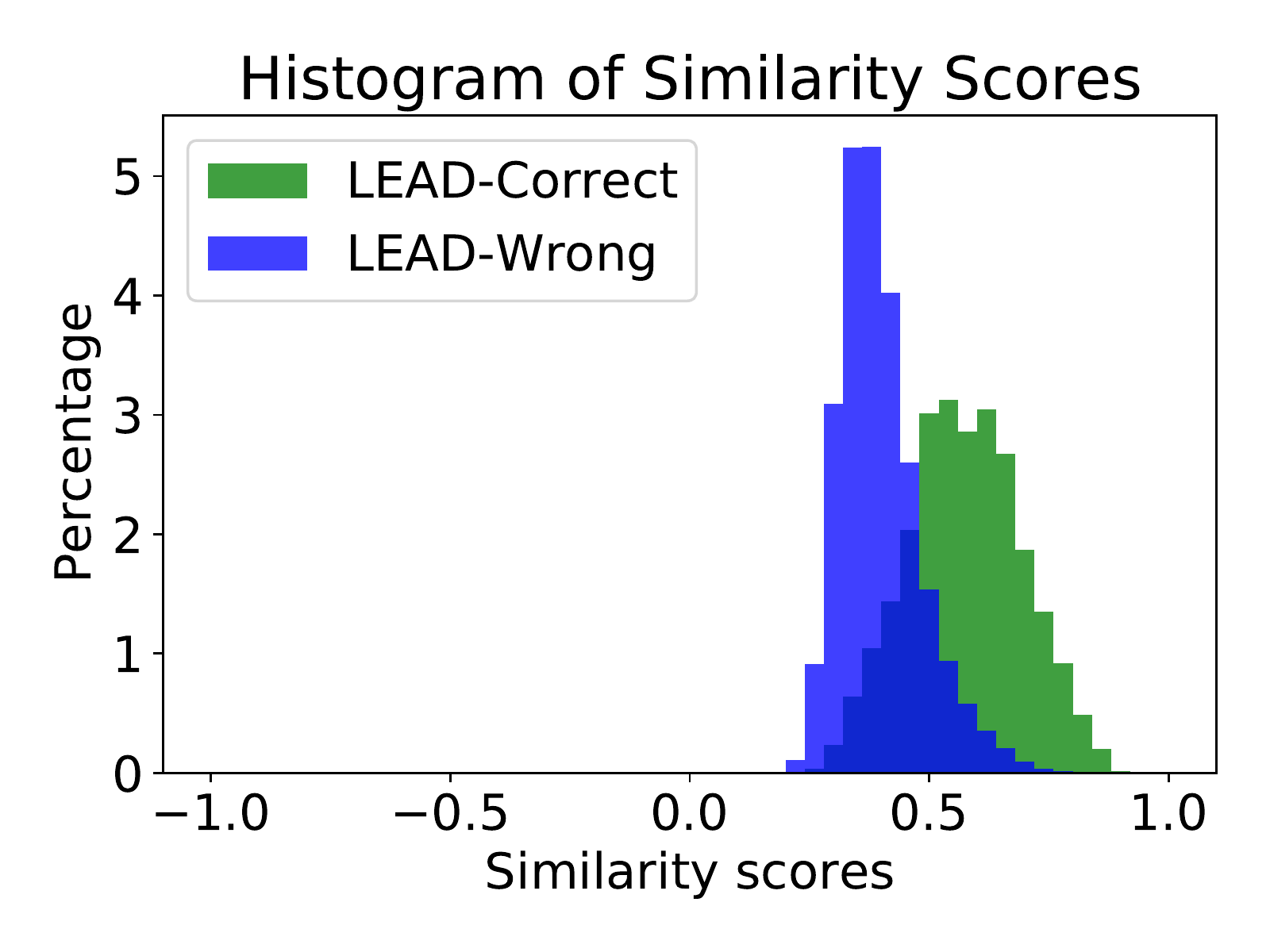}
     \end{subfigure}
     \hfill
    \begin{subfigure}[b]{0.32\textwidth}
        \centering
         \includegraphics[width=\textwidth, trim=0 0 0 0, clip]{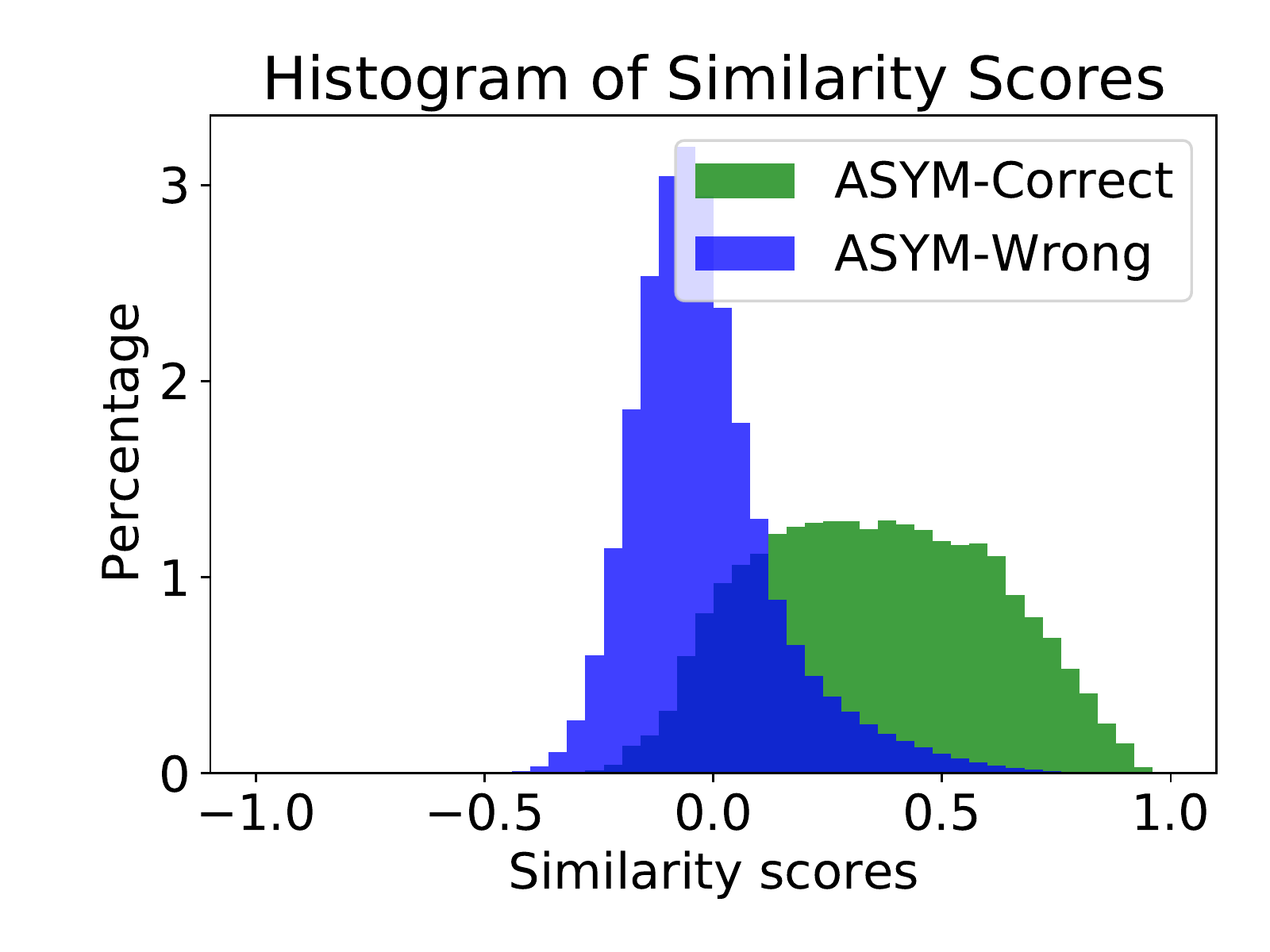}
     \end{subfigure}
    
\parbox{\textwidth}{\center \textbf{Awa}}
    \centering
     \begin{subfigure}[b]{0.32\textwidth}
        \centering
         \includegraphics[width=\textwidth, trim=0 0 0 0, clip]{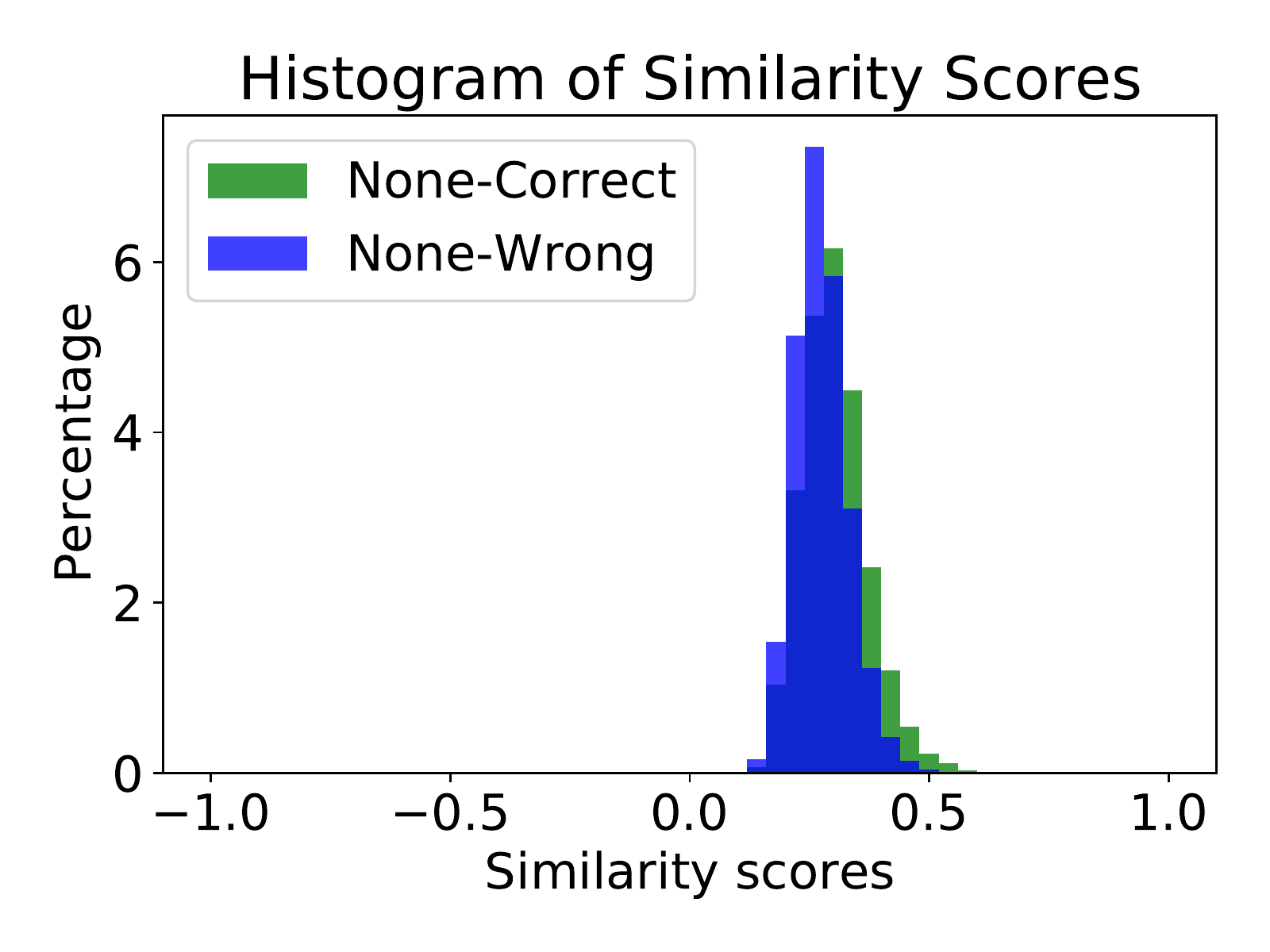}
         \vspace{-15pt}\caption{None}
     \end{subfigure}
    \hfill
        \begin{subfigure}[b]{0.32\textwidth}
        \centering
         \includegraphics[width=\textwidth, trim=0 0 0 0, clip]{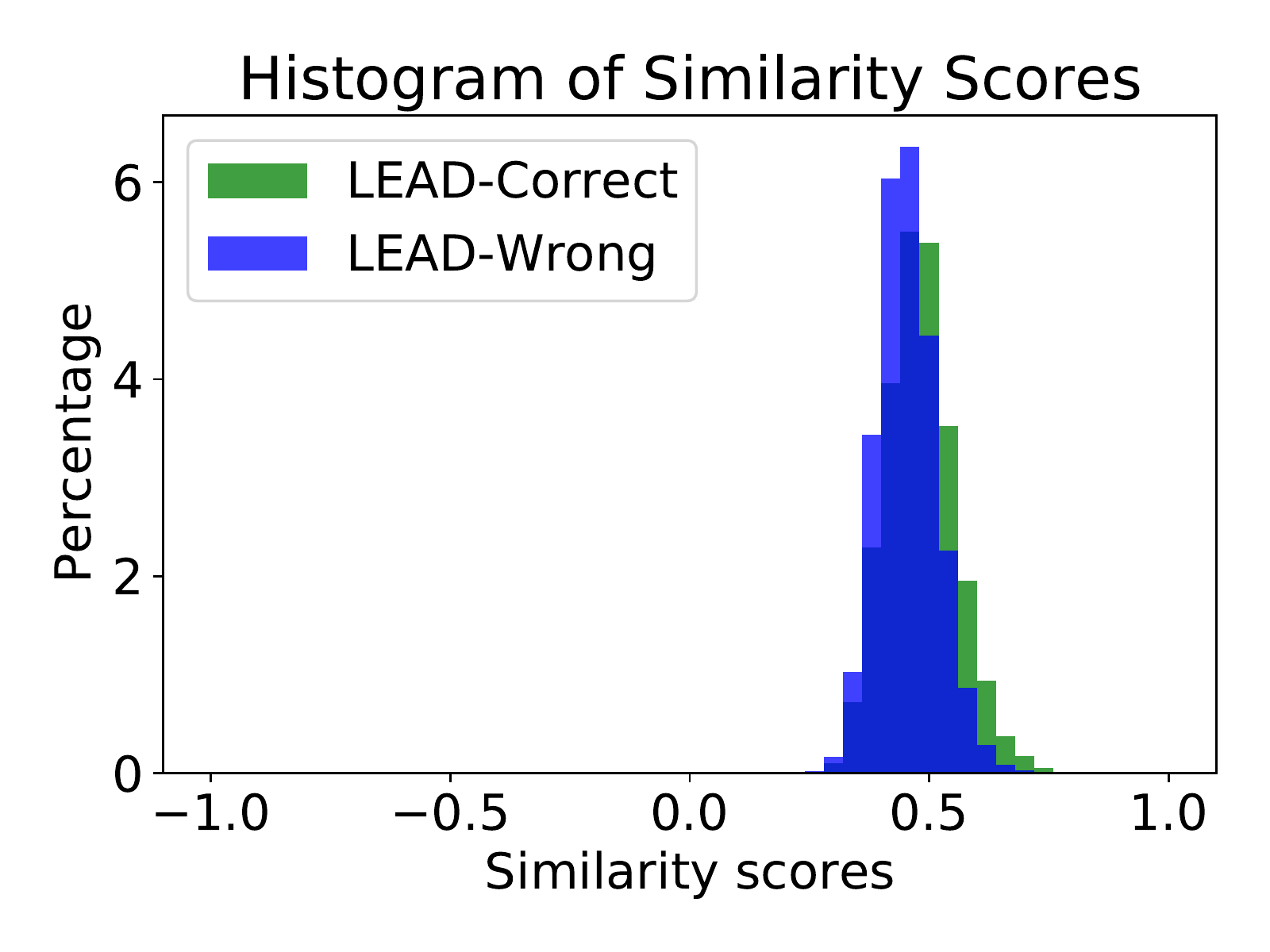}
         \vspace{-15pt}\caption{LEAD}
     \end{subfigure}
     \hfill
    \begin{subfigure}[b]{0.32\textwidth}
        \centering
         \includegraphics[width=\textwidth, trim=0 0 0 0, clip]{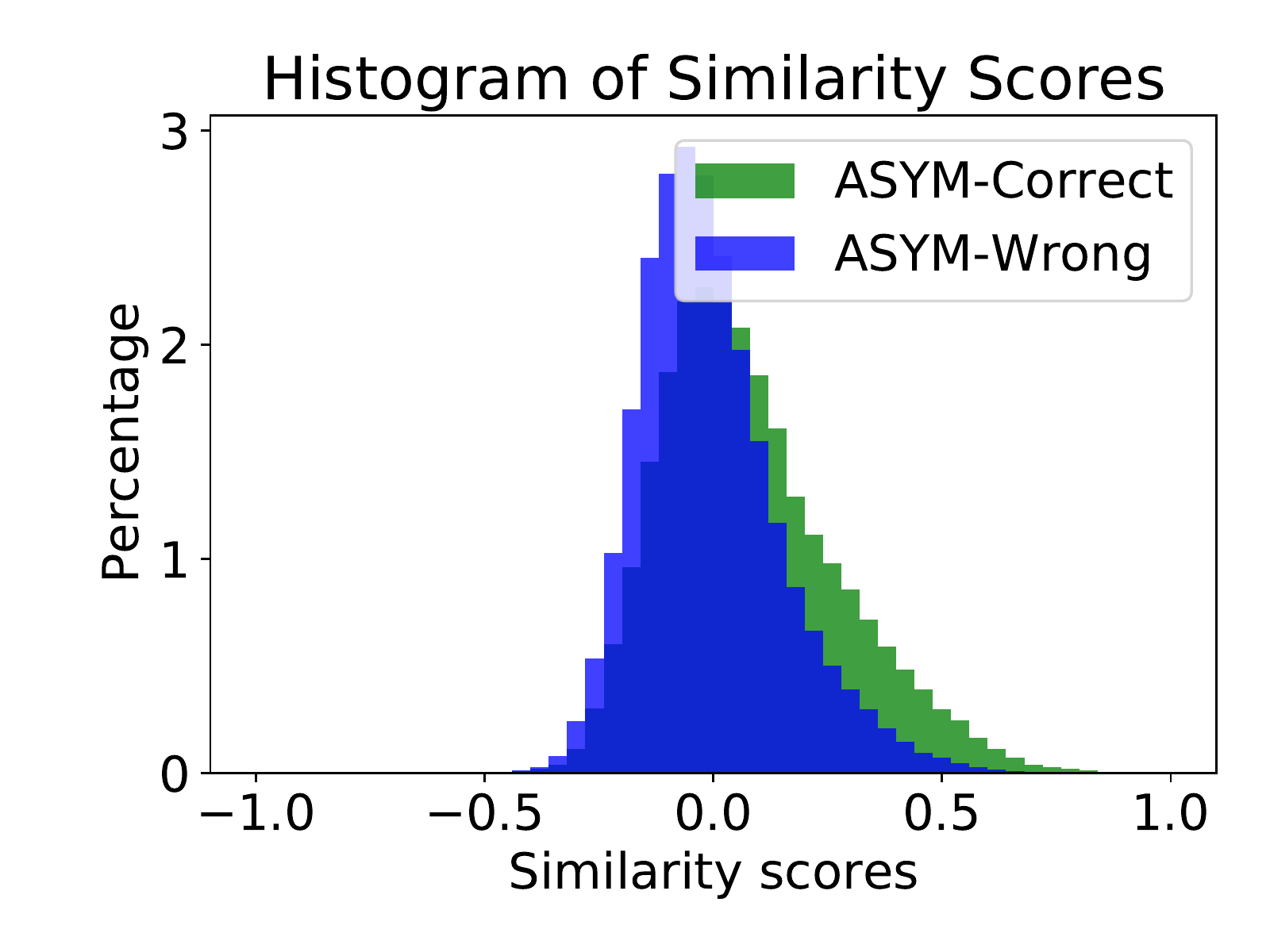}
         \vspace{-15pt}\caption{ASYM}
     \end{subfigure}
    \caption{Histograms for cosine similarity scores of embeddings for (a) None, (b) LEAD, and (c) ASYM. Each row is a different dataset. }
    \label{fig:hist_lead_vs_asym}
\end{figure}

%% file: tables/feature_layer.tex
\begin{table}
\centering
\caption{Evaluation of using pre-trained features from different layers for the Resnet50 trained with Imagenet. 
The results here for $\text{conv}_{3}$ correspond to the no projection model (\ie `None) from Table 2 (a) in the main paper. 
} 
\begin{tabular}{lccccc}
\toprule
Layer & ~Spair-71K~ & ~SDogs~ & ~CUB~ & ~AFLW~ & ~Awa~ \\ \midrule
$\text{conv}_{1}$  & 7.3 & 5.1 & 7.9  & 11.6 & 5.6  \\
$\text{conv}_{2}$  & 12.9 & 8.6 & 13.3 & 27.2 & 9.1  \\
$\text{conv}_{3}$  & 31.8 & 34.9 & 51.3 & 57.4 & 28.8 \\
$\text{conv}_{4}$ & 15.8 & 10.3 & 14.0 & 31.3 & 9.3  \\
\bottomrule
\end{tabular}
\label{table:layers}
\end{table}

%% file: tables/pckdag_otherdatasets.tex
\begin{table}
\caption{Evaluation of error types across four different datasets. In addition to PCK, we also report scores for our $\text{PCK}^{\dag}$ metric. Results for  Spair-71 are presented in Table~3 in the main paper.}
\centering
\subfloat[SDogs]{
\resizebox{0.48\linewidth}{!}{
\begin{tabular}{lccccc}
\toprule
Method & ~Miss$\downarrow$~ & ~Jitter$\downarrow$~ & ~Swap$\downarrow$~ & ~PCK$\uparrow$~  & ~$\text{PCK}^{\dag}\uparrow$\\ \midrule
EQ & 55.9 & 21.4 & 25.9 & 21.2 & 18.2 \\
DVE & 57.7 & 21.8 & 24.8 & 20.5 & 17.5 \\
CL & 40.9 & 17.9 & 27.3 & 37.0 & 31.9 \\
LEAD & 38.0 & 16.2 & 31.2 & 35.1 & 30.8 \\
ASYM & 33.1 & 16.3 & 31.4 & 40.4 & 35.5 \\ \midrule
Supervised & 23.7 & 16.7 & 29.0 & 53.2 & 47.3 \\
\bottomrule
\end{tabular}
}
}
\subfloat[CUB]{
\resizebox{0.48\linewidth}{!}{
\begin{tabular}{lccccc}
\toprule
Method & ~Miss$\downarrow$~ & ~Jitter$\downarrow$~ & ~Swap$\downarrow$~ & ~PCK$\uparrow$~  & ~$\text{PCK}^{\dag}\uparrow$\\ \midrule
EQ & 44.0 & 24.8 & 35.2 & 28.1 & 20.9 \\
DVE & 44.3 & 24.6 & 35.7 & 27.7 & 20.0 \\
CL & 24.8 & 20.1 & 34.6 & 54.5 & 40.7 \\
LEAD & 28.1 & 17.4 & 31.8 & 51.5 & 40.1 \\
ASYM & 21.7 & 16.9 & 29.8 & 60.8 & 48.5 \\ \midrule
Supervised & 14.3 & 15.2 & 25.4 & 72.7 & 60.2 \\
\bottomrule
\end{tabular}
}
}

\subfloat[AFLW]{
\resizebox{0.48\linewidth}{!}{
\begin{tabular}{lccccc}
\toprule
Method & ~Miss$\downarrow$~ & ~Jitter$\downarrow$~ & ~Swap$\downarrow$~ & ~PCK$\uparrow$~  & ~$\text{PCK}^{\dag}\uparrow$\\ \midrule
EQ & 38.0 & 26.0 & 14.2 & 48.5 & 47.8 \\
DVE & 24.9 & 21.2 & 17.3 & 58.7 & 57.8 \\
CL & 18.0 & 11.4 & 15.2 & 67.3 & 66.8 \\
LEAD & 13.6 & 10.7 & 28.8 & 58.0 & 57.5 \\
ASYM & 11.7 & 7.9 & 25.2 & 63.6 & 63.1 \\ \midrule
Supervised & 7.0 & 4.7 & 12.7 & 80.8 & 80.4 \\
\bottomrule
\end{tabular}
}
}
\subfloat[AWA]{
\resizebox{0.48\linewidth}{!}{
\begin{tabular}{lccccc}
\toprule
Method & ~Miss$\downarrow$~ & ~Jitter$\downarrow$~ & ~Swap$\downarrow$~ & ~PCK$\uparrow$~  & ~$\text{PCK}^{\dag}\uparrow$\\ \midrule
EQ & 52.0 & 19.6 & 38.7 & 15.6 & 10.3 \\
DVE & 52.1 & 19.2 & 37.8 & 15.4 & 10.1 \\
CL & 38.4 & 16.8 & 41.5 & 31.7 & 20.1 \\
LEAD & 37.1 & 16.3 & 44.0 & 29.1 & 18.9 \\
ASYM & 32.2 & 16.7 & 45.6 & 34.1 & 22.1 \\ \midrule
Supervised & 23.4 & 18.3 & 46.3 & 46.1 & 30.3 \\
\bottomrule
\end{tabular}
}
}
\label{table:pckdag_alldatasets}
\end{table}

%% file: figs/examples_from_data.tex
\begin{figure}
        \centering 
        \parbox{0.195\textwidth}{\center Spair}
        \parbox{0.175\textwidth}{\center SDogs}
        \parbox{0.185\textwidth}{\center CUB}
        \parbox{0.145\textwidth}{\center AFLW}
        \parbox{0.195\textwidth}{\center Awa}

        \includegraphics[width=0.195\textwidth]{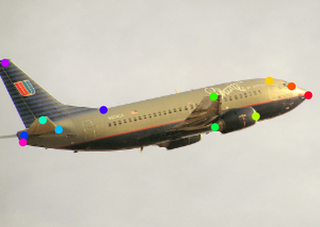}
        \includegraphics[width=0.175\textwidth]{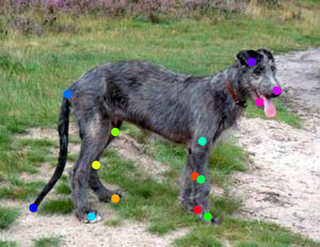}
        \includegraphics[width=0.185\textwidth]{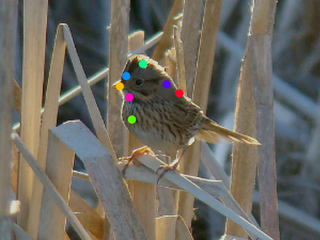}
        \includegraphics[width=0.145\textwidth]{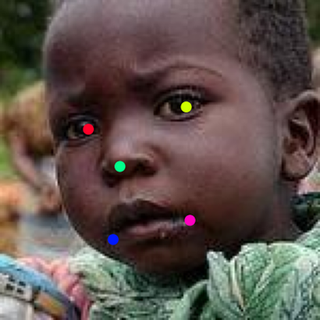}
        \includegraphics[width=0.195\textwidth]{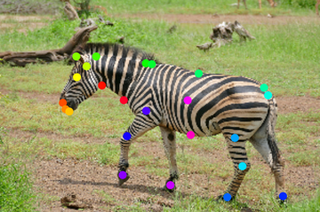}

        \includegraphics[width=0.195\textwidth]{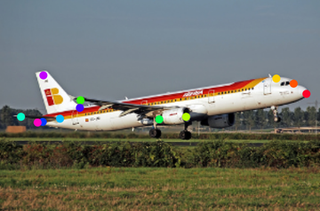}
        \includegraphics[width=0.175\textwidth]{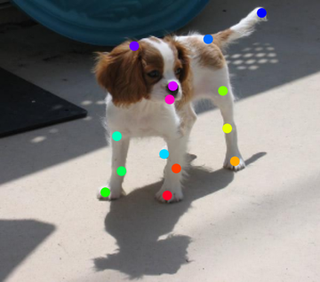}
        \includegraphics[width=0.185\textwidth]{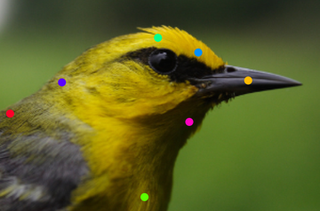}
        \includegraphics[width=0.145\textwidth]{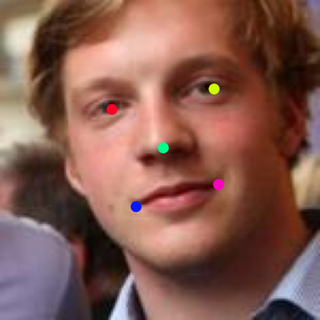}
        \includegraphics[width=0.195\textwidth]{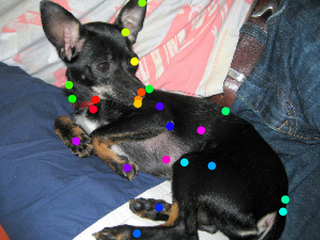}

    \caption{Examples from each of the datasets with the keypoint annotations that we consider in our paper. The pop row illustrates a source instance and the bottom a target instance. %
    }
    \label{fig:examples}
\end{figure}

%% file: figs/qualitative.tex
\begin{figure}
     \centering
     \begin{subfigure}[b]{0.17\textwidth}
        \centering
         \includegraphics[width=\textwidth, trim=0 0 0 0, clip]{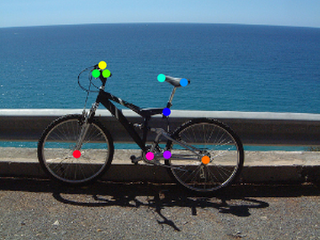}
     \end{subfigure}
     \hfill
     \begin{subfigure}[b]{0.19\textwidth}
        \centering
         \includegraphics[width=\textwidth, trim=0 0 0 0, clip]{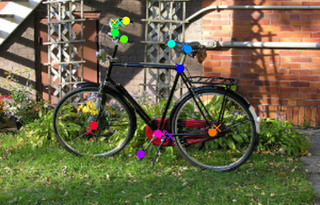}
     \end{subfigure}
     \hfill
     \begin{subfigure}[b]{0.19\textwidth}
        \centering
         \includegraphics[width=\textwidth,  trim=0 0 0 0, clip]{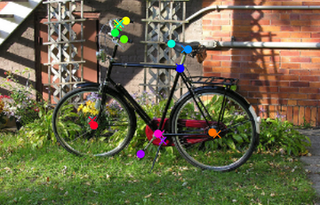}
     \end{subfigure}
     \hfill
     \begin{subfigure}[b]{0.19\textwidth}
        \centering
         \includegraphics[width=\textwidth,  trim=0 0 0 0, clip]{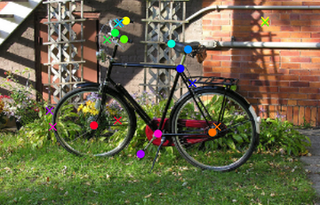}
     \end{subfigure}
     \hfill
     \begin{subfigure}[b]{0.19\textwidth}
        \centering
         \includegraphics[width=\textwidth,  trim=0 0 0 0, clip]{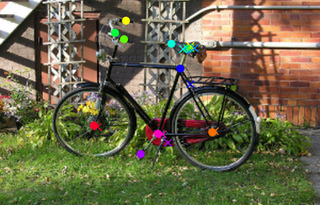}
     \end{subfigure}
     
     \centering
     \begin{subfigure}[b]{0.24\textwidth}
        \centering
         \includegraphics[width=\textwidth, trim=0 0 0 0, clip]{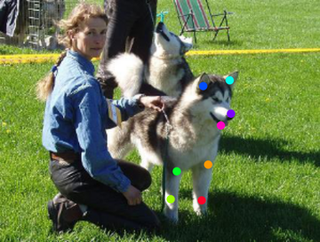}
     \end{subfigure}
     \hfill
     \begin{subfigure}[b]{0.17\textwidth}
        \centering
         \includegraphics[width=\textwidth, trim=0 40 0 10, clip]{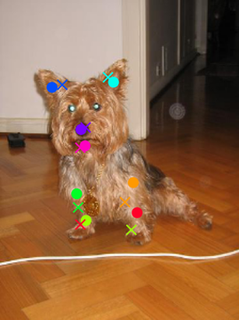}
     \end{subfigure}
     \hfill
     \begin{subfigure}[b]{0.17\textwidth}
        \centering
         \includegraphics[width=\textwidth,  trim=0 40 0 10, clip]{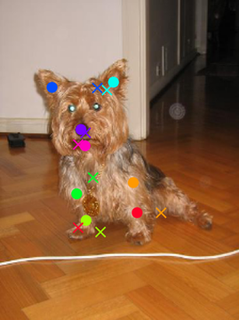}
     \end{subfigure}
     \hfill
     \begin{subfigure}[b]{0.17\textwidth}
        \centering
         \includegraphics[width=\textwidth,  trim=0 40 0 10, clip]{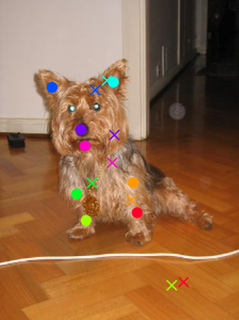}
     \end{subfigure}
     \hfill
     \begin{subfigure}[b]{0.17\textwidth}
        \centering
         \includegraphics[width=\textwidth,  trim=0 40 0 10, clip]{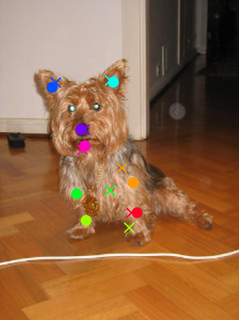}
     \end{subfigure}
     \centering
     \begin{subfigure}[b]{0.18\textwidth}
        \centering
         \includegraphics[width=\textwidth, trim=0 0 0 0, clip]{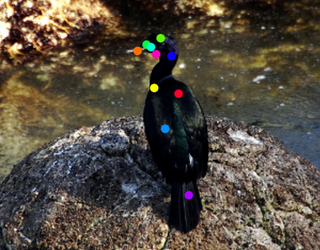}
     \end{subfigure}
     \hfill
     \begin{subfigure}[b]{0.19\textwidth}
     \centering
         \includegraphics[width=\textwidth, trim=0 0 0 0, clip]{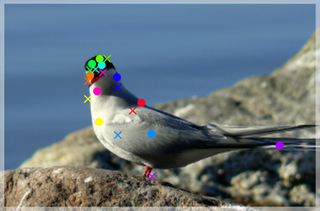}
     \end{subfigure}
     \hfill
      \begin{subfigure}[b]{0.19\textwidth}
     \centering
         \includegraphics[width=\textwidth, trim=0 0 0 0, clip]{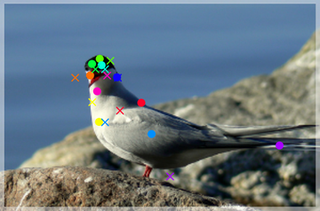}
     \end{subfigure}
     \hfill
          \begin{subfigure}[b]{0.19\textwidth}
     \centering
         \includegraphics[width=\textwidth, trim=0 0 0 0, clip]{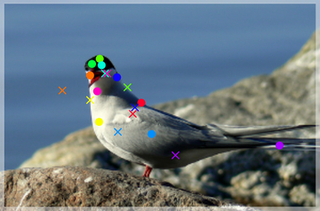}
     \end{subfigure}
     \hfill
          \begin{subfigure}[b]{0.19\textwidth}
     \centering
         \includegraphics[width=\textwidth, trim=0 0 0 0, clip]{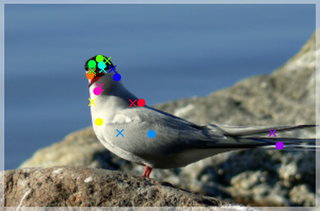}
     \end{subfigure}

     \centering
     \begin{subfigure}[b]{0.19\textwidth}
        \centering
         \includegraphics[width=\textwidth, trim=0 0 0 0, clip]{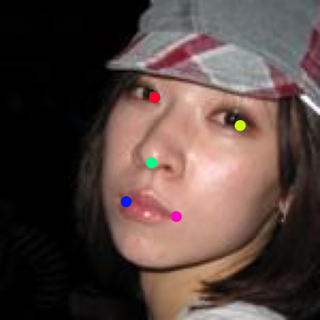}
     \end{subfigure}
     \hfill
     \begin{subfigure}[b]{0.19\textwidth}
        \centering
         \includegraphics[width=\textwidth, trim=0 0 0 0, clip]{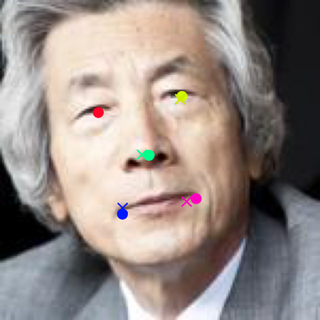}
     \end{subfigure}
     \hfill
     \begin{subfigure}[b]{0.19\textwidth}
        \centering
         \includegraphics[width=\textwidth,  trim=0 0 0 0, clip]{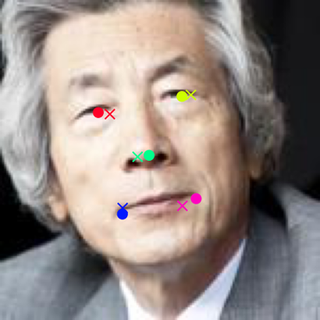}
     \end{subfigure}
     \hfill
     \begin{subfigure}[b]{0.19\textwidth}
        \centering
         \includegraphics[width=\textwidth,  trim=0 0 0 0, clip]{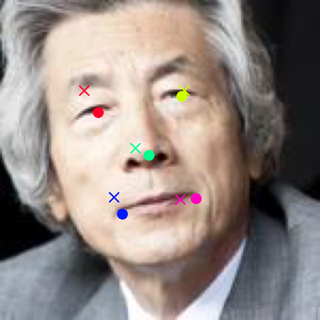}
     \end{subfigure}
     \hfill
     \begin{subfigure}[b]{0.19\textwidth}
        \centering
         \includegraphics[width=\textwidth,  trim=0 0 0 0, clip]{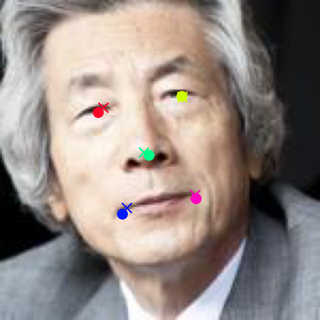}
     \end{subfigure}
     
     \centering
     \begin{subfigure}[b]{0.15\textwidth}
        \centering
         \includegraphics[width=\textwidth, trim=0 0 0 0, clip]{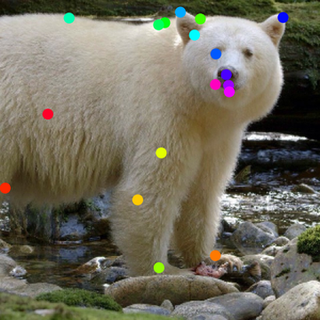} \caption{Source}
     \end{subfigure}
     \hfill
     \begin{subfigure}[b]{0.203\textwidth}
        \centering
         \includegraphics[width=\textwidth, trim=0 0 0 0, clip]{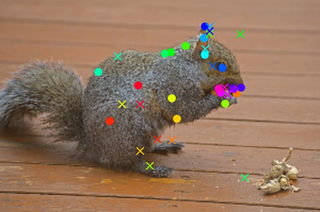} \caption{ASYM}
     \end{subfigure}
     \hfill
     \begin{subfigure}[b]{0.203\textwidth}
        \centering
         \includegraphics[width=\textwidth,  trim=0 0 0 0, clip]{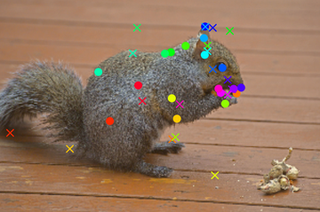} \caption{CL}
     \end{subfigure}
     \hfill
     \begin{subfigure}[b]{0.203\textwidth}
        \centering
         \includegraphics[width=\textwidth,  trim=0 0 0 0, clip]{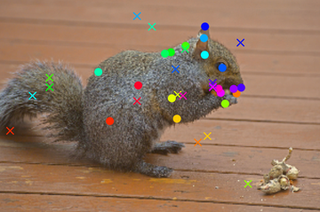}\caption{DVE}
     \end{subfigure}
     \hfill
     \begin{subfigure}[b]{0.203\textwidth} 
        \centering
         \includegraphics[width=\textwidth,  trim=0 0 0 0, clip]{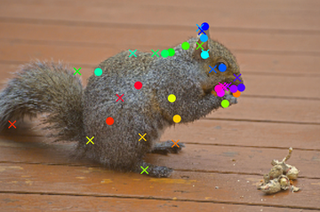}\caption{Sup}
     \end{subfigure}
        \caption{Qualitative matching results. 
        Each row is a different dataset: Spair, SDogs, CUB, AFLW, and Awa, from top to bottom. 
        The left most image for each row is a source example, and the remaining images visualize matches from different unsupervised methods, where 'o' indicates a ground truth location and  'x' indicates a prediction. Overall, while ASYM cannot match with the performance of Supervised projection, it is better than other unsupervised methods. For instance, in the AFLW example, only our proposed ASYM and supervised baseline able to precisely find correspondences for the all keypoints. }
        \label{fig:qual}
\end{figure}

%% file: figs/qualitative2.tex
\begin{figure}
     \centering
     \begin{subfigure}[b]{0.18\textwidth}
        \centering
         \includegraphics[width=\textwidth, trim=0 30 0 120, clip]{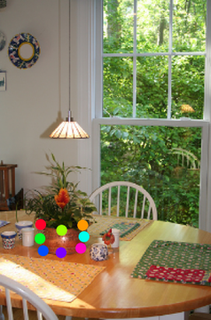}
     \end{subfigure}
     \hfill
     \begin{subfigure}[b]{0.19\textwidth}
        \centering
         \includegraphics[width=\textwidth, trim=0 0 0 0, clip]{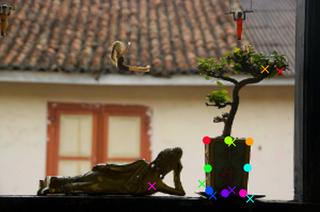}
     \end{subfigure}
     \hfill
     \begin{subfigure}[b]{0.19\textwidth}
        \centering
         \includegraphics[width=\textwidth,  trim=0 0 0 0, clip]{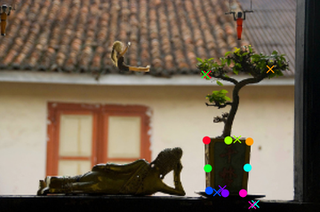}
     \end{subfigure}
     \hfill
     \begin{subfigure}[b]{0.19\textwidth}
        \centering
         \includegraphics[width=\textwidth,  trim=0 0 0 0, clip]{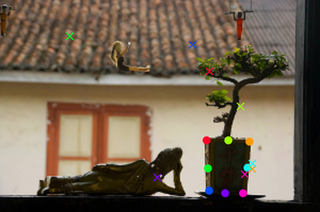}
     \end{subfigure}
     \hfill
     \begin{subfigure}[b]{0.19\textwidth}
        \centering
         \includegraphics[width=\textwidth,  trim=0 0 0 0, clip]{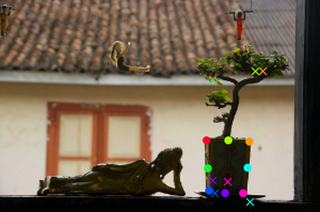}
     \end{subfigure}
     
     \centering
     \begin{subfigure}[b]{0.20\textwidth}
        \centering
         \includegraphics[width=\textwidth, trim=0 0 0 0, clip]{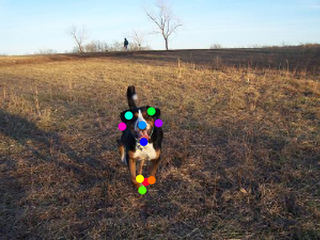}
     \end{subfigure}
     \hfill
     \begin{subfigure}[b]{0.18\textwidth}
        \centering
         \includegraphics[width=\textwidth, trim=0 0 0 0, clip]{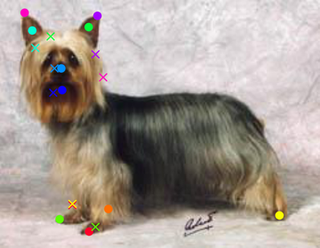}
     \end{subfigure}
     \hfill
     \begin{subfigure}[b]{0.18\textwidth}
        \centering
         \includegraphics[width=\textwidth,  trim=0 0 0 0, clip]{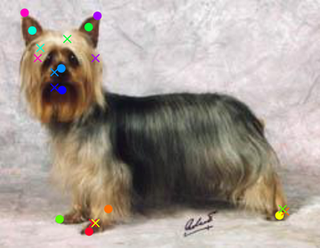}
     \end{subfigure}
     \hfill
     \begin{subfigure}[b]{0.18\textwidth}
        \centering
         \includegraphics[width=\textwidth,  trim=0 0 0 0, clip]{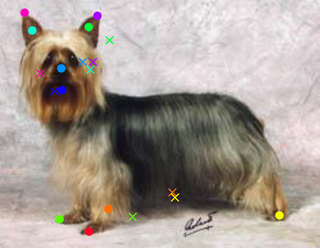}
     \end{subfigure}
     \hfill
     \begin{subfigure}[b]{0.18\textwidth}
        \centering
         \includegraphics[width=\textwidth,  trim=0 0 0 0, clip]{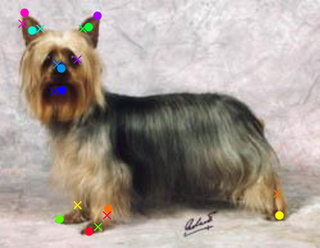}
     \end{subfigure}
     \centering
     \begin{subfigure}[b]{0.18\textwidth}
        \centering
         \includegraphics[width=\textwidth, trim=0 50 0 20, clip]{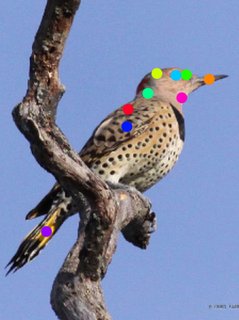}
     \end{subfigure}
     \hfill
     \begin{subfigure}[b]{0.19\textwidth}
     \centering
         \includegraphics[width=\textwidth, trim=0 0 0 0, clip]{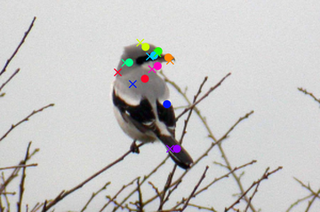}
     \end{subfigure}
     \hfill
      \begin{subfigure}[b]{0.19\textwidth}
     \centering
         \includegraphics[width=\textwidth, trim=0 0 0 0, clip]{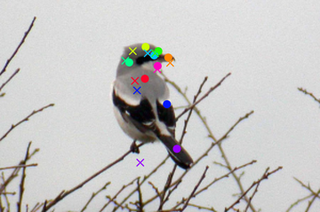}
     \end{subfigure}
     \hfill
          \begin{subfigure}[b]{0.19\textwidth}
     \centering
         \includegraphics[width=\textwidth, trim=0 0 0 0, clip]{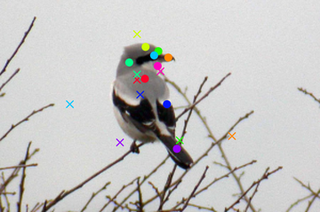}
     \end{subfigure}
     \hfill
          \begin{subfigure}[b]{0.19\textwidth}
     \centering
         \includegraphics[width=\textwidth, trim=0 0 0 0, clip]{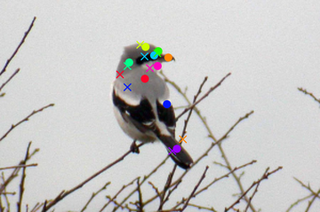}
     \end{subfigure}

     \centering
     \begin{subfigure}[b]{0.19\textwidth}
        \centering
         \includegraphics[width=\textwidth, trim=0 0 0 0, clip]{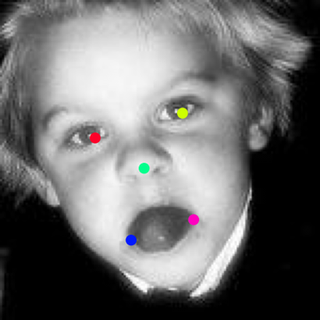}
     \end{subfigure}
     \hfill
     \begin{subfigure}[b]{0.19\textwidth}
        \centering
         \includegraphics[width=\textwidth, trim=0 0 0 0, clip]{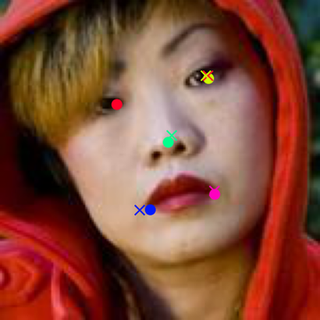}
     \end{subfigure}
     \hfill
     \begin{subfigure}[b]{0.19\textwidth}
        \centering
         \includegraphics[width=\textwidth,  trim=0 0 0 0, clip]{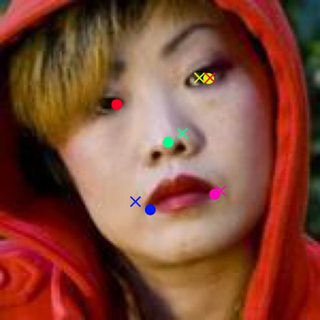}
     \end{subfigure}
     \hfill
     \begin{subfigure}[b]{0.19\textwidth}
        \centering
         \includegraphics[width=\textwidth,  trim=0 0 0 0, clip]{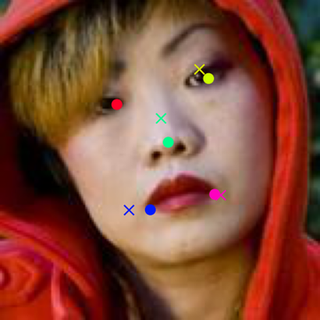}
     \end{subfigure}
     \hfill
     \begin{subfigure}[b]{0.19\textwidth}
        \centering
         \includegraphics[width=\textwidth,  trim=0 0 0 0, clip]{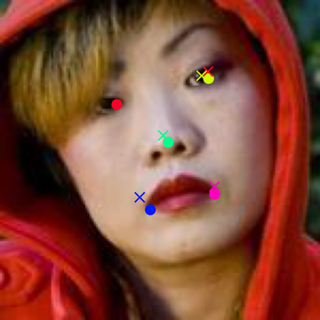}
     \end{subfigure}
     
     \centering
     \begin{subfigure}[b]{0.15\textwidth}
        \centering
         \includegraphics[width=\textwidth, trim=0 0 0 0, clip]{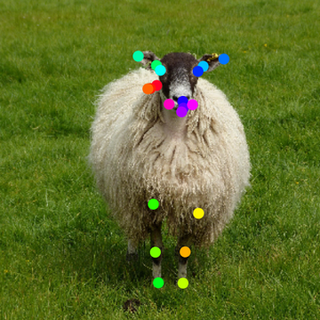} \caption{Source}
     \end{subfigure}
     \hfill
     \begin{subfigure}[b]{0.203\textwidth}
        \centering
         \includegraphics[width=\textwidth, trim=0 0 0 0, clip]{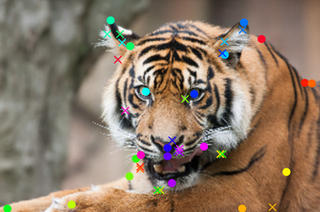} \caption{ASYM}
     \end{subfigure}
     \hfill
     \begin{subfigure}[b]{0.203\textwidth}
        \centering
         \includegraphics[width=\textwidth,  trim=0 0 0 0, clip]{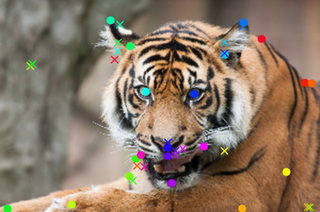} \caption{CL}
     \end{subfigure}
     \hfill
     \begin{subfigure}[b]{0.203\textwidth}
        \centering
         \includegraphics[width=\textwidth,  trim=0 0 0 0, clip]{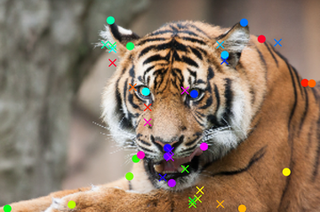}\caption{DVE}
     \end{subfigure}
     \hfill
     \begin{subfigure}[b]{0.203\textwidth} 
        \centering
         \includegraphics[width=\textwidth,  trim=0 0 0 0, clip]{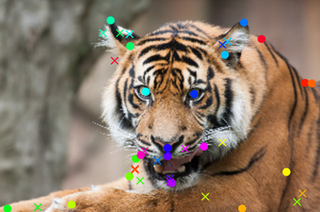}\caption{Sup}
     \end{subfigure}
        \caption{More qualitative matching results. 
        Each row is a different dataset: Spair, SDogs, CUB, AFLW, and Awa, from top to bottom. 
        The left most image for each row is a source example, and the remaining images visualize matches from different unsupervised methods, where 'o' indicates a ground truth location and  'x' indicates a prediction. For the Awa-Pose dataset example in the bottom row, all of the methods struggle as visual diversity is high between instances and the target example is in a different pose.}
        \label{fig:qual2}
\end{figure}

%% file: figs/qualitative3.tex
\begin{figure}
     \centering
     \begin{subfigure}[b]{0.18\textwidth}
        \centering
         \includegraphics[width=\textwidth, trim=0 0 0 0, clip]{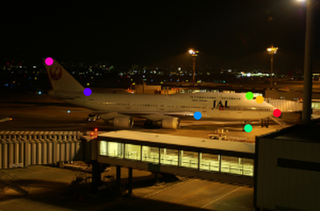}
     \end{subfigure}
     \hfill
     \begin{subfigure}[b]{0.19\textwidth}
        \centering
         \includegraphics[width=\textwidth, trim=0 0 0 0, clip]{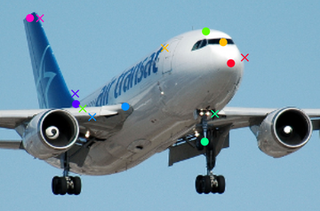}
     \end{subfigure}
     \hfill
     \begin{subfigure}[b]{0.19\textwidth}
        \centering
         \includegraphics[width=\textwidth,  trim=0 0 0 0, clip]{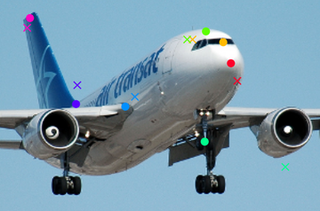}
     \end{subfigure}
     \hfill
     \begin{subfigure}[b]{0.19\textwidth}
        \centering
         \includegraphics[width=\textwidth,  trim=0 0 0 0, clip]{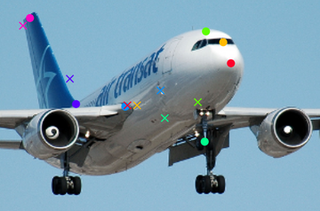}
     \end{subfigure}
     \hfill
     \begin{subfigure}[b]{0.19\textwidth}
        \centering
         \includegraphics[width=\textwidth,  trim=0 0 0 0, clip]{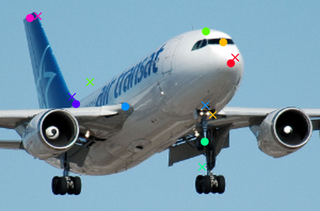}
     \end{subfigure}
     
     \centering
     \begin{subfigure}[b]{0.20\textwidth}
        \centering
         \includegraphics[width=\textwidth, trim=0 0 0 0, clip]{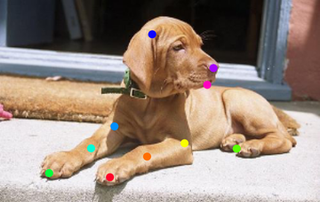}
     \end{subfigure}
     \hfill
     \begin{subfigure}[b]{0.18\textwidth}
        \centering
         \includegraphics[width=\textwidth, trim=0 0 0 0, clip]{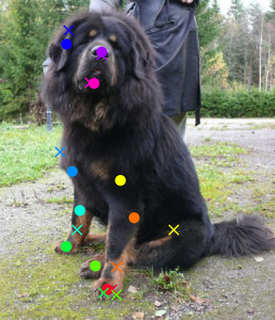}
     \end{subfigure}
     \hfill
     \begin{subfigure}[b]{0.18\textwidth}
        \centering
         \includegraphics[width=\textwidth,  trim=0 0 0 0, clip]{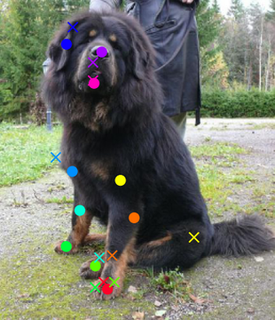}
     \end{subfigure}
     \hfill
     \begin{subfigure}[b]{0.18\textwidth}
        \centering
         \includegraphics[width=\textwidth,  trim=0 0 0 0, clip]{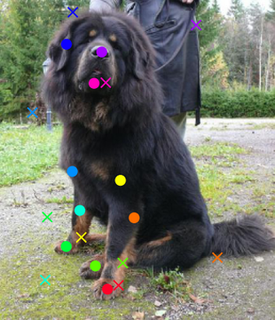}
     \end{subfigure}
     \hfill
     \begin{subfigure}[b]{0.18\textwidth}
        \centering
         \includegraphics[width=\textwidth,  trim=0 0 0 0, clip]{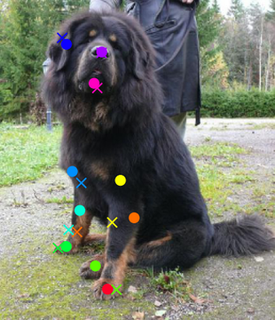}
     \end{subfigure}
     \centering
     \begin{subfigure}[b]{0.18\textwidth}
        \centering
         \includegraphics[width=\textwidth, trim=0 50 0 20, clip]{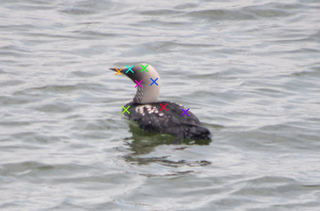}
     \end{subfigure}
     \hfill
     \begin{subfigure}[b]{0.19\textwidth}
     \centering
         \includegraphics[width=\textwidth, trim=0 0 0 0, clip]{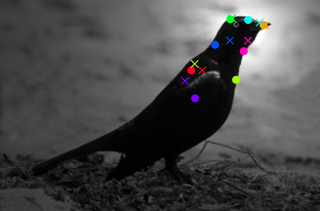}
     \end{subfigure}
     \hfill
      \begin{subfigure}[b]{0.19\textwidth}
     \centering
         \includegraphics[width=\textwidth, trim=0 0 0 0, clip]{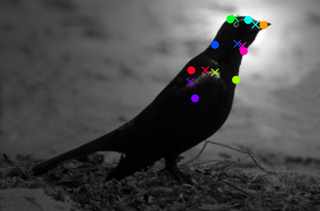}
     \end{subfigure}
     \hfill
          \begin{subfigure}[b]{0.19\textwidth}
     \centering
         \includegraphics[width=\textwidth, trim=0 0 0 0, clip]{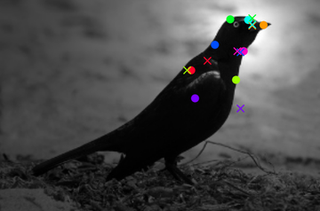}
     \end{subfigure}
     \hfill
          \begin{subfigure}[b]{0.19\textwidth}
     \centering
         \includegraphics[width=\textwidth, trim=0 0 0 0, clip]{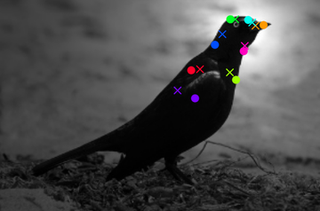}
     \end{subfigure}

     \centering
     \begin{subfigure}[b]{0.19\textwidth}
        \centering
         \includegraphics[width=\textwidth, trim=0 0 0 0, clip]{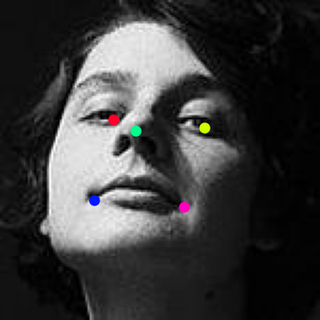}
     \end{subfigure}
     \hfill
     \begin{subfigure}[b]{0.19\textwidth}
        \centering
         \includegraphics[width=\textwidth, trim=0 0 0 0, clip]{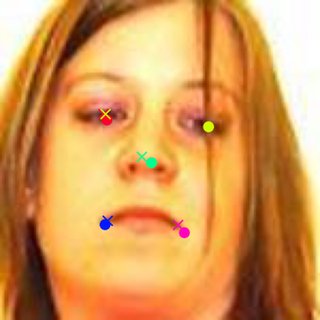}
     \end{subfigure}
     \hfill
     \begin{subfigure}[b]{0.19\textwidth}
        \centering
         \includegraphics[width=\textwidth,  trim=0 0 0 0, clip]{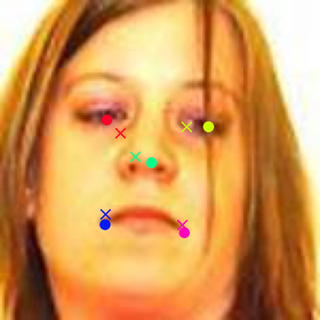}
     \end{subfigure}
     \hfill
     \begin{subfigure}[b]{0.19\textwidth}
        \centering
         \includegraphics[width=\textwidth,  trim=0 0 0 0, clip]{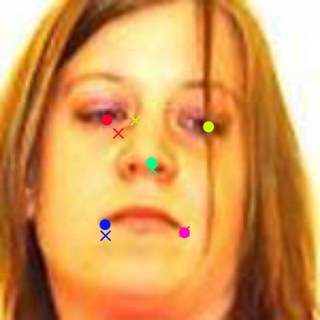}
     \end{subfigure}
     \hfill
     \begin{subfigure}[b]{0.19\textwidth}
        \centering
         \includegraphics[width=\textwidth,  trim=0 0 0 0, clip]{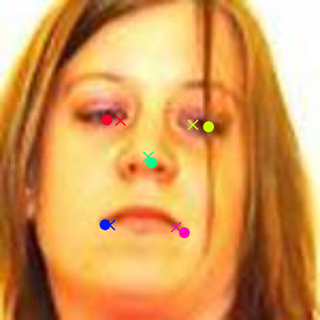}
     \end{subfigure}
     
     \centering
     \begin{subfigure}[b]{0.15\textwidth}
        \centering
         \includegraphics[width=\textwidth, trim=0 0 0 0, clip]{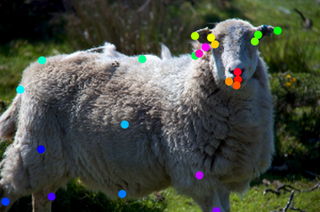} \caption{Source}
     \end{subfigure}
     \hfill
     \begin{subfigure}[b]{0.203\textwidth}
        \centering
         \includegraphics[width=\textwidth, trim=0 0 0 0, clip]{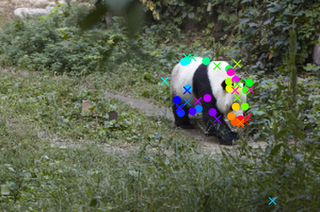} \caption{ASYM}
     \end{subfigure}
     \hfill
     \begin{subfigure}[b]{0.203\textwidth}
        \centering
         \includegraphics[width=\textwidth,  trim=0 0 0 0, clip]{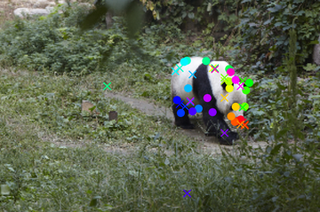} \caption{CL}
     \end{subfigure}
     \hfill
     \begin{subfigure}[b]{0.203\textwidth}
        \centering
         \includegraphics[width=\textwidth,  trim=0 0 0 0, clip]{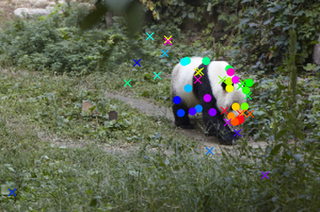}\caption{DVE}
     \end{subfigure}
     \hfill
     \begin{subfigure}[b]{0.203\textwidth} 
        \centering
         \includegraphics[width=\textwidth,  trim=0 0 0 0, clip]{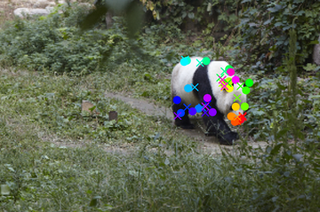}\caption{Sup}
     \end{subfigure}
        \caption{More qualitative matching results. 
        Each row is a different dataset: Spair, SDogs, CUB, AFLW, and Awa, from top to bottom. 
        The left most image for each row is a source example, and the remaining images visualize matches from different unsupervised methods, where 'o' indicates a ground truth location and  'x' indicates a prediction. While most methods perform reasonably good on the AFLW dataset instance, the predictions for the highly articulated objects (\eg animals), even the supervised baseline cannot obtain satisfactory results. }
        \label{fig:qual3}
\end{figure}

%% file: figs/tsne/tsne.tex
\begin{figure}
        \vspace{-0.0cm}
        \centering 
        \rotatebox{90}{\parbox{0.15\textwidth}{}}
        \parbox{0.15\textwidth}{\center None}
        \parbox{0.15\textwidth}{\center DVE}
        \parbox{0.15\textwidth}{\center CL}
        \parbox{0.15\textwidth}{\center LEAD}
        \parbox{0.15\textwidth}{\center ASYM}
        \parbox{0.15\textwidth}{\center Supervised}
        \rotatebox{90}{\parbox{0.15\textwidth}{\center AFLW}}
        \includegraphics[width=0.15\textwidth, trim=10 10 10 10, clip]{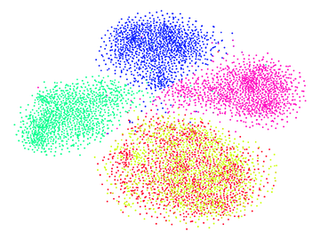}
        \includegraphics[width=0.15\textwidth, trim=10 10 10 10, clip]{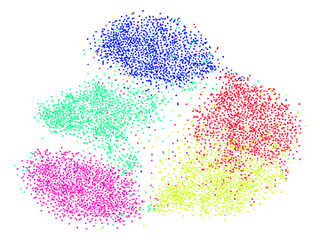}        
        \includegraphics[width=0.15\textwidth, trim=10 10 10 10, clip]{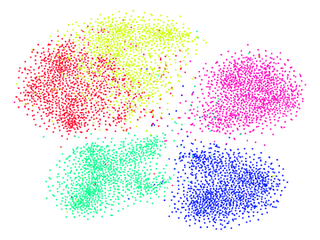}       
        \includegraphics[width=0.15\textwidth, trim=10 10 10 10, clip]{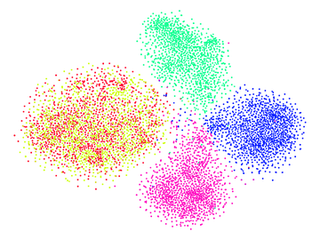}        
        \includegraphics[width=0.15\textwidth, trim=10 10 10 10, clip]{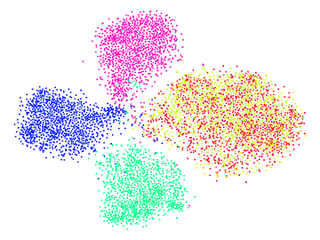}
        \includegraphics[width=0.15\textwidth, trim=10 10 10 10, clip]{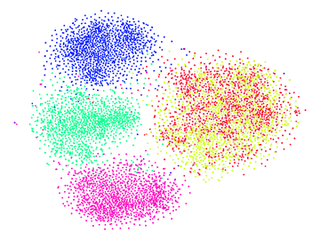}
        
        \rotatebox{90}{\parbox{0.15\textwidth}{\center CUB}}
        \includegraphics[width=0.15\textwidth, trim=10 10 10 10, clip]{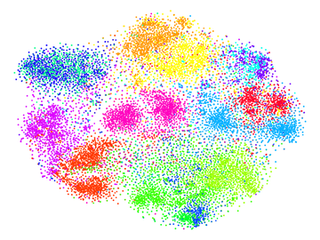}
        \includegraphics[width=0.15\textwidth, trim=10 10 10 10, clip]{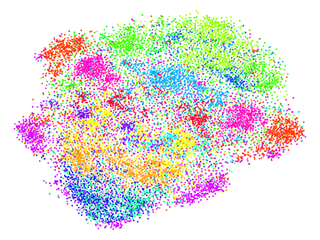}        
        \includegraphics[width=0.15\textwidth, trim=10 10 10 10, clip]{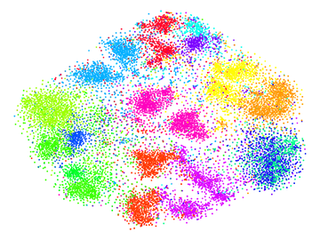}        
        \includegraphics[width=0.15\textwidth, trim=10 10 10 10, clip]{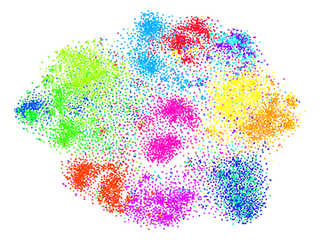}    
        \includegraphics[width=0.15\textwidth, trim=10 10 10 10, clip]{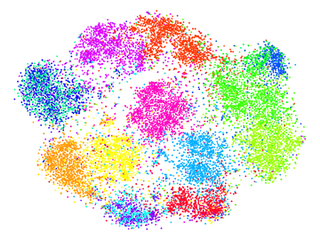}       
        \includegraphics[width=0.15\textwidth, trim=10 10 10 10, clip]{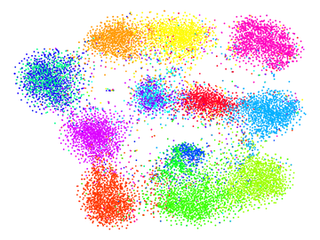}
      
        \rotatebox{90}{\parbox{0.15\textwidth}{\center SDogs}}
        \includegraphics[width=0.15\textwidth, trim=10 10 10 10, clip]{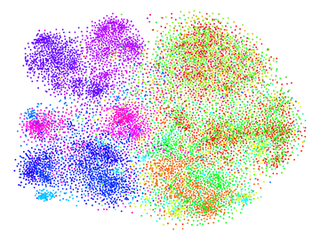}
        \includegraphics[width=0.15\textwidth, trim=10 10 10 10, clip]{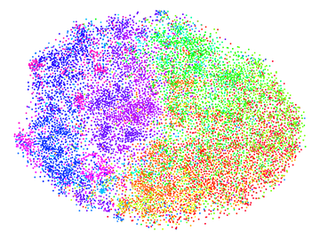}        \includegraphics[width=0.15\textwidth, trim=10 10 10 10, clip]{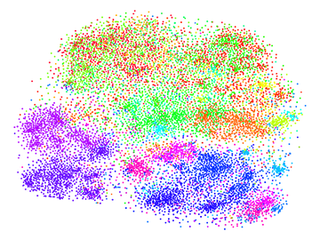}        \includegraphics[width=0.15\textwidth, trim=10 10 10 10, clip]{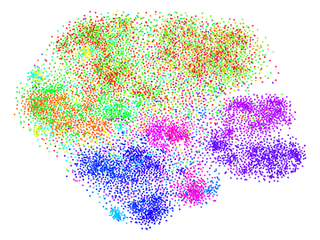}        \includegraphics[width=0.15\textwidth, trim=10 10 10 10, clip]{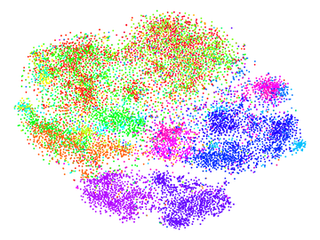}
        \includegraphics[width=0.15\textwidth, trim=10 10 10 10, clip]{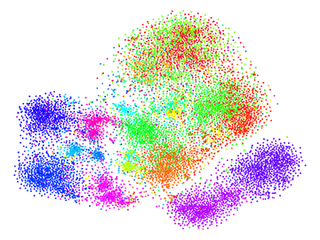}
        
    \caption{t-SNE visualization the embeddings learned by different unsupervised losses. Each row is a different dataset, and the colors indicate the ground truth identity of different keypoints.}
    \label{fig:tsne}
\end{figure}

%% file: tables/regression_exp.tex
\begin{table}
\footnotesize
\centering
\caption{Keypoint regression results with percentage of inter-ocular distance.
The rows marked as `Original' are numbers taken from the original papers and differ in the network architecture and in some cases the amount of supervision used. 
Note that AFLW is referred to as $\text{AFLW}_{M}$ in some of the works below. 
The numerical scores represent the percentage of inter-ocular distance, where lower scores are better. 
}
\begin{tabular}{lllcc}
\toprule
Implementation~~ & Method~~ & Feat.dim.~~ & MAFL~~ & AFLW \\ \midrule
\multirow{3}{*}{Original}& DVE & 64 & 2.86 & 7.53  \\ 
&CL & 256 & 2.64 & 7.17  \\ 
&LEAD & 256 & 2.87 & 6.51  \\ \midrule
\multirow{4}{*}{Ours}& DVE & 256 & 3.07 & 8.57  \\ 
&CL & 256 & 2.96 & 7.73  \\ 
&LEAD & 256 & 2.80 &  7.97   \\ 
&ASYM & 256 & 2.94 & 7.98  \\
\bottomrule
\end{tabular}

\label{table:regression}
\end{table}

%% file: tables/pretraining_dataset.tex
\begin{table}
\centering
\caption{Results for using different sources of pre-training dataset. These numbers correspond to those presented in Fig.~3 in the main paper.} 
\subfloat[Imagenet]{
\resizebox{0.32\linewidth}{!}{
\begin{tabular}{lccccc}
\toprule
Projection($\rho$)~  &  Spair-71K & SDogs & CUB & AFLW & Awa \\ \midrule
None & 30.7 & 34.3 & 47.5 & 64.3 & 27.6  \\
NMF  & 20.6 & 19.9 & 44.0 & 40.8 & 15.6  \\
PCA  & 27.4 & 29.8 & 50.7 & 51.0 & 24.1  \\ 
Random  & 26.6 & 31.5 & 40.0 & 60.2 & 23.3  \\\midrule
Supervised  & 39.5 & 54.0 & 73.4 & 83.8 & 48.2  \\ \midrule
EQ\cite{thewlis2017unsupervisedneurips} & 14.3 & 20.5 & 26.4 & 62.8 & 15.5  \\
DVE\cite{thewlis2019unsupervised} &  15.0 & 19.4 & 28.7 & 60.6 & 14.7  \\
CL\cite{cheng2021equivariant} & 29.7 & 37.9 & 54.1 & 77.1 & 33.4 \\
LEAD\cite{karmali2022lead}  & 30.5 & 34.4 & 48.3 & 64.9 & 28.1  \\\midrule
ASYM (Ours) & 33.2 & 38.2 & 54.4 & 69.7 & 32.1  \\
\bottomrule
\end{tabular}
}
}
\subfloat[iNat2021]{
\resizebox{0.32\linewidth}{!}{
\begin{tabular}{lccccc}
\toprule
Projection($\rho$)~  &  Spair-71K & SDogs & CUB & AFLW & Awa \\ \midrule
None & 21.6 & 19.3 & 44.5 & 42.0 & 16.1  \\ 
NMF  & 18.8 & 17.9 & 45.2 & 33.6 & 15.6   \\
PCA  & 21.7 & 20.2 & 45.2 & 42.2 & 16.7  \\ 
Random  & 17.0 & 14.5 & 35.8 & 37.2 & 12.3  \\
\midrule
Supervised  & 28.1 & 36.4 & 70.6 & 58.6 & 32.6  \\ \midrule
EQ\cite{thewlis2017unsupervisedneurips}& 10.7 & 15.4 & 26.3 & 40.8 & 11.7 \\
DVE\cite{thewlis2019unsupervised} & 10.6 & 15.4 & 25.8 & 38.5 & 11.2 \\
CL\cite{cheng2021equivariant} & 19.9 & 19.9 & 51.9 & 44.8 & 18.1  \\
LEAD\cite{karmali2022lead} & 21.1 & 19.4 & 44.1 & 41.9 & 16.0  \\  \midrule
ASYM (Ours)  & 21.8 & 21.8 & 51.7 & 44.4 & 17.7  \\
\bottomrule
\end{tabular}
}
}
\subfloat[CelebA]{
\resizebox{0.32\linewidth}{!}{
\begin{tabular}{lccccc}
\toprule
Projection($\rho$)~  &  Spair-71K & SDogs & CUB & AFLW & Awa \\ \midrule
None & 11.6 & 8.8 & 13.6 & 50.3 & 8.0  \\ 
NMF  & 10.0 & 8.5 & 12.4 & 47.6 & 8.0   \\
PCA  & 11.7 & 9.0 & 13.8 & 51.2 & 8.3  \\
Random  & 10.4 & 8.1 & 11.8 & 43.7 & 7.5  \\  \midrule
Supervised  & 14.3 & 17.4 & 28.0 & 65.2 & 14.4  \\ \midrule
EQ\cite{thewlis2017unsupervisedneurips}& 8.6 & 9.5 & 12.1 & 54.0 & 8.1  \\
DVE\cite{thewlis2019unsupervised} & 9.0 & 9.4 & 12.4 & 48.3 & 8.1  \\
CL\cite{cheng2021equivariant} & 10.8 & 10.1 & 12.8 & 62.4 & 8.0 \\
LEAD\cite{karmali2022lead} & 11.5 & 8.8 & 13.3 & 50.0 & 8.0  \\  \midrule
ASYM (Ours)  & 11.5 & 8.9 & 13.4 & 60.7 & 8.0  \\
\bottomrule
\end{tabular}
}
}
\label{table:different_datasource}
\end{table}

%% file: tables/cross_proj_sup.tex
\begin{table}
\caption{Cross dataset evaluation results. These results use the  `Sup. pre-trained - CNN' and correspond to the results in Fig.~4 in the main paper.} 
\subfloat[CL]{
\resizebox{0.48\linewidth}{!}{
\begin{tabular}{l|ccccc}
Test/Train  & Spair-71K & SDogs & CUB & AFLW & Awa \\ \midrule
Spair-71K&30.8&31.1&31.4&29.1&31.5 \\
SDogs&36.4&37.0&36.8&35.4&36.9 \\
CUB&49.1&47.5&54.5&45.6&48.3 \\
AFLW&62.7&62.1&62.2&67.3&62.7 \\
Awa&30.6&29.9&30.1&27.0&31.7 \\
\bottomrule
\end{tabular}
}
}
\subfloat[ASYM]{
\resizebox{0.48\linewidth}{!}{
\begin{tabular}{l|ccccc}
Test/Train & Spair-71K & SDogs & CUB & AFLW & Awa \\ \midrule
Spair-71K & 34.0 & 30.9 & 28.3 & 25.9 & 30.2 \\
SDogs & 38.4 & 40.4 & 31.1 & 30.9 & 38.3 \\
CUB & 56.2 & 50.5 & 60.8 & 42.8 & 51.1 \\
AFLW & 54.4 & 58.2 & 48.6 & 63.6 & 56.3 \\
Awa & 33.5 & 33.9 & 26.6 & 25.4 & 34.1 \\
\bottomrule
\end{tabular}
}
}

\subfloat[DVE]{
\resizebox{0.48\linewidth}{!}{
\begin{tabular}{l|ccccc}
Test/Train & Spair-71K & SDogs & CUB & AFLW & Awa \\ \midrule
Spair-71K&16.3&13.9&15.5&17.3&14.7 \\
SDogs&21.9&20.5&21.3&23.3&20.3 \\
CUB&26.2&24.1&27.7&25.2&23.4 \\
AFLW&41.0&41.2&43.9&58.7&41.4 \\
Awa&16.0&14.2&14.8&17.6&15.4 \\
\bottomrule
\end{tabular}
}
}
\subfloat[Supervised]{
\resizebox{0.48\linewidth}{!}{
\begin{tabular}{l|ccccc}
Test/Train  & Spair-71K & SDogs & CUB & AFLW & Awa \\ \midrule
Spair-71K & 38.7 & 26.7 & 24.5 & 17.4 & 29.2 \\
SDogs & 40.1 & 53.2 & 29.0 & 25.1 & 42.9 \\
CUB & 52.5 & 40.2 & 72.7 & 25.4 & 47.6 \\
AFLW & 57.4 & 56.6 & 46.5 & 80.8 & 58.7 \\
Awa & 35.1 & 34.9 & 26.9 & 18.1 & 46.1 \\
\bottomrule
\end{tabular}
}
}
\label{table:cross_projection_sup}
\end{table}

%% file: main.bbl
\begin{thebibliography}{10}
\providecommand{\url}[1]{\texttt{#1}}
\providecommand{\urlprefix}{URL }
\providecommand{\doi}[1]{https://doi.org/#1}

\bibitem{alwassel2018diagnosing}
Alwassel, H., Heilbron, F.C., Escorcia, V., Ghanem, B.: Diagnosing error in
  temporal action detectors. In: ECCV. pp. 256--272 (2018)

\bibitem{amir2021deep}
Amir, S., Gandelsman, Y., Bagon, S., Dekel, T.: Deep vit features as dense
  visual descriptors. arXiv:2112.05814  (2021)

\bibitem{araslanov2021dense}
Araslanov, N., Schaub-Meyer, S., Roth, S.: Dense unsupervised learning for
  video segmentation. NeurIPS  (2021)

\bibitem{awapose}
Banik, P., Li, L., Dong, X.: A novel dataset for keypoint detection of
  quadruped animals from images. arXiv:2108.13958  (2021)

\bibitem{biggs2020left}
Biggs, B., Boyne, O., Charles, J., Fitzgibbon, A., Cipolla, R.: Who left the
  dogs out? 3d animal reconstruction with expectation maximization in the loop.
  In: ECCV (2020)

\bibitem{bristow2015dense}
Bristow, H., Valmadre, J., Lucey, S.: Dense semantic correspondence where every
  pixel is a classifier. In: ICCV. pp. 4024--4031 (2015)

\bibitem{caron2021emerging}
Caron, M., Touvron, H., Misra, I., J\'egou, H., Mairal, J., Bojanowski, P.,
  Joulin, A.: Emerging properties in self-supervised vision transformers. In:
  ICCV (2021)

\bibitem{chen2020simple}
Chen, T., Kornblith, S., Norouzi, M., Hinton, G.: A simple framework for
  contrastive learning of visual representations. In: ICML (2020)

\bibitem{mocov2}
Chen, X., Fan, H., Girshick, R., He, K.: Improved baselines with momentum
  contrastive learning. arXiv:2003.04297  (2020)

\bibitem{chen2021mocov3}
Chen, X., Xie, S., He, K.: An empirical study of training self-supervised
  vision transformers. arXiv:2104.02057  (2021)

\bibitem{cheng2021equivariant}
Cheng, Z., Su, J.C., Maji, S.: On equivariant and invariant learning of object
  landmark representations. In: ICCV (2021)

\bibitem{cho2021cats}
Cho, S., Hong, S., Jeon, S., Lee, Y., Sohn, K., Kim, S.: Cats: Cost aggregation
  transformers for visual correspondence. NeurIPS  (2021)

\bibitem{choe2020evaluating}
Choe, J., Oh, S.J., Lee, S., Chun, S., Akata, Z., Shim, H.: Evaluating weakly
  supervised object localization methods right. In: CVPR. pp. 3133--3142 (2020)

\bibitem{choy2016universal}
Choy, C.B., Gwak, J., Savarese, S., Chandraker, M.: Universal correspondence
  network. NeurIPS  (2016)

\bibitem{david2016correspondence}
David, M.: The correspondence theory of truth. The Oxford Handbook of Truth
  (2016)

\bibitem{everingham2015pascal}
Everingham, M., Eslami, S.A., Van~Gool, L., Williams, C.K., Winn, J.,
  Zisserman, A.: The pascal visual object classes challenge: A retrospective.
  In: IJCV (2015)

\bibitem{gonzalez2018semantic}
Gonzalez-Garcia, A., Modolo, D., Ferrari, V.: Do semantic parts emerge in
  convolutional neural networks? IJCV  (2018)

\bibitem{grill2020bootstrap}
Grill, J.B., Strub, F., Altch{\'e}, F., Tallec, C., Richemond, P., Buchatskaya,
  E., Doersch, C., Avila~Pires, B., Guo, Z., Gheshlaghi~Azar, M., et~al.:
  Bootstrap your own latent-a new approach to self-supervised learning. NeurIPS
   (2020)

\bibitem{ham2016proposal}
Ham, B., Cho, M., Schmid, C., Ponce, J.: Proposal flow. In: CVPR (2016)

\bibitem{han2017scnet}
Han, K., Rezende, R.S., Ham, B., Wong, K.Y.K., Cho, M., Schmid, C., Ponce, J.:
  Scnet: Learning semantic correspondence. In: ICCV. pp. 1831--1840 (2017)

\bibitem{he2020momentum}
He, K., Fan, H., Wu, Y., Xie, S., Girshick, R.: Momentum contrast for
  unsupervised visual representation learning. In: CVPR (2020)

\bibitem{he2016deep}
He, K., Zhang, X., Ren, S., Sun, J.: Deep residual learning for image
  recognition. In: CVPR (2016)

\bibitem{hoiem2012diagnosing}
Hoiem, D., Chodpathumwan, Y., Dai, Q.: Diagnosing error in object detectors.
  In: ECCV. pp. 340--353 (2012)

\bibitem{huang2019dynamic}
Huang, S., Wang, Q., Zhang, S., Yan, S., He, X.: Dynamic context correspondence
  network for semantic alignment. In: ICCV. pp. 2010--2019 (2019)

\bibitem{jakab2018unsupervised}
Jakab, T., Gupta, A., Bilen, H., Vedaldi, A.: Unsupervised learning of object
  landmarks through conditional image generation. NeurIPS  (2018)

\bibitem{jakab2020self}
Jakab, T., Gupta, A., Bilen, H., Vedaldi, A.: Self-supervised learning of
  interpretable keypoints from unlabelled videos. In: CVPR (2020)

\bibitem{jiang2021cotr}
Jiang, W., Trulls, E., Hosang, J., Tagliasacchi, A., Yi, K.M.: Cotr:
  Correspondence transformer for matching across images. In: ICCV. pp.
  6207--6217 (2021)

\bibitem{kanazawa2016warpnet}
Kanazawa, A., Jacobs, D.W., Chandraker, M.: Warpnet: Weakly supervised matching
  for single-view reconstruction. In: CVPR (2016)

\bibitem{karmali2022lead}
Karmali, T., Atrishi, A., Harsha, S.S., Agrawal, S., Jampani, V., Babu, R.V.:
  Lead: Self-supervised landmark estimation by aligning distributions of
  feature similarity. In: WACV (2022)

\bibitem{khosla2011novel}
Khosla, A., Jayadevaprakash, N., Yao, B., Li, F.F.: Novel dataset for
  fine-grained image categorization: Stanford dogs. In: CVPR Workshop on
  Fine-Grained Visual Categorization (2011)

\bibitem{kim2013deformable}
Kim, J., Liu, C., Sha, F., Grauman, K.: Deformable spatial pyramid matching for
  fast dense correspondences. In: CVPR. pp. 2307--2314 (2013)

\bibitem{kim2018recurrent}
Kim, S., Lin, S., Jeon, S.R., Min, D., Sohn, K.: Recurrent transformer networks
  for semantic correspondence. NeurIPS  (2018)

\bibitem{kim2017fcss}
Kim, S., Min, D., Ham, B., Jeon, S., Lin, S., Sohn, K.: Fcss: Fully
  convolutional self-similarity for dense semantic correspondence. In: CVPR.
  pp. 6560--6569 (2017)

\bibitem{kingma2014adam}
Kingma, D.P., Ba, J.: Adam: A method for stochastic optimization.
  arXiv:1412.6980  (2014)

\bibitem{AFLW}
Koestinger, M., Wohlhart, P., Roth, P.M., Bischof, H.: Annotated facial
  landmarks in the wild: A large-scale, real-world database for facial landmark
  localization. In: ICCV workshops (2011)

\bibitem{dosovitskiy2020image}
Kolesnikov, A., Dosovitskiy, A., Weissenborn, D., Heigold, G., Uszkoreit, J.,
  Beyer, L., Minderer, M., Dehghani, M., Houlsby, N., Gelly, S., Unterthiner,
  T., Zhai, X.: An image is worth 16x16 words: Transformers for image
  recognition at scale. In: ICLR (2021)

\bibitem{kulkarni2019unsupervised}
Kulkarni, T.D., Gupta, A., Ionescu, C., Borgeaud, S., Reynolds, M., Zisserman,
  A., Mnih, V.: Unsupervised learning of object keypoints for perception and
  control. NeurIPS  (2019)

\bibitem{lee2021patchmatch}
Lee, J.Y., DeGol, J., Fragoso, V., Sinha, S.N.: Patchmatch-based neighborhood
  consensus for semantic correspondence. In: CVPR. pp. 13153--13163 (2021)

\bibitem{lee2019sfnet}
Lee, J., Kim, D., Ponce, J., Ham, B.: Sfnet: Learning object-aware semantic
  correspondence. In: CVPR. pp. 2278--2287 (2019)

\bibitem{li2020correspondence}
Li, S., Han, K., Costain, T.W., Howard-Jenkins, H., Prisacariu, V.:
  Correspondence networks with adaptive neighbourhood consensus. In: CVPR. pp.
  10196--10205 (2020)

\bibitem{liu2010sift}
Liu, C., Yuen, J., Torralba, A.: Sift flow: Dense correspondence across scenes
  and its applications. PAMI  (2010)

\bibitem{liu2020semantic}
Liu, Y., Zhu, L., Yamada, M., Yang, Y.: Semantic correspondence as an optimal
  transport problem. In: CVPR. pp. 4463--4472 (2020)

\bibitem{liu2015deep}
Liu, Z., Luo, P., Wang, X., Tang, X.: Deep learning face attributes in the
  wild. In: ICCV (2015)

\bibitem{long2014convnets}
Long, J.L., Zhang, N., Darrell, T.: Do convnets learn correspondence? NeurIPS
  (2014)

\bibitem{van2008visualizing}
Van~der Maaten, L., Hinton, G.: Visualizing data using t-sne. JMLR  (2008)

\bibitem{min2021convolutional}
Min, J., Cho, M.: Convolutional hough matching networks. In: CVPR (2021)

\bibitem{min2019hyperpixel}
Min, J., Lee, J., Ponce, J., Cho, M.: Hyperpixel flow: Semantic correspondence
  with multi-layer neural features. In: ICCV (2019)

\bibitem{min2019spair}
Min, J., Lee, J., Ponce, J., Cho, M.: Spair-71k: A large-scale benchmark for
  semantic correspondence. arXiv:1908.10543  (2019)

\bibitem{min2020learning}
Min, J., Lee, J., Ponce, J., Cho, M.: Learning to compose hypercolumns for
  visual correspondence. In: ECCV. pp. 346--363 (2020)

\bibitem{musgrave2020metric}
Musgrave, K., Belongie, S., Lim, S.N.: A metric learning reality check. In:
  ECCV. pp. 681--699 (2020)

\bibitem{o2020unsupervised}
O~Pinheiro, P.O., Almahairi, A., Benmalek, R., Golemo, F., Courville, A.C.:
  Unsupervised learning of dense visual representations. NeurIPS  (2020)

\bibitem{van2018representation}
Van~den Oord, A., Li, Y., Vinyals, O.: Representation learning with contrastive
  predictive coding. arXiv:1807.03748  (2018)

\bibitem{rocco2017convolutional}
Rocco, I., Arandjelovic, R., Sivic, J.: Convolutional neural network
  architecture for geometric matching. In: CVPR. pp. 6148--6157 (2017)

\bibitem{rocco2020efficient}
Rocco, I., Arandjelovi{\'c}, R., Sivic, J.: Efficient neighbourhood consensus
  networks via submanifold sparse convolutions. In: ECCV. pp. 605--621 (2020)

\bibitem{rocco2018neighbourhood}
Rocco, I., Cimpoi, M., Arandjelovi{\'c}, R., Torii, A., Pajdla, T., Sivic, J.:
  Neighbourhood consensus networks. NeurIPS  (2018)

\bibitem{roh2021spatially}
Roh, B., Shin, W., Kim, I., Kim, S.: Spatially consistent representation
  learning. In: CVPR. pp. 1144--1153 (2021)

\bibitem{ruggero2017benchmarking}
Ruggero~Ronchi, M., Perona, P.: Benchmarking and error diagnosis in
  multi-instance pose estimation. In: ICCV (2017)

\bibitem{russakovsky2013detecting}
Russakovsky, O., Deng, J., Huang, Z., Berg, A.C., Fei-Fei, L.: Detecting
  avocados to zucchinis: what have we done, and where are we going? In: ICCV
  (2013)

\bibitem{russakovsky2015imagenet}
Russakovsky, O., Deng, J., Su, H., Krause, J., Satheesh, S., Ma, S., Huang, Z.,
  Karpathy, A., Khosla, A., Bernstein, M., et~al.: Imagenet large scale visual
  recognition challenge. IJCV  (2015)

\bibitem{ryou2021weakly}
Ryou, S., Perona, P.: Weakly supervised keypoint discovery. arXiv:2109.13423
  (2021)

\bibitem{sarlin2020superglue}
Sarlin, P.E., DeTone, D., Malisiewicz, T., Rabinovich, A.: Superglue: Learning
  feature matching with graph neural networks. In: CVPR. pp. 4938--4947 (2020)

\bibitem{sigurdsson2017actions}
Sigurdsson, G.A., Russakovsky, O., Gupta, A.: What actions are needed for
  understanding human actions in videos? In: ICCV. pp. 2137--2146 (2017)

\bibitem{thewlis2019unsupervised}
Thewlis, J., Albanie, S., Bilen, H., Vedaldi, A.: Unsupervised learning of
  landmarks by descriptor vector exchange. In: ICCV (2019)

\bibitem{thewlis2017unsupervisedneurips}
Thewlis, J., Bilen, H., Vedaldi, A.: Unsupervised learning of object frames by
  dense equivariant image labelling. NeurIPS  (2017)

\bibitem{thewlis2017unsupervised}
Thewlis, J., Bilen, H., Vedaldi, A.: Unsupervised learning of object landmarks
  by factorized spatial embeddings. In: ICCV (2017)

\bibitem{ufer2017deep}
Ufer, N., Ommer, B.: Deep semantic feature matching. In: CVPR. pp. 6914--6923
  (2017)

\bibitem{van2021benchmarking}
Van~Horn, G., Cole, E., Beery, S., Wilber, K., Belongie, S., Mac~Aodha, O.:
  Benchmarking representation learning for natural world image collections. In:
  CVPR (2021)

\bibitem{CUB}
Wah, C., Branson, S., Welinder, P., Perona, P., Belongie, S.: The caltech-ucsd
  birds-200-2011 dataset  (2011)

\bibitem{wang2021dense}
Wang, X., Zhang, R., Shen, C., Kong, T., Li, L.: Dense contrastive learning for
  self-supervised visual pre-training. In: CVPR. pp. 3024--3033 (2021)

\bibitem{wang2021exploring}
Wang, Z., Li, Q., Zhang, G., Wan, P., Zheng, W., Wang, N., Gong, M., Liu, T.:
  Exploring set similarity for dense self-supervised representation learning.
  arXiv:2107.08712  (2021)

\bibitem{wei2021aligning}
Wei, F., Gao, Y., Wu, Z., Hu, H., Lin, S.: Aligning pretraining for detection
  via object-level contrastive learning. NeurIPS  (2021)

\bibitem{wu2018unsupervised}
Wu, Z., Xiong, Y., Yu, S.X., Lin, D.: Unsupervised feature learning via
  non-parametric instance discrimination. In: CVPR (2018)

\bibitem{xian2018zero}
Xian, Y., Lampert, C.H., Schiele, B., Akata, Z.: Zero-shot learning—a
  comprehensive evaluation of the good, the bad and the ugly. PAMI  (2018)

\bibitem{zeiler2014visualizing}
Zeiler, M.D., Fergus, R.: Visualizing and understanding convolutional networks.
  In: ECCV (2014)

\bibitem{zhang2016far}
Zhang, S., Benenson, R., Omran, M., Hosang, J., Schiele, B.: How far are we
  from solving pedestrian detection? In: CVPR. pp. 1259--1267 (2016)

\bibitem{zhang2018unsupervised}
Zhang, Y., Guo, Y., Jin, Y., Luo, Y., He, Z., Lee, H.: Unsupervised discovery
  of object landmarks as structural representations. In: CVPR (2018)

\bibitem{MTFL}
Zhang, Z., Luo, P., Loy, C.C., Tang, X.: Facial landmark detection by deep
  multi-task learning. In: European conference on computer vision. pp. 94--108.
  Springer (2014)

\bibitem{MAFL}
Zhang, Z., Luo, P., Loy, C.C., Tang, X.: Learning deep representation for face
  alignment with auxiliary attributes. PAMI  (2015)

\bibitem{zhao2021multi}
Zhao, D., Song, Z., Ji, Z., Zhao, G., Ge, W., Yu, Y.: Multi-scale matching
  networks for semantic correspondence. In: ICCV. pp. 3354--3364 (2021)

\bibitem{zhong2021pixel}
Zhong, Y., Yuan, B., Wu, H., Yuan, Z., Peng, J., Wang, Y.X.: Pixel
  contrastive-consistent semi-supervised semantic segmentation. In: ICCV. pp.
  7273--7282 (2021)

\bibitem{zhou2016cvpr}
Zhou, B., Khosla, A., Lapedriza, A., Oliva, A., Torralba, A.: Learning deep
  features for discriminative localization. In: CVPR (2016)

\end{thebibliography}
